\documentclass{article}

     \usepackage[final]{neurips_2025}

\usepackage{natbib}
\usepackage{graphicx}
\usepackage{amsmath}
\usepackage{makecell}
\setcitestyle{numbers,square}
\usepackage[utf8]{inputenc} % allow utf-8 input
\usepackage[T1]{fontenc}    % use 8-bit T1 fonts
\usepackage{hyperref}       % hyperlinks
\usepackage{url}            % simple URL typesetting
\usepackage{booktabs}       % professional-quality tables
\usepackage{amsfonts}       % blackboard math symbols
\usepackage{nicefrac}       % compact symbols for 1/2, etc.

\usepackage{microtype}      % microtypography
\usepackage{xcolor}         % colors

\usepackage[most]{tcolorbox}
\usepackage{colortbl}
\usepackage{multirow}
\usepackage{graphicx}
\usepackage{lipsum}
\usepackage{tablefootnote}
\usepackage{tcolorbox}
\usepackage[tikz]{bclogo}
\usepackage{subfigure}
\usepackage{subcaption}
\usepackage{multirow}
\usepackage{wrapfig}

\usepackage{listings} % 代码风格展示
\usepackage{CJKutf8}

\usepackage{tabularx}

\newcommand{\Romannum}[1]{\texorpdfstring{\uppercase\expandafter{\romannumeral #1}}{#1}}

\newtcolorbox{titleEnv}{
colframe=black!80,
colback=gray!10,
fonttitle=\bfseries,
coltitle=black,
left=3pt,
right=3pt,
top=3pt,
bottom=3pt,
boxrule=0.4mm,
arc=3mm
}
\definecolor{mydeepblue}{RGB}{46, 90, 168}
\definecolor{myblue}{RGB}{166, 202, 236}
\definecolor{my_blue}{RGB}{0,120,255}
\definecolor{my_purple}{RGB}{161, 27, 155}
\definecolor{my_green}{RGB}{0, 176, 80}
\definecolor{msftBlue}{RGB}{0,164,239}
\definecolor{msftGreen}{RGB}{127,186,0}
\definecolor{msftYello}{RGB}{255,185,0}
\definecolor{msftBlack}{RGB}{0,0,0}

\newenvironment{findingBox}[2]{%
	\begin{tcolorbox}[
colframe=mydeepblue!80,
colback=myblue!50,
 boxrule=.5pt,
 left=1pt,
 right = 1pt,
 top= 0pt,
 bottom=0pt,
 size=small,
 fonttitle=\bfseries,
coltitle=black,
boxrule=0.4mm,
arc=2mm
 ]{\textbf{Finding #1:} #2} %~\itshape %
}{%
	\end{tcolorbox}
}

\title{EvaLearn: Quantifying the Learning Capability and Efficiency of LLMs via Sequential Problem Solving
}

\author{\textbf{Shihan Dou}$^{1}$\thanks{{ }{ }Equal contribution.},
\ \textbf{Ming Zhang}$^{1*}$,
\ \textbf{Chenhao Huang}$^{1}$, 
\ \textbf{Jiayi Chen}$^{1}$,
\ \textbf{Feng Chen}$^{2,3}$,
\\
\ \textbf{Shichun Liu}$^{1}$,
\ \textbf{Yan Liu}$^{2}$,
\ \textbf{Chenxiao Liu}$^{2}$,
\ \textbf{Cheng Zhong}$^{2}$,
\ \textbf{Zongzhang Zhang}$^{3}$,
\ \textbf{Tao Gui}$^{1,4}$\thanks{{ }{ }Corresponding author.},
\\
\ \textbf{Chao Xin}$^{2}$,
\ \textbf{Chengzhi Wei}$^{2}$,
\ \textbf{Lin Yan}$^{2}$,
\ \textbf{Yonghui Wu}$^{2}$,
\ \textbf{Qi Zhang}$^{1,5\dagger}$,
\ \textbf{Xuanjing Huang}$^{1,5\dagger}$
\\
$^{1}$Fudan University \ 
$^{2}$ByteDance Seed \ 
$^{3}$Nanjing University \
$^{4}$Shanghai Innovation Institute \\
$^{5}$Shanghai Collaborative Innovation Center of Intelligent Visual Computing\\
\texttt{\{shdou24, mingzhang23\}@m.fudan.edu.cn}   \\
\texttt{\{tgui, qz, xjhuang\}@fudan.edu.cn}
}

\begin{document}

\maketitle

\begin{abstract}

We introduce \textbf{EvaLearn}, a pioneering benchmark designed to evaluate large language models (LLMs) on their learning capability and efficiency in challenging tasks, a critical, yet underexplored aspect of model potential. 
EvaLearn contains 648 challenging problems across six task types, grouped into 182 sequences, each sequence dedicated to one task type.
Diverging from most existing benchmarks that evaluate models in parallel, EvaLearn requires models to solve problems sequentially, allowing them to leverage the experience gained from previous solutions. 
EvaLearn provides five comprehensive automated metrics to evaluate models and quantify their learning capability and efficiency.
We extensively benchmark nine frontier models and observe varied performance profiles: some models, such as Claude-3.7-sonnet, start with moderate initial performance but exhibit strong learning ability, while some models struggle to benefit from experience and may even show negative transfer.
Moreover, we investigate model performance under two learning settings and find that instance-level rubrics and teacher-model feedback further facilitate model learning.
Importantly, we observe that current LLMs with stronger static abilities do not show a clear advantage in learning capability across all tasks, highlighting that EvaLearn evaluates a new dimension of model performance.
We hope EvaLearn provides a novel evaluation perspective for assessing LLM potential and understanding the gap between models and human capabilities, promoting the development of deeper and more dynamic evaluation approaches.
All datasets, the automatic evaluation framework, and the results studied in this paper are available at \url{https://github.com/ByteDance-Seed/EvaLearn}.

% We extensively benchmark nine frontier models and observe varied performance profiles: some models, such as DeepSeek-R1, start with moderate initial performance but exhibit strong learning ability; others, such as OpenAI o3-mini, demonstrate both high initial accuracy and robust learning; while some models struggle to benefit from experience and may even show negative transfer.
% Moreover, we investigate model performance under two learning settings and find that instance-level rubrics and teacher model feedback further facilitate model learning.

% to help us further understand the gap between models and human capabilities.

\end{abstract}

\section{Introduction}
\label{sec:introduction}

Large language models (LLMs) have advanced rapidly in recent years, driving remarkable progress in a wide range of applications \cite{qwen2.5,doubao1.5pro,seed2025seed,guo2025deepseek,o3,claude3.7}. 
Rigorous evaluation of these models is essential for understanding their current capabilities, identifying areas for improvement, and guiding the development of more advanced LLMs \cite{chang2024survey,laskar2024systematic,guo2023evaluating,gu2024survey,zheng2023judging}.

Although numerous benchmarks have been proposed to assess various aspects of model performance, the overwhelming majority rely on a parallel evaluation paradigm \cite{zellers2019hellaswag,srivastava2022beyond,lin2022truthfulqa,bisk2020piqa,liu2024alignbench,li2024crowdsourced,jiang-etal-2024-followbench,xia-etal-2024-fofo}. 
In this setting, models are tested on independent and identically distributed (i.i.d.) samples, and their overall performance is reported as aggregate metrics. 
However, such benchmarks primarily measure static abilities, while overlooking an equally important dimension: a model’s capability to learn and adapt within a specific task (i.e., \textbf{learning capability}), as well as the speed at which this learning occurs (i.e., \textbf{learning efficiency}). 
These dynamic learning abilities are fundamental indicators of human learning potential and intelligence \cite{bassett2011dynamic,deary2010neuroscience,wang2021meta,lake2017building}, yet remain largely unexplored in the evaluation of LLMs—mainly because the prevailing parallel evaluation paradigm is inherently unable to capture such learning dynamics.

To address this gap, we introduce EvaLearn, a challenging benchmark designed to systematically quantify the learning capability and efficiency of LLMs through a novel sequential evaluation paradigm.
Considering the scarcity of carefully categorized and challenging related problems in existing benchmarks, we construct 648 challenging problems from scratch.
These problems are organized into 182 sequences, with each sequence containing seven problems from the same task type and spanning six distinct task categories.
Models are required to sequentially solve each problem within a sequence, with each solution automatically evaluated using a combination of instance-level rubrics and an LLM-as-a-judge \cite{zheng2023judging,sirdeshmukh2025multichallenge} framework.
EvaLearn is designed to assess these two core aspects by testing whether models can leverage experience gained from solving previous problems to improve their performance on subsequent ones. 
Moreover, EvaLearn includes a suite of five metrics to comprehensively quantify the learning capability and efficiency of a model.
Importantly, these metrics are decoupled from the specific learning methods employed (i.e., how models utilize prior experience from earlier problems), highlighting the extensibility and flexibility of EvaLearn.

\begin{figure}[t]
\centering
\includegraphics[width=0.95\textwidth]{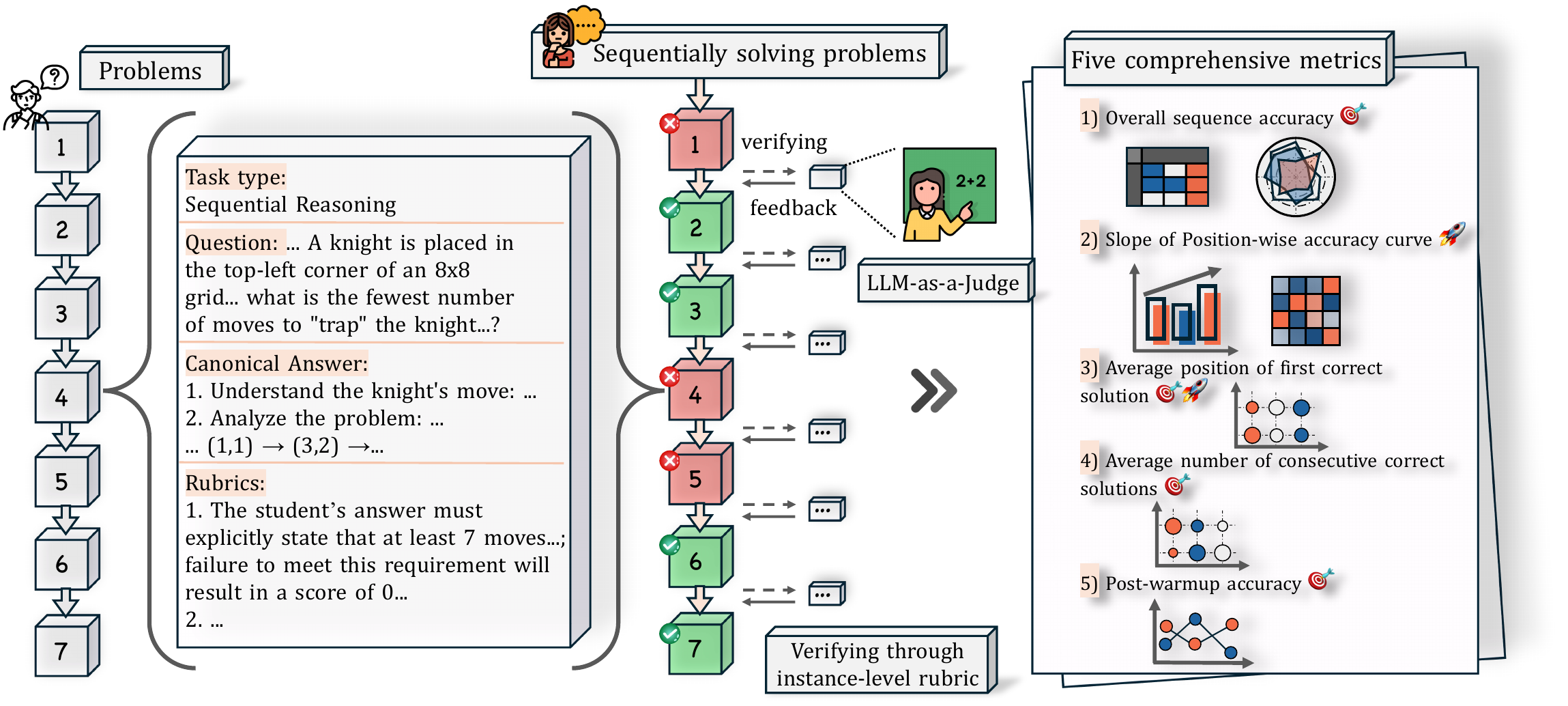}
\caption{
Overview of EvaLearn. 
Instead of parallel evaluation, EvaLearn requires models to solve problems sequentially, thereby systematically evaluating the LLM's learning ability and efficiency.
}
\label{fig:main}
  \vspace{-1.5em}
\end{figure}

We conducted a comprehensive study with nine frontier LLMs on EvaLearn.
We summarize several key findings:
\textbf{(a)} Models exhibit diverse learning capabilities across different task types.
They are generally more adept at leveraging prior experience to solve tasks involving mathematical and logical reasoning, while tasks such as summarization tend to rely more on knowledge acquired during pre-training and the model’s inherent abilities.
Moreover, learning efficiency also varies significantly across tasks.
\textbf{(b)} Thinking-based LLMs typically outperform non-thinking-based LLMs in both learning capability and learning efficiency.
They are better able to utilize experience to solve new problems and also show greater learning stability, being more likely to solve multiple problems consecutively within a sequence.
\textbf{(c)} Feedback learning, which enables models to solve problems with the help of feedback from previous solutions and rubric-based evaluations, significantly enhances both the learning capability and efficiency of models.
Moreover, it is often more effective than demonstration learning, which simply provides previous problems and canonical solutions as context.
\textbf{(d)} Indicators of learning capability and learning efficiency should be considered jointly. These metrics together provide a comprehensive assessment of a model’s learning potential.
Moreover, they are not strongly correlated with static model capability.
Even LLMs with higher static performance do not demonstrate a clear advantage in learning capability across all tasks.

More insightful findings are presented in Section~\ref{sec:exp} and Appendix~\ref{sec:addition-results}.
We hope EvaLearn offers a new perspective on assessing the learning potential of LLMs, which is a key indicator of human-like intelligence.
EvaLearn has a significant potential to foster a deeper understanding of model capabilities within the community, promote the development of more effective learning methods, and serve as a pioneering step toward more dynamic and realistic evaluation paradigms.

% to help us further understand the gap between models and human capabilities.

\section{EvaLearn}
\label{sec:evalearn}

In this section, we first describe dataset composition and task types.
We also provide an example to clearly illustrate the data structure.
Section~\ref{sec:evaluation} then presents the automated evaluation procedure based on instance-level rubrics.
Section~\ref{sec:metrics} introduces five metrics we designed to quantify the learning capability and efficiency of LLMs.
We also present the data annotation process in Appendix~\ref{sec:annotation}.

\subsection{Datasets}
\label{sec:datasets}

EvaLearn is designed to evaluate the LLM's learning capability and efficiency.
It contains 648 challenging problems, all carefully constructed from scratch and organized into 182 sequences.
Each sequence contains seven problems under the same task category. 
Problems within a sequence are interrelated and collectively challenge the model's learning potential. 
Figure~\ref{fig:main} provides an overview of EvaLearn.
Each problem within a sequence includes a question and a canonical answer. 
Since most of these challenging questions cannot be reliably evaluated using pre-defined rules, each problem is also accompanied by a human-written rubric that defines the criteria for assessing the correctness of model responses (details on the evaluation process are provided in Section~\ref{sec:evaluation}).

In contrast to most existing benchmarks that evaluate models in a parallel setting, EvaLearn requires models to solve problems sequentially within each sequence.
This sequential setup is specifically designed to evaluate a model's ability to accumulate experience and leverage feedback from previous problems, as reflected in its performance on subsequent problems within the same sequence.

\textbf{Task types.}
EvaLearn comprises six distinct task types, with each sequence belonging to one of these categories.
% For each task, we examine whether models can, like humans, improve their performance by learning from experience on solving the same task problems.
These task types include:
\textbf{(1) Summarization (Sum)} that evaluates whether models can improve the accuracy and coverage of summaries by leveraging prior experience;
\textbf{(2) Classification (Cla)} that assesses a model's ability to enhance its categorization skills from solving a series of classification problems;
\textbf{(3) Extraction (Ex)} that measures whether models can progressively improve the accuracy and completeness of key information extraction;
\textbf{(4) Logical Reasoning (LR)} that tests whether models can learn from previous errors and improve logic reasoning ability;
\textbf{(5) Mathematical Reasoning (MR)} that examines whether models can quickly master mathematical problem-solving methods by utilizing feedback from earlier problems;
\textbf{(6) Sequential Reasoning (SR)} that evaluates whether models can enhance their ability to solve sequence-based problems by learning from historical experience, including clarifying event steps and reasoning logic.

Table~\ref{tab:statistics} presents the statistics of EvaLearn.
Across all tasks, the average word counts per question is 315.45.
Solving these problems requires models to possess strong and diverse capabilities.
The results and analysis for various models on EvaLearn are provided in Section~\ref{sec:exp}.

\textbf{An example of sequence.}
We use an example sequence to showcase the relationships among problems within a sequence, as illustrated in Figure~\ref{fig:datasets-case1}, which involves solving a $4\times4$ sliding puzzle. 
In this case, the puzzle consists of 16 tiles, one of which is a blank space (denoted by `>'), while the remaining tiles are filled with different symbols. 
By moving the blank space and swapping it with adjacent tiles, the goal is to restore the puzzle from a given initial state to a target configuration through a series of logical moves. 
This task is designed to assess the model's logical reasoning ability, and all problems within this sequence are centered around the sliding puzzles and logical reasoning.
Figure~\ref{fig:datasets-case2} presents another problem from the same sequence, which differs mainly in grid size, the initial state of the puzzle, and symbol representation.

These problems are related and together assess the model’s capability for logical reasoning.
Humans are able to gain experience and improve their performance by solving a series of such problems~\cite{LANGLEY2009141,tenisonPhasesLearningHow2016}. 
By requiring models to solve these problems sequentially within a sequence, EvaLearn evaluates whether models can similarly learn from experience, thereby measuring their learning efficiency.

\subsection{Automatic Evaluation}
\label{sec:evaluation}

Most of the challenging problems in EvaLearn cannot be reliably evaluated using rule-based verifiers, as many questions may have answers that are difficult to verify with pre-defined rules or may allow for multiple correct answers.
To address this, we employ an instance-level rubric combined with LLM-as-a-judge \cite{gu2024survey} to assess whether model outputs satisfy the corresponding rubrics. 
The evaluation prompt is shown in Figure~\ref{fig:prompt-judge}, which includes three demonstrations to ensure that the judging model follows the instructions.
In all experiments, we use GPT-4o as the judging model.

\begin{table}[t]
  \centering
  \footnotesize
  \caption{Statistics of EvaLearn.}
  \resizebox{0.86\textwidth}{!}{
    \begin{tabularx}{\textwidth}{l|ccccc}
    \toprule
    \textbf{Task Type} & \textbf{\# Problems} & \textbf{\# Sequences} & \textbf{\makecell{Problem\\reuse rate}} & \textbf{\makecell{Average words\\per question}} & \textbf{\makecell{Average words\\per canonical answer}} \\
    \midrule
    Summarization & 60    & 16    & 1.87 & 959.97  & 220.79 \\
    Classification & 48    & 13    & 1.90 & 203.18 & 149.31 \\
    Extraction & 60    & 17    & 1.98  & 674.89 & 78.60 \\
    Logical Reasoning & 360   & 102   & 1.98  & 227.49 & 81.98 \\
    Mathematical Reasoning & 60    & 17    & 1.98 & 137.03 & 336.76 \\
    Sequential Reasoning & 60    & 17    & 1.98 & 141.38 & 334.30 \\
    \midrule
    Overall & 648   & 182   & 1.966 & 315.45 & 146.05 \\
    \bottomrule
    \end{tabularx}%
    }
  \label{tab:statistics}%
  \vspace{-1.5em}
\end{table}%

We conduct an additional experiment to validate the effectiveness of our automatic evaluation process. 
Specifically, for each problem in EvaLearn, we randomly select three models from a model list (as shown in Table~\ref{tab:model-list}) to generate answers, which are then evaluated using our evaluation framework. 
Human annotators further verify whether the judging model can successfully assess the correctness of the model outputs through the instance-level rubric. 
Results show that the evaluation accuracy exceeds 95\% for all tasks, demonstrating the high reliability of our evaluation framework that combines instance-level rubrics with LLM-as-a-judge, consistent with findings from previous studies~\cite{sirdeshmukh2025multichallenge,yao2024tau}.

\subsection{Evaluation Metrics}
\label{sec:metrics}

In EvaLearn, models are required to sequentially solve all problems within a sequence. 
All evaluation metrics are computed based on the model’s correctness across these problems.
Notably, the choice of learning method, such as learning from demonstrations (i.e., in-context learning)~\cite{brown2020language,min-etal-2022-rethinking} or learning from feedback~\cite{shinn2023reflexion}, is decoupled from the evaluation metrics.
This design ensures that EvaLearn remains extensible and flexible, supporting a wide range of learning strategies.

Let $N=182$ denote the number of sequences and $M=7$ the number of problems in each sequence in EvaLearn.  
Let $y_{n,m} \in \{0,1\}$ indicate whether the $m$-th problem in the $n$-th sequence is answered correctly ($1$ for correct, $0$ for incorrect).
We define the following five metrics to comprehensively evaluate the learning potential of models in sequential problem-solving:

\textbf{(1) Overall sequence accuracy} ($\text{Acc}$).  
This metric reflects the model’s overall performance in sequential problem-solving and serves as an indirect indicator of its learning capability.  
It is computed as the average accuracy across all problems and all sequences:
\begin{equation}
\small
    \text{Acc} = \frac{1}{N M} \sum_{n=1}^N \sum_{m=1}^M y_{n,m}
\end{equation}
Higher values indicate better overall performance.
We further compute the \textbf{position-wise accuracy curve}, i.e., the accuracy at each problem position $m$ across all sequences:
\begin{equation}
\small
    \text{Acc}_m = \frac{1}{N} \sum_{n=1}^N y_{n,m}
\end{equation}

Based on this curve, we define a helpful metric, i.e., the slope of the fitted accuracy curve, to capture dynamic learning behaviors.

\textbf{(2) Slope of fitted accuracy curve} ($k$):  
This metric measures the model’s learning speed across all sequences by fitting a straight line to the position-wise accuracy curve using least squares regression \cite{hastie2009elements}.  
It quantifies how quickly the model’s accuracy improves as it progresses through sequences.  

Specifically, let $\text{Acc}_m$ denote the average accuracy at position $m$ across all sequences. The fitted line is given by $y = kx + b$, where $x$ is the problem position and $k$ is the estimated slope obtained by minimizing the sum of squared errors:
\begin{equation}
\small
    k = \arg\min_{k, b} \sum_{m=1}^{M} \left( \text{Acc}_m - (k m + b) \right)^2.
\end{equation}
A higher $k$ indicates greater learning efficiency and a faster rate of improvement.

\textbf{(3) Average position of first correct solution} ($P_{\text{first}}$).  
This metric measures the average position in the sequence where the model achieves its first correct solution, indicating its initial ability, learning capability, and efficiency.  
For each sequence $n$, let $p_n$ denote the position of the first correct solution (or $M+1$ if none is correct). The metric is calculated as:
\begin{equation}
\small
    P_{\text{first}} = \frac{1}{N} \sum_{n=1}^N p_n
\end{equation}
Lower values indicate that the model can achieve its first correct solution earlier in the sequence.

\textbf{(4) Average offset of first learned correct solution} ($P_{\text{offset}}$).  
This metric measures how quickly a model begins to learn within a sequence while discounting pre-existing knowledge.  
For each sequence $n$, let $i_0^{(n)}$ be the position of the first problem answered incorrectly under zero-shot (parallel) solving; if all problems are correct in zero-shot, set $i_0^{(n)} = M{+}1$. Let $A^{\text{zero}}_{n,i}, A^{\text{feedback}}_{n,i} \in \{0,1\}$ denote correctness at position $i$ under zero-shot and under the sequential solving setting, respectively. We define the per-sequence offset to the first learned correct solution as:
\begin{equation}
\small
    d_n \,=\, \min \left\{\, i - i_0^{(n)} \;\middle|\; i \ge i_0^{(n)},\; A^{\text{feedback}}_{n,i} = 1,\; A^{\text{zero}}_{n,i} = 0 \,\right\},
\end{equation}
with the convention that $d_n = M{+}1 - i_0^{(n)}$ if no such $i$ exists, and $d_n = 0$ when $i_0^{(n)} = M{+}1$. The metric is computed by averaging over sequences:
\begin{equation}
\small
    P_{\text{offset}} = \frac{1}{N} \sum_{n=1}^N d_n.
\end{equation}
Lower values indicate that the model begins to learn earlier after excluding problems solved due to pre-existing knowledge.

\textbf{(5) Average number of consecutive correct solutions} ($N_{\text{consec}}$). 
This metric reflects the model’s capability to leverage experience to consistently solve problems within a sequence, resulting in fewer errors. 
It indirectly indicates how effectively and stably the model learns.
For each sequence $n$, we compute the longest run of consecutive correct solutions, then average across all sequences:
\begin{equation}
\small
    N_{\text{consec}} = \frac{1}{N} \sum_{n=1}^N \max_{\substack{1 \leq a \leq b \leq M}} \left\{ b-a+1 : y_{n,a}=y_{n,a+1}=\dots=y_{n,b}=1 \right\}
\end{equation}
Higher values indicate greater consistency, while lower values suggest intermittent errors and less stable learning.

\textbf{(6) Post-warmup accuracy} ($\text{Acc}_{\text{pw}}\text{-K}$).  
This metric reflects model performance after an initial ``warmup'' phase, i.e., after some experience has been accumulated.  
For each sequence, we exclude the first $K$ problems and compute the average accuracy on the remaining problems:
\begin{equation}
\small
    \text{Acc}_{\text{pw}}\text{-K} = \frac{1}{N (M-K)} \sum_{n=1}^N \sum_{m=K+1}^{M} y_{n,m}
\end{equation}
Higher values indicate better adaptation and learning after the warmup.

These metrics provide a comprehensive and nuanced evaluation of dynamic learning behaviors of models in sequential problem-solving, fully capturing learning capability and efficiency. 
In the following sections, we use these metrics to thoroughly analyze the learning potential of various LLMs.

\section{Benchmarking LLMs with EvaLearn}
\label{sec:exp}

\begin{table}[t]
    \centering
    \footnotesize
    \caption{Comparison of accuracy between zero-shot (parallel solving) and feedback Learning (sequential solving).
    Values in parentheses indicate the difference between the two methods.
    Full task names for the abbreviations can be found in Section~\ref{sec:datasets}. Blue-colored cells indicate the best performance for each task category under few-shot and feedback learning settings, respectively. All values are shown as \%.
    }
    \resizebox{1\textwidth}{!}{
    \begin{tabular}{l|c|cccccc|c}
    \toprule
    \textbf{Model} & \textbf{Paradigm} & \textbf{Sum} & \textbf{Cla} & \textbf{Ex} & \textbf{LR} & \textbf{MR} & \textbf{SR}  & \textbf{Overall} \\
    \midrule
    \rowcolor[rgb]{ .71,  .773,  .945} \multicolumn{9}{l}{\textbf{Non-thinking-based}} \\
    \multirow{2}{*}{DeepSeek-V3} & Zero-shot & 81.7 & 72.9 & 45.0 & 25.6 & 71.7 & 60.0 & 43.5 \\
    & Feedback Learning & 76.8 (-4.9) & 74.7 (+1.8) & 36.1 (-8.9) & 24.9 (-0.7) & 74.8 (+3.1) & 54.6 (-5.4) & 41.5 (-2.0) \\
    \midrule
    \multirow{2}{*}{Claude-3.7-Sonnet} & Zero-shot & 76.7 & 64.6 & 48.3 & 8.1 & 48.3 & 33.3 & 28.4 \\
    & Feedback Learning & 75.0 (-1.7) & 75.8 (\textbf{+11.2}) & 46.2 (-2.1) & 15.5 (\textbf{+7.4}) & 63.9 (+15.6) & 49.6 (\textbf{+16.3}) & 35.6 (\textbf{+7.2}) \\
    \midrule
    \multirow{2}{*}{GPT-4o} & Zero-shot & 81.7 & 70.8 & 45.0 & 15.6 & 43.3 & 31.7 & 32.6 \\
    & Feedback Learning & 78.6 (-3.1) & 73.6 (+2.8) & 48.7 (\textbf{+3.7}) & 12.9 (-2.7) & 61.3 (\textbf{+18.0}) & 32.2 (+0.5) & 32.7 (+0.1) \\
    \midrule
    \multirow{2}{*}{Doubao-1.5-Pro} & Zero-shot & 71.7 & 70.8 & 45.0 & 10.3 & 61.7 & 41.7 & 31.3 \\
    & Feedback Learning & 71.4 (-0.3) & 68.1 (-2.7) & 42.9 (-2.1) & 10.8 (+0.5) & 71.4 (+9.7) & 36.1 (-5.6) & 31.2 (-0.1) \\
    \midrule
    \multirow{2}{*}{Qwen2.5-32b-Instruct} & Zero-shot & 66.7 & 60.4 & 31.7 & 6.9 & 46.7 & 21.7 & 23.8 \\
    & Feedback Learning & 59.8 (-6.9) & 62.6 (+2.2) & 30.3 (-1.4) & 9.8 (+2.9) & 49.6 (+2.9) & 30.3 (+8.6) & 25.5 (+1.7) \\
    \midrule
    \rowcolor[rgb]{ .71,  .773,  .945} \multicolumn{9}{l}{\textbf{Thinking-based}} \\
    \multirow{2}{*}{OpenAI-o3-mini} & Zero-shot & 65.0 & 64.6 & 48.3 & 45.8 & 73.3 & 73.3 & 54.3 \\
    & Feedback Learning & 75.9 (\textbf{+10.9}) & 73.6 (+9.0) & 47.1 (-1.2) & \cellcolor{myblue!50}59.9 (\textbf{+14.1}) & 80.7 (+7.4) & 78.2 (+4.9) & 64.8 (\textbf{+10.5}) \\
    \midrule
    \multirow{2}{*}{Doubao-1.5-Thinking-Pro} & Zero-shot & 85.0 & 79.2 & \cellcolor{myblue!50}55.0 & 42.5 & 85.0 & 73.3 & 57.1 \\
    & Feedback Learning & 82.1 (-2.9) & 70.3 (-8.9) & \cellcolor{myblue!50}52.9 (-2.1) & 39.4 (-3.1) & 77.3 (-7.7) & 55.5 (-17.8) & 51.6 (-5.5) \\
    \midrule
    \multirow{2}{*}{DeepSeek-R1} & Zero-shot & \cellcolor{myblue!50}86.7 & \cellcolor{myblue!50}89.6 & 48.3 & 41.7 & 78.3 & 66.7 & 55.7 \\
    & Feedback Learning & \cellcolor{myblue!50}89.3 (+2.6) & 79.1 (-10.5) & 48.7 (+0.4) & 29.6 (-12.1) & 74.8 (-3.5) & 51.3 (-15.4) & 46.4 (-9.3) \\
    \midrule
    \multirow{2}{*}{Claude-3.7-Sonnet-Thinking} & Zero-shot & \cellcolor{myblue!50}86.7 & 66.7 & 43.3 & 12.5 & 46.7 & 31.7 & 31.2 \\
    & Feedback Learning & 78.6 (-8.1) & \cellcolor{myblue!50}80.2 (\textbf{+13.5}) & 48.7 (\textbf{+5.4}) & 18.8 (+6.3) & 58.0 (\textbf{+11.3}) & 46.2 (\textbf{+14.5}) & 37.4 (+6.2) \\
    \midrule
    \multirow{2}{*}{Gemini-2.5-Pro} & Zero-shot & 85.0 & 72.9 & \cellcolor{myblue!50}55.0 & \cellcolor{myblue!50}62.5 & \cellcolor{myblue!50}90.0 & \cellcolor{myblue!50}76.7 & \cellcolor{myblue!50}68.5 \\
    & Feedback Learning & \cellcolor{myblue!50}89.3 (+4.3) & 75.8 (+2.9) & 52.1 (-2.9) & \cellcolor{myblue!50}59.9 (-2.6) & \cellcolor{myblue!50}85.7 (-4.3) & \cellcolor{myblue!50}79.8 (+3.1) & \cellcolor{myblue!50}67.2 (-1.3) \\
    \midrule
    \multirow{2}{*}{Gemini-2.5-Flash} & Zero-shot & 81.7 & 66.7 & 38.3 & 16.1 & 70.0 & 51.7 & 36.3 \\
    & Feedback Learning & 67.9 (-13.8) & 59.3 (-7.4) & 42.0 (+3.7) & 18.3 (+2.2) & 81.5 (+11.5) & 58.8 (+7.1) & 37.5 (+1.2) \\
    \bottomrule
    \end{tabular}%
    }
    \label{tab:RQ-1}
    \vspace{-1.5em}
\end{table}

In this section, we compare two problem-solving paradigms, including parallel solving and sequential solving, to investigate the learning capability and efficiency of LLMs.

\subsection{Setup}
\label{sec:setup}

\textbf{Parallel Solving.}
This paradigm includes two settings:
\textbf{(I) Zero-shot}: Models solve each problem independently, without access to any experience from previous problems. 
This setting aligns with the evaluation approach used in most existing benchmarks, assessing a model's inherent ability to solve challenging problems without any learning opportunity. 
The system prompt is shown in Figure~\ref{fig:Prompt for Zero-shot}.
\textbf{(II) Few-shot}: For each problem, we provide three demonstrations from the same task as examples (i.e., 3-shot), offering models guidance on output format and problem-solving approach. 
The demonstrations are identical for all problems within each task type. 
The system prompt is shown in Figure~\ref{fig:Prompt for Few-shot}.

\textbf{Sequential Solving.}
To investigate models' ability to learn from experience, we utilize two sequential learning paradigms:
\textbf{(I) Demonstration learning}: Models are provided with all previous problems and their corresponding canonical answers from the same sequence before solving the current problem, similar to in-context learning.
The system prompt is shown in Figure~\ref{fig:Prompt for Demonstration Learning}.
\textbf{(II) Feedback learning}: When solving the current problem, models receive as context all previous problems, their solutions, and detailed feedback on their own prior solutions, as assessed by a judge using instance-level rubrics.
This setting evaluates whether models can leverage their previous experience to improve on subsequent problems.
The system prompt is shown in Figure~\ref{fig:Prompt for Feedback Learning}.

In all experiments, we evaluate nine frontier LLMs, including both thinking-based and non-thinking-based models: Claude-3.7-Sonnet \cite{claude3.7}, Claude-3.7-Sonnet-Thinking \cite{claude3.7}, DeepSeek-R1 \cite{guo2025deepseek}, DeepSeek-V3 \cite{liu2024deepseek}, Doubao-1.5-Pro \cite{doubao1.5pro}, Doubao-1.5-Thinking-Pro \cite{seed2025seed}, OpenAI-o3-mini \cite{o3}, GPT-4o \cite{gpt4}, and Qwen2.5-32b-Instruct \cite{qwen2.5}.
See Appendix~\ref{sec:open-source-models} for details on these models.

\subsection{RQ \Romannum{1}: Can LLMs learn a task by engaging with a sequence of problems?}

\begin{findingBox}{1}{
LLMs exhibit varying abilities to learn from problem sequences, with differences observed across both models and task types. Moreover, most models also demonstrate better performance after a warm-up phase.
}
\end{findingBox}

Table~\ref{tab:RQ-1} summarizes the differences in overall accuracy between the feedback learning and zero-shot paradigms.
We observe that five models benefit from sequence learning, while four models experience slight declines, indicating that some models can effectively leverage prior experience to solve problems within the same task.
Thinking-based models, in particular, exhibit more pronounced performance shifts, with OpenAI-o3-mini achieving the highest overall improvement (+10.5\%).

Moreover, most LLMs show improvement on mathematical reasoning and classification tasks when learning from experience.
For example, GPT-4o and Claude-3.7-Sonnet-Thinking achieve gains of 18.0\% and 13.5\% in mathematical reasoning and classification, respectively.
This improvement likely results from these tasks having clear solution steps, enabling feedback to pinpoint reasoning errors and help models effectively learn specific problem-solving strategies.

In contrast, most models experience performance declines on the summarization task; for instance, 7 out of 9 models perform worse after feedback in summarization.
This may be because the summarization task rely more heavily on knowledge and instruction-following abilities acquired during pre-training, and additional experience may sometimes interfere with the model’s ability to solve the current problem.

\begin{figure}[t]
    \centering
    \begin{minipage}{0.45\textwidth}
        \centering
        \includegraphics[width=\textwidth]{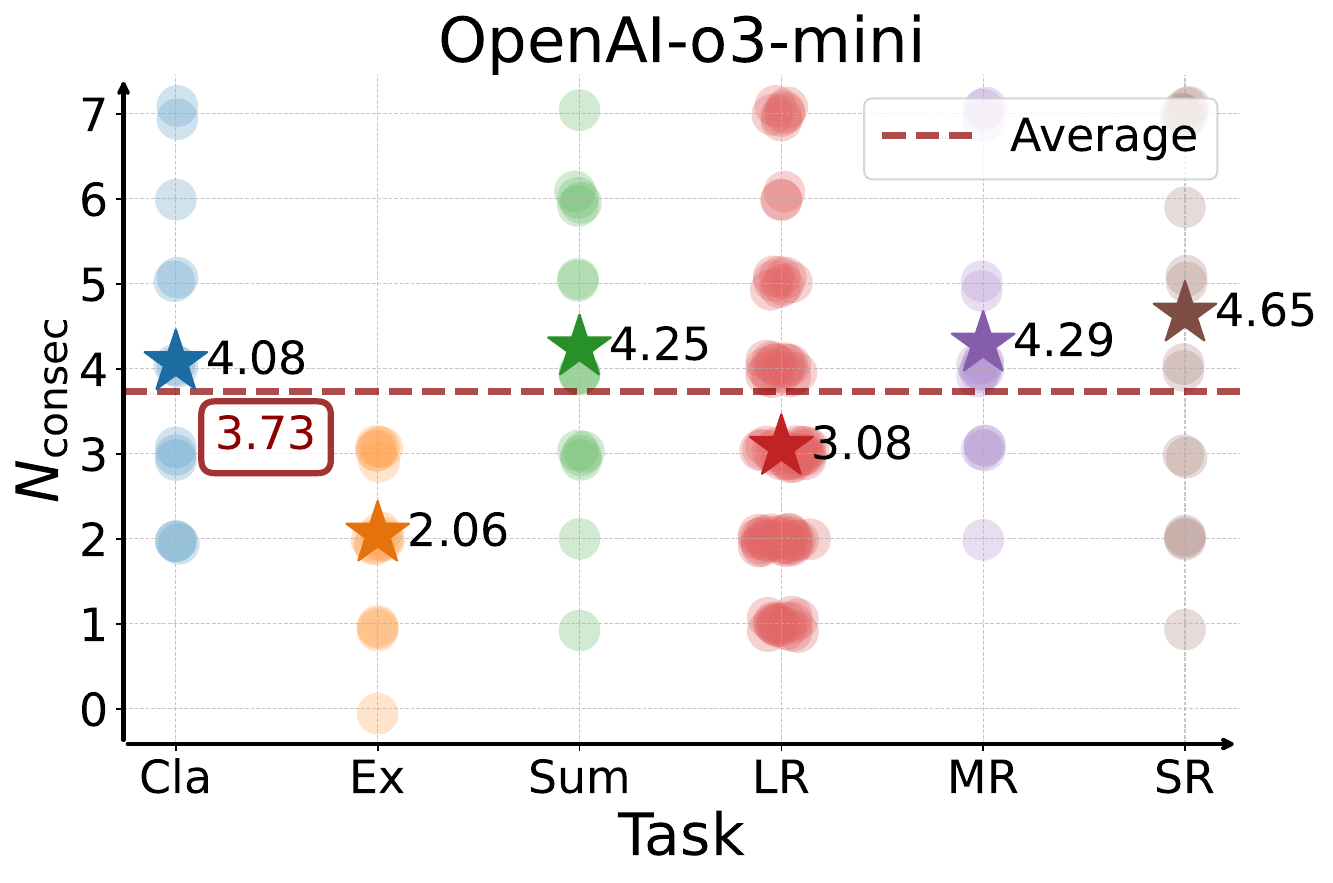}
    \end{minipage}
    \hfill
    \begin{minipage}{0.45\textwidth}
        \centering
        \includegraphics[width=\textwidth]{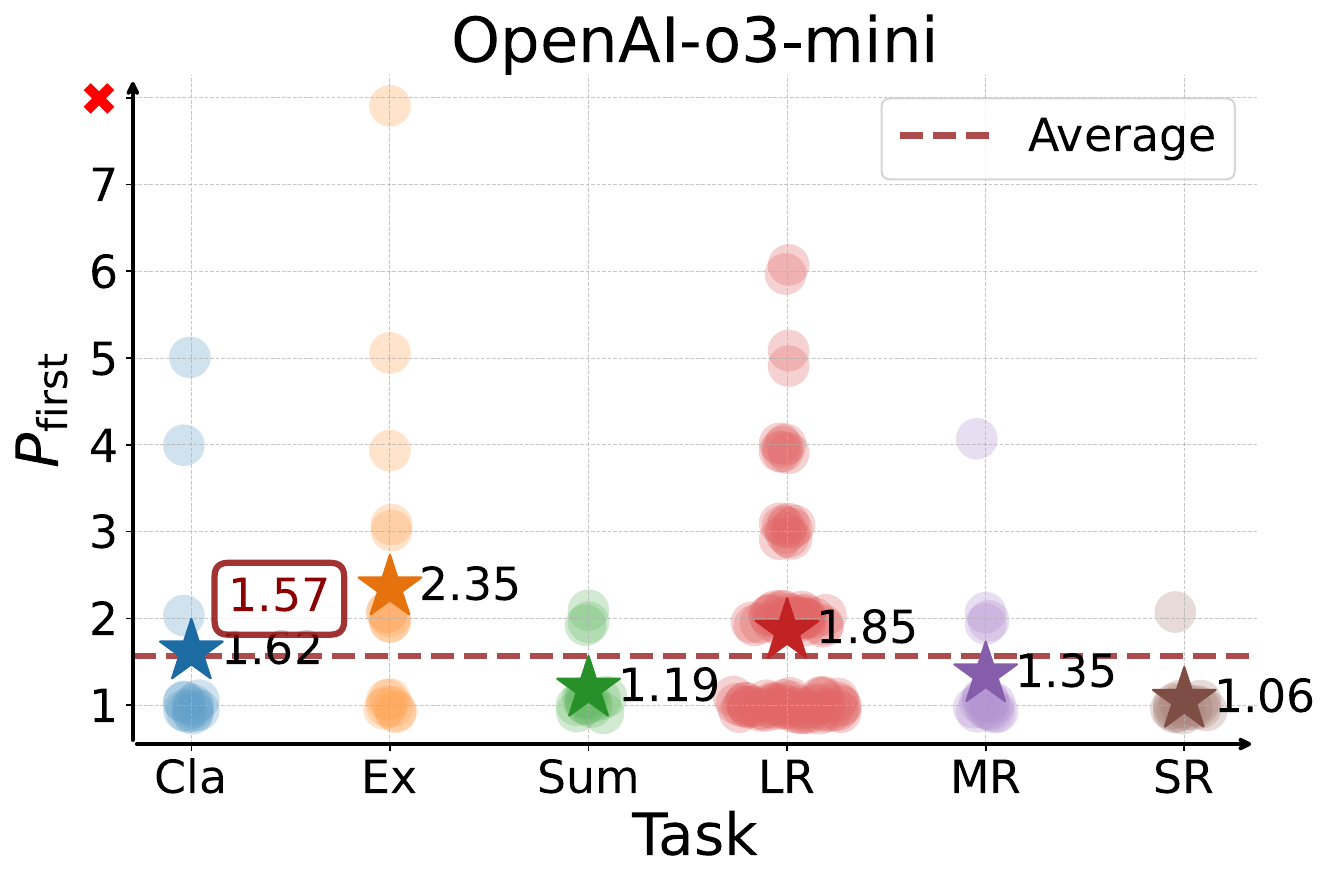}
    \end{minipage}
    \caption{
    \textbf{(Left)} Average number of consecutive correct solutions ($N_{\mathrm{consec}}$).
    \textbf{(Right)} Average position of the first correct solution ($P_\text{first}$).
    Results are shown for OpenAI-o3-mini, with each node representing a sequence.
    }
    \label{fig:main-consec}
    \vspace{-1.5em}
\end{figure}

We also analyze models' Post-Warmup Accuracy ($\mathrm{Acc}_{\mathrm{pw}}\text{-K}$), with results for Claude-3.7-Sonnet-Thinking shown in the left side of Figure~\ref{fig:main-curve-and-pwk} and additional models presented in Appendix~\ref{app:metric-5-pwk}.  
We observe that most models achieve higher accuracy in the later stages of the sequence, particularly among thinking-based models.  
This further demonstrates that models can use early problems as practice, leveraging the experience gained to solve subsequent, related problems more effectively.

\begin{findingBox}{2}{
Learning stability varies significantly across different tasks and models.
For certain tasks, such as summarization, current models are more adept at leveraging their inherent knowledge to solve problems rather than drawing on experience gained from previous problems.
}\end{findingBox}

We further investigate the learning stability of LLMs by analyzing the average number of consecutive correct solutions ($N_\text{consec}$).  
The left side of Figure~\ref{fig:main-consec} presents the results for o3-mini, with additional results for other models provided in Appendix~\ref{app:metric-4-consec}.  
Each colored node in the figure represents a sequence in the dataset, and nodes positioned higher indicate that the model solves more problems consecutively within that sequence.

From the results, we first observe that the average number of consecutive correct solutions varies notably across tasks.  
For most models, logical reasoning tasks present greater challenges for maintaining long streaks of correct solutions.  
For example, Claude-3.7-Sonnet and GPT-4o fail entirely or manage to solve only one problem consecutively in over half of the sequences.
However, there are substantial differences between thinking-based and non-thinking-based models.  
For instance, while Doubao-1.5-Pro struggles to maintain consecutive correct solutions in logical reasoning, Doubao-1.5-Thinking-Pro exhibits much greater learning stability.  
Similar patterns appear in other model pairs, such as DeepSeek-V3 vs DeepSeek-R1 and GPT-4o vs o3-mini, where the thinking-based variants generally achieve higher average numbers of consecutive correct solutions.  
For example, o3-mini achieves an average maximum streak of 3.42 consecutive correct solutions per sequence, compared to only 2.58 for GPT-4o.
Combining these results with Table~\ref{tab:RQ-1}, we conclude that thinking-based models not only benefit more from prior experience to improve performance, but also maintain higher stability and are more likely to solve multiple related problems in succession.

Interestingly, in the summarization task, models tend to have relatively high $N_\text{consec}$ values.  
However, these same models show a marked decline in $\mathrm{Acc}_{\mathrm{pw}\text{-}K}$, indicating that their performance on later problems in the sequence is worse than on earlier ones.  
This suggests that, for such tasks, the capability to solve multiple problems consecutively may rely more on the model’s inherent knowledge and static ability, rather than on learning from experience within the sequence.  
This observation further supports the conclusion in Findings 1.

Overall, these metrics allow us to comprehensively assess the learning capability and stability of LLMs, providing deeper insight into their potential to learn from experience within a task.

\begin{findingBox}{3}{
Learning capability provides a new perspective for evaluating models, independent of their static performance, and reveals their underlying learning potential.
}\end{findingBox}

Results in Table~\ref{tab:RQ-1} show that strong performance in parallel solving does not necessarily imply strong learning capability.
For example, although DeepSeek-R1 outperforms Claude-3.7-Sonnet in the parallel setting, it experiences a 9\% drop in the sequential solving paradigm, while Claude-3.7-Sonnet achieves a 7.2\% gain.
In the logic reasoning task, Claude-3.7-Sonnet-Thinking does not outperform DeepSeek-R1 or Doubao-1.5-Thinking-Pro in the zero-shot setting, yet it demonstrates stronger learning capability by substantially improving its performance through experience.
o3-mini, on the other hand, exhibits the best performance in this task under both zero-shot and feedback learning settings, showing remarkable learning capability by leveraging feedback to achieve a 14.1 percentage point improvement.
Another example is DeepSeek-R1 in the math reasoning task. Although DeepSeek-R1 achieves higher zero-shot performance than o3-mini, it fails to learn from experience as effectively as o3-mini and instead suffers a decline in performance.

% \begin{figure}[t]
%     \centering
%     \begin{subfigure}{0.45\textwidth}
%         \centering
%         \includegraphics[width=\textwidth]{figures/RQ1acc_pw_k/Learning/Claude3.7Sonnet-Thinking_acc_pw_k.pdf}
%     \end{subfigure}
%     \centering
%     \begin{subfigure}{0.45\textwidth}
%         \includegraphics[width=\textwidth]{figures/RQ2position_curve/Learning/DeepSeek-R1accuracy_curve.pdf}
%     \end{subfigure}
%     % \vspace{-2em}
% \caption{
% \textbf{(Left)} Post-warmup accuracy ($\mathrm{Acc}_{\mathrm{pw}}\text{-K}$) results of Claude-3.7-Sonnet-Thinking.
% \textbf{(Right)} Position-wise accuracy curve and its slope $k$ of DeepSeek-R1.
% }
% \label{fig:main-curve-and-pwk}
% \label{fig:main-consec}
% \end{figure}

\begin{figure}[t]
    \centering
    \begin{minipage}{0.45\textwidth}
        \centering
        \includegraphics[width=\textwidth]{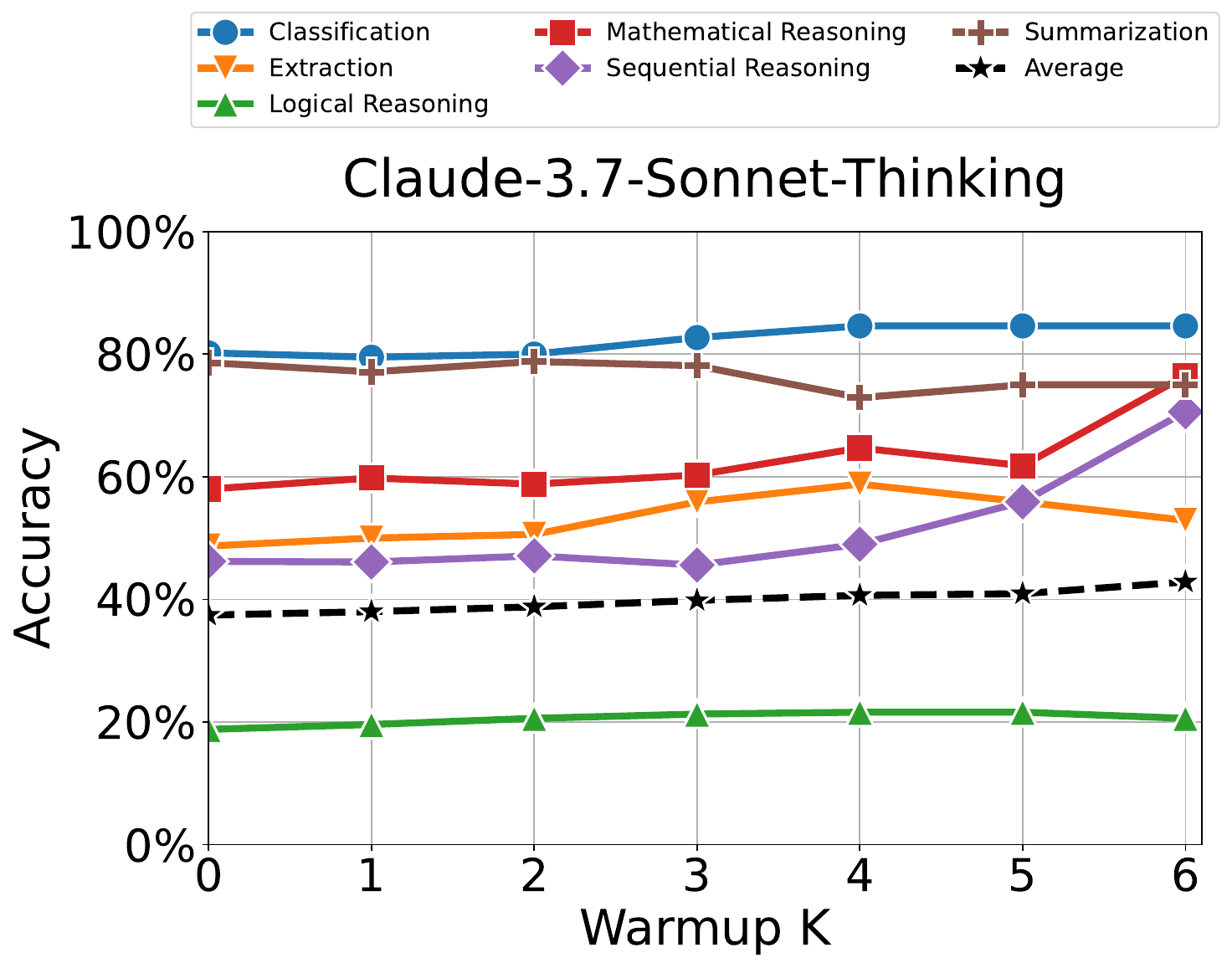}
    \end{minipage}
    \hfill
    \begin{minipage}{0.45\textwidth}
        \centering
        \includegraphics[width=\textwidth]{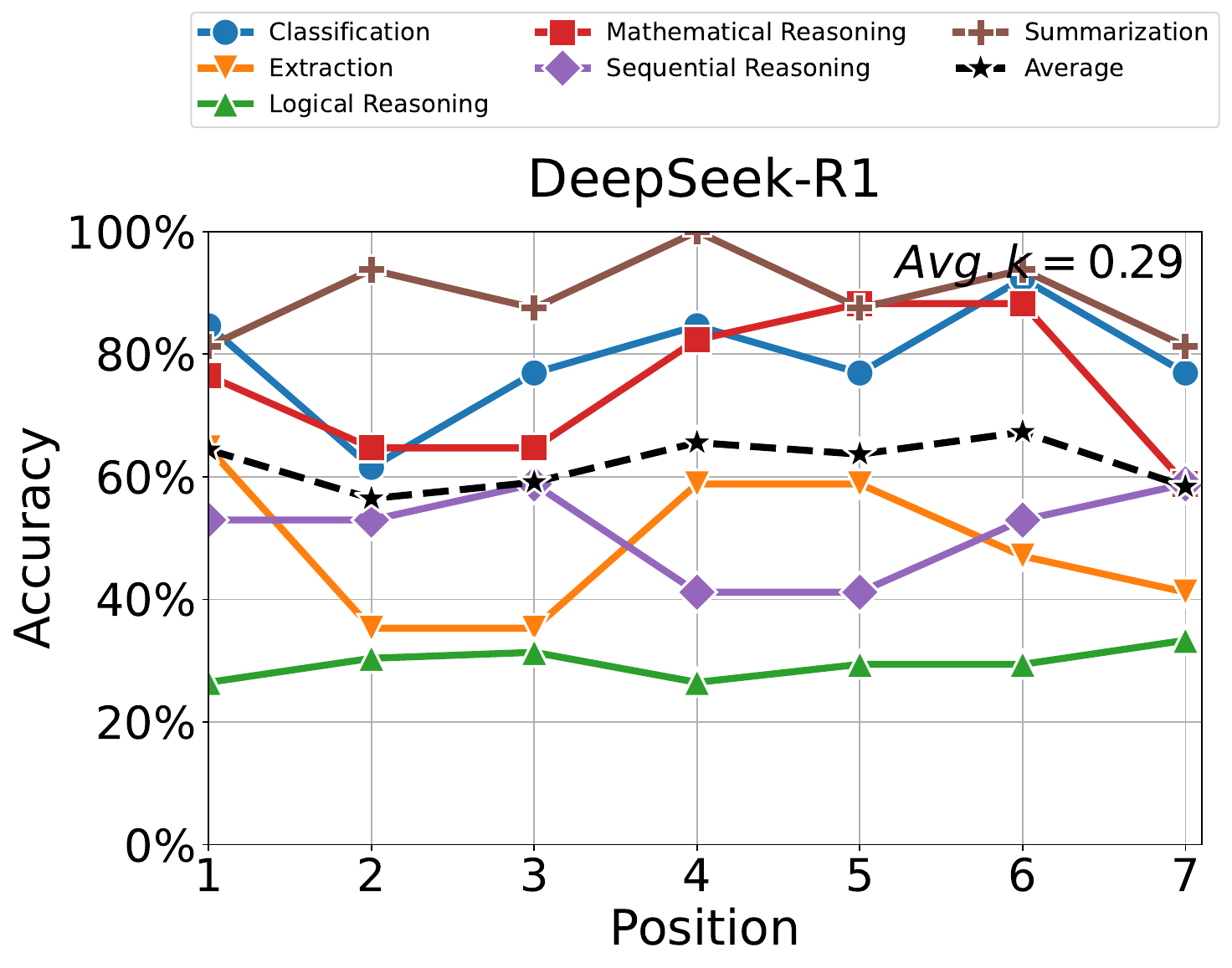}
    \end{minipage}
    \caption{
    \textbf{(Left)} Post-warmup accuracy ($\mathrm{Acc}_{\mathrm{pw}}\text{-K}$) results of Claude-3.7-Sonnet-Thinking.
    \textbf{(Right)} Position-wise accuracy curve and its slope $k$ of DeepSeek-R1.
    }
    \label{fig:main-curve-and-pwk}
    \vspace{-1.5em}
\end{figure}

Moreover, models with similar static abilities may exhibit markedly different learning capabilities.
For instance, Claude-3.7-Sonnet-Thinking and GPT-4o both achieve 31.7\% accuracy in the sequential reasoning task under the zero-shot paradigm, yet their learning capabilities differ significantly: the former improves by 14.5 percentage points, while the latter only improves by 0.5 points.
However, GPT-4o demonstrates stronger learning capability than Claude-3.7-Sonnet-Thinking in the math reasoning task.
These results indicate that learning capability also varies across different tasks.

Taken together, these observations suggest that learning capability is an important and distinct metric for model evaluation, independent of static performance and varying across different tasks.
This dimension of model assessment deserves further attention and systematic investigation by the research community.
In Research Question III (§~\ref{sec:rq3}), we further analyze model performance under various learning paradigms, providing additional evidence that learning capability is not solely determined by static ability.

\subsection{RQ \Romannum{2}: How efficient are LLMs at learning from a sequence of problems?}

\begin{findingBox}{4}{
Learning efficiency differs markedly across models and task types.  
On average, most non-thinking-based models improve more rapidly with experience, while thinking-based models tend to achieve more stable gains.
}
\end{findingBox}

The right side of Figure~\ref{fig:main-curve-and-pwk} presents the position-wise accuracy curve for DeepSeek-R1, with corresponding curves for other models shown in Appendix~\ref{app:acc-per-position}.  
The heatmap in Figure~\ref{fig:feedback heatmap} summarizes the slope ($k$) of these curves, directly indicating model learning efficiency.

We observe substantial differences in learning efficiency across models.  
For instance, Claude-3.7-Sonnet achieves the highest overall learning efficiency, with $k=2.08$.  
On average, most non-thinking-based models exhibit steeper slopes in their position-wise accuracy curves compared to thinking-based models, as shown in Figure~\ref{fig:feedback heatmap} (e.g., Claude-3.7-Sonnet, DeepSeek-V3, and GPT-4o).  
This may be because non-thinking-based models often start from a lower baseline, allowing them to quickly capitalize on ``low-hanging fruit'' as they accumulate experience.

In contrast, learning in thinking-based models is more stable.  
For example, as shown in Figure~\ref{fig:Feedback-acc-per-position}, o3-mini’s position-wise accuracy curve shows a steady upward trend, even when its initial accuracy is already high.  
Non-thinking-based models often display larger fluctuations between adjacent positions, such as GPT-4o, DeepSeek-V3, and Doubao-1.5-Pro.  
This suggests that thinking-based models can better leverage information from previous problems, including feedback from the judging model, to reason more effectively about subsequent solutions, resulting in more stable performance gains.
Moreover, most models display positive learning efficiency in math reasoning tasks.  
However, sequential reasoning tasks remain particularly challenging, with widespread performance declines; only Claude-3.7-Sonnet-Thinking shows meaningful positive learning on these sequences ($k=1.89$).  
This indicates that even in reasoning tasks with clear solution paths and feedback, learning efficiency can vary significantly depending on the specific task, highlighting that it depends on multiple factors.

Overall, these results suggest that static model capability, task type, and reasoning approach all influence learning efficiency.  
Non-thinking-based models tend to improve performance more rapidly through experience, while thinking-based models achieve more stable improvements.  
Case studies in Appendix~\ref{sec:case-study} further illustrate how thinking-based models can utilize prior experience to solve new problems.

\subsection{RQ \Romannum{3}: Do different learning methods lead to differences in performance?}
\label{sec:rq3}

\begin{findingBox}{5}{
Different solving approaches significantly affect model performance.
Models can acquire experience from demonstrations, and feedback further enhances their learning.
Moreover, learning capability is not strongly correlated with a model’s inherent static ability.
}\end{findingBox}

Table~\ref{tab:RQ-3} and Figure~\ref{fig:radar-map} present model performance across four solving methods.  
We observe that, for most models, demonstration learning in the sequential setting generally yields better results than few-shot parallel solving.  
For instance, Claude-3.7-Sonnet-Thinking achieves consistent performance improvements across five tasks compared to the few-shot setting.  
The key distinction is that demonstration learning allows models to access all previous problems and canonical solutions within the sequence, enabling them to learn from prior experience.

Furthermore, Figures~\ref{fig:feedback-acc-per-k}, \ref{fig:demonstration-acc-per-k}, and~\ref{fig:feedback-demonstration heatmap} compare feedback learning and demonstration learning in terms of post-warmup accuracy ($\mathrm{Acc}_{\mathrm{pw}\text{-}K}$) and the slope $k$, with additional results provided in Appendix~\ref{sec:addition-results}.  
Results show that, for most models, feedback learning achieves higher average overall accuracy and greater learning efficiency than demonstration learning. 
For example, DeepSeek-V3 and Doubao-1.5-Thinking-Pro show the ability to learn from feedback provided by the judging model and apply this experience to subsequent problems across five tasks.

Notably, we find that each LLM exhibits strong learning capability on certain tasks, but no model can consistently achieve stable learning improvements across all tasks.  
Even the state-of-the-art LLMs, as widely recognized in recent studies~\cite{aime,rein2024gpqa,singh2024globalmmluunderstandingaddressing,hendrycks2021math,wang2019superglue}, such as o3-mini and Claude-3.7-Sonnet-Thinking, do not demonstrate a clear advantage in learning capability and efficiency across all tasks.  
Moreover, for every task, there are always some models that can learn from prior experience and improve their performance.  
For example, o3-mini exhibits strong learning capability in all tasks except sequential reasoning, whereas Claude-3.7-Sonnet can improve its performance on this task by leveraging experience from feedback.

These results indicate that each model has its own strengths, and that learning ability is not strongly correlated with static performance.  
Learning capability and efficiency thus provide a valuable new perspective for assessing model performance and understanding the gap between current models and human abilities.  
We also present a case study in Appendix~\ref{sec:case-study} to further illustrate how models learn from prior experience.

\begin{findingBox}{6}{
The average position of the first correct solution $P_\text{first}$ and the average offset of the first learned correct solution $P_\text{offset}$ vary across models and tasks, providing important insights into model potential.  
All metrics in EvaLearn capture different aspects of a model’s learning ability and efficiency, collectively revealing the model's learning potential.
}
\end{findingBox}

We analyze the average position of the first correct solution $P_{\text{first}}$, which measures how quickly models achieve initial success within a sequence.  
The right side of Figure~\ref{fig:main-consec} presents $P_{\text{first}}$ for o3-mini, with results for other models provided in Appendix~\ref{app:metric-3-first}.  
We observe that o3-mini and Doubao-1.5-Thinking-Pro achieve the best performance, with average positions of 1.57 and 1.80, respectively.  
These two models also achieve the highest overall accuracy in the feedback learning setting.  
Although $P_{\text{first}}$ is closely related to static model capability (since stronger models are more likely to solve problems earlier), notable exceptions exist.  
For example, DeepSeek-V3 achieves the same $P_{\text{first}}$ value as o3-mini on the mathematical reasoning task, yet its overall accuracy is 5.9\% lower, indicating that o3-mini can better improve its problem-solving ability through experience.

We also find substantial variation in model performance across different tasks.  
For example, most models struggle to solve their first problem early in logical reasoning sequences, with only o3-mini and Doubao-1.5-Thinking-Pro consistently achieving early success.
Additionally, comparing demonstration learning and feedback learning settings, as shown in Figures~\ref{fig:Demonstration P first} and~\ref{fig:Feedback_P_first}, seven out of nine models can solve a problem earlier when provided with feedback from the judging model.  
This indicates that, compared to simply providing canonical answers, feedback is more effective in facilitating model learning and mastery of a task.  
For instance, in the logical reasoning task, o3-mini sometimes fails to solve any problems in a sequence when using only demonstrations, but with feedback learning, it can solve at least one problem in every sequence.

Beyond $P_{\text{first}}$, we also analyze the average offset of the first learned correct solution $P_{\text{offset}}$, as shown in Table~\ref{tab:P-first} . On average, frontier LLMs attain their first learned success relatively late in a sequence—typically around three to four additional positions after the first zero-shot miss, such as Qwen2.5-32b-Instruct, DeepSeek-R1, and DeepSeek-V3. By contrast, o3-mini tends to reach this point earlier on average.
Clear task effects emerge. Models generally learn earlier on classification and extraction than on logical reasoning.
For models such as GPT-4o and Qwen2.5-32b-Instruct, the first learned correct solution typically appears earlier in classification and extraction, whereas logical reasoning usually requires more steps. Summarization displays substantial variability across models—early for o3-mini but much later for Doubao-1.5-Thinking-Pro and DeepSeek-R1.

The long chain-of-thought reasoning paradigm provides a slight efficiency gain within families. For example, Claude-3.7-Sonnet-Thinking attains a lower average $P_{\text{offset}}$ than Claude-3.7-Sonnet, and Doubao-1.5-Thinking-Pro slightly improves over Doubao-1.5-Pro, while the DeepSeek pair is mixed. Across families, the reasoning-focused o3-mini also tends to learn earlier than GPT-4o. We find that $P_{\text{offset}}$ highlights systematic task effects and modest benefits from long-form reasoning, while revealing substantial cross-model variability in true test-time learning efficiency.
Consistent with $P_{\text{first}}$, $P_{\text{offset}}$ varies widely across models and tasks and does not align with overall accuracy, underscoring that EvaLearn captures a dynamic learning dimension that is distinct from static capability.

Overall, all metrics comprehensively assess the learning performance from multiple perspectives.  
EvaLearn evaluates models in a more realistic and dynamic way, helping researchers better understand the gap between models and humans, and driving the development of more powerful models.

\section{Conclusion}

We present EvaLearn, a novel benchmark that sequentially evaluates the learning capability and efficiency of models within specific tasks.
EvaLearn is equipped with a suite of comprehensive metrics, revealing significant performance differences among frontier models across diverse tasks, including both thinking-based and non-thinking-based models.
Moreover, we find that while some models can effectively leverage teacher model feedback on previous solutions to enhance learning, others struggle to benefit from such feedback.
EvaLearn offers a new perspective on assessing the potential of LLMs and serves as a pioneering step toward dynamic evaluation.
The limitations and future work of EvaLearn are discussed in Appendix~\ref{sec:limitations}, and related work is presented in Appendix~\ref{sec:related-work}.

\section*{Acknowledgments}

We gratefully acknowledge the significant contributions made by the annotation teams at ByteDance, whose diligent work was essential to the success of this paper. 
The core members of the annotation team include Di Cheng, Linhua Deng, Yanxi Fu, Yafei Qiao, Chaoqian Ren, Mei Su, Ying Wu, Baitong Yang, and Xingyu Zhu.
We wish to express our sincere appreciation to an undisclosed third-party annotation company for their substantial support in data annotation.
We also want to thank all individuals who participated in and supported this project for their valuable input.
Finally, we wish to thank the AC and anonymous reviewers for their constructive comments.
This work was partially funded by National Natural Science Foundation of China (No.62476061, 62441602, 62206057), Shanghai Rising-Star Program (23QA1400200), and Natural Science Foundation of Shanghai (23ZR1403500).

% \section*{Contributions and Acknowledgments}

% \textbf{Contributors}

% Shihan Dou$^{1,2}$, Ming Zhang$^{2}$, Chenhao Huang$^{2}$, Jiayi Chen$^{2}$, Feng Chen$^{1,3}$, Shichun Liu$^{2}$, Yan Liu$^{1,2}$, Chenxiao Liu$^{1}$, Cheng Zhong$^{1}$, Chao Xin$^{1}$, Chengzhi Wei$^{1}$, Zongzhang Zhang$^{3}$, Tao Gui$^{2}$, Qi Zhang$^{2}$, Lin Yan$^{1}$, Yonghui Wu$^{1}$, Xuanjing Huang$^{2}$

% \textbf{Supervision}

% Chao Xin$^{1}$, Chengzhi Wei$^{1}$, Zongzhang Zhang$^{3}$, Tao Gui$^{2}$, Qi Zhang$^{2}$, Lin Yan$^{1}$, Yonghui Wu$^{1}$, Xuanjing Huang$^{2}$

% \textbf{Affiliations}

% $^1$ ByteDance Seed

% $^2$ NLP Group, Fudan University

% $^3$ LAMDA, Nanjing University

% \textbf{Acknowledgments}

% We gratefully acknowledge the significant contributions made by the annotation teams at ByteDance, whose diligent work was essential to the success of this paper. The core members of the annotation team include Di Cheng, Linhua Deng, Yanxi Fu, Yafei Qiao, Chaoqian Ren, Mei Su, Ying Wu, Baitong Yang, and Xingyu Zhu.
% We also wish to express our sincere appreciation to an undisclosed third-party annotation company for their substantial support in data annotation.
% Finally, we would like to thank all individuals who participated in and supported this project for their valuable input.

\bibliography{evalearn}

\begin{thebibliography}{105}
\providecommand{\natexlab}[1]{#1}
\providecommand{\url}[1]{\texttt{#1}}
\expandafter\ifx\csname urlstyle\endcsname\relax
  \providecommand{\doi}[1]{doi: #1}\else
  \providecommand{\doi}{doi: \begingroup \urlstyle{rm}\Url}\fi

\bibitem[Agarwal et~al.(2024)Agarwal, Singh, Zhang, Bohnet, Rosias, Chan, Zhang, Anand, Abbas, Nova, et~al.]{agarwal2024many}
Rishabh Agarwal, Avi Singh, Lei Zhang, Bernd Bohnet, Luis Rosias, Stephanie Chan, Biao Zhang, Ankesh Anand, Zaheer Abbas, Azade Nova, et~al.
\newblock Many-shot in-context learning.
\newblock \emph{Advances in Neural Information Processing Systems}, 37:\penalty0 76930--76966, 2024.

\bibitem[Aky{\"u}rek et~al.(2024)Aky{\"u}rek, Wang, Kim, and Andreas]{akyurek2024context}
Ekin Aky{\"u}rek, Bailin Wang, Yoon Kim, and Jacob Andreas.
\newblock In-context language learning: Architectures and algorithms.
\newblock In \emph{International Conference on Machine Learning}, pages 787--812. PMLR, 2024.

\bibitem[Anthropic(2025{\natexlab{a}})]{claude3.5}
Anthropic.
\newblock Claude 3.5 sonnet, 2025{\natexlab{a}}.
\newblock URL \url{https://www.anthropic.com/news/claude-3-5-sonnet}.

\bibitem[Anthropic(2025{\natexlab{b}})]{claude3.7}
Anthropic.
\newblock Claude 3.7 sonnet and claude code, 2025{\natexlab{b}}.
\newblock URL \url{https://www.anthropic.com/news/claude-3-7-sonnet}.

\bibitem[Austin et~al.(2021)Austin, Odena, Nye, Bosma, Michalewski, Dohan, Jiang, Cai, Terry, Le, et~al.]{austin2021program}
Jacob Austin, Augustus Odena, Maxwell Nye, Maarten Bosma, Henryk Michalewski, David Dohan, Ellen Jiang, Carrie Cai, Michael Terry, Quoc Le, et~al.
\newblock Program synthesis with large language models.
\newblock \emph{arXiv preprint arXiv:2108.07732}, 2021.

\bibitem[Bassett et~al.(2011)Bassett, Wymbs, Porter, Mucha, Carlson, and Grafton]{bassett2011dynamic}
Danielle~S Bassett, Nicholas~F Wymbs, Mason~A Porter, Peter~J Mucha, Jean~M Carlson, and Scott~T Grafton.
\newblock Dynamic reconfiguration of human brain networks during learning.
\newblock \emph{Proceedings of the National Academy of Sciences}, 108\penalty0 (18):\penalty0 7641--7646, 2011.

\bibitem[Bisk et~al.(2020)Bisk, Zellers, Gao, Choi, et~al.]{bisk2020piqa}
Yonatan Bisk, Rowan Zellers, Jianfeng Gao, Yejin Choi, et~al.
\newblock Piqa: Reasoning about physical commonsense in natural language.
\newblock In \emph{Proceedings of the AAAI conference on artificial intelligence}, volume~34, pages 7432--7439, 2020.

\bibitem[Brown et~al.(2020)Brown, Mann, Ryder, Subbiah, Kaplan, Dhariwal, Neelakantan, Shyam, Sastry, Askell, et~al.]{brown2020language}
Tom Brown, Benjamin Mann, Nick Ryder, Melanie Subbiah, Jared~D Kaplan, Prafulla Dhariwal, Arvind Neelakantan, Pranav Shyam, Girish Sastry, Amanda Askell, et~al.
\newblock Language models are few-shot learners.
\newblock \emph{Advances in neural information processing systems}, 33:\penalty0 1877--1901, 2020.

\bibitem[ByteDance(2025)]{doubao1.5pro}
ByteDance.
\newblock Doubao-1.5-pro, 2025.
\newblock URL \url{https://team.doubao.com/en/special/doubao_1_5_pro}.

\bibitem[Chang et~al.(2024)Chang, Wang, Wang, Wu, Yang, Zhu, Chen, Yi, Wang, Wang, et~al.]{chang2024survey}
Yupeng Chang, Xu~Wang, Jindong Wang, Yuan Wu, Linyi Yang, Kaijie Zhu, Hao Chen, Xiaoyuan Yi, Cunxiang Wang, Yidong Wang, et~al.
\newblock A survey on evaluation of large language models.
\newblock \emph{ACM transactions on intelligent systems and technology}, 15\penalty0 (3):\penalty0 1--45, 2024.

\bibitem[Chen et~al.(2024{\natexlab{a}})Chen, Lou, Chen, Bai, Xiang, Yang, Zhao, and Zhang]{chen2024dual}
Andong Chen, Lianzhang Lou, Kehai Chen, Xuefeng Bai, Yang Xiang, Muyun Yang, Tiejun Zhao, and Min Zhang.
\newblock Dual-reflect: Enhancing large language models for reflective translation through dual learning feedback mechanisms.
\newblock In \emph{Proceedings of the 62nd Annual Meeting of the Association for Computational Linguistics (Volume 2: Short Papers)}, pages 693--704, 2024{\natexlab{a}}.

\bibitem[Chen et~al.(2024{\natexlab{b}})Chen, Hu, Liu, Huang, Tu, He, and Wen]{chen2024llmarena}
Junzhe Chen, Xuming Hu, Shuodi Liu, Shiyu Huang, Wei-Wei Tu, Zhaofeng He, and Lijie Wen.
\newblock Llmarena: Assessing capabilities of large language models in dynamic multi-agent environments.
\newblock In \emph{Proceedings of the 62nd Annual Meeting of the Association for Computational Linguistics (Volume 1: Long Papers)}, pages 13055--13077, 2024{\natexlab{b}}.

\bibitem[Chen et~al.(2021)Chen, Tworek, Jun, Yuan, Pinto, Kaplan, Edwards, Burda, Joseph, Brockman, et~al.]{chen2021evaluating}
Mark Chen, Jerry Tworek, Heewoo Jun, Qiming Yuan, Henrique Ponde De~Oliveira Pinto, Jared Kaplan, Harri Edwards, Yuri Burda, Nicholas Joseph, Greg Brockman, et~al.
\newblock Evaluating large language models trained on code.
\newblock \emph{arXiv preprint arXiv:2107.03374}, 2021.

\bibitem[Chen et~al.(2025)Chen, Lin, Zhou, Huang, Jia, Cao, and Wen]{chen2025icleval}
Wentong Chen, Yankai Lin, ZhenHao Zhou, HongYun Huang, Yantao Jia, Zhao Cao, and Ji-Rong Wen.
\newblock Icleval: Evaluating in-context learning ability of large language models.
\newblock In \emph{Proceedings of the 31st International Conference on Computational Linguistics}, pages 10398--10422, 2025.

\bibitem[Chen et~al.(2024{\natexlab{c}})Chen, Lin, Sch{\"a}rli, and Zhou]{chen2024teaching}
Xinyun Chen, Maxwell Lin, Nathanael Sch{\"a}rli, and Denny Zhou.
\newblock Teaching large language models to self-debug.
\newblock In \emph{The Twelfth International Conference on Learning Representations}, 2024{\natexlab{c}}.
\newblock URL \url{https://openreview.net/forum?id=KuPixIqPiq}.

\bibitem[Chiang et~al.(2023)Chiang, Li, Lin, Sheng, Wu, Zhang, Zheng, Zhuang, Zhuang, Gonzalez, et~al.]{chiang2023vicuna}
Wei-Lin Chiang, Zhuohan Li, Ziqing Lin, Ying Sheng, Zhanghao Wu, Hao Zhang, Lianmin Zheng, Siyuan Zhuang, Yonghao Zhuang, Joseph~E Gonzalez, et~al.
\newblock Vicuna: An open-source chatbot impressing gpt-4 with 90\%* chatgpt quality.
\newblock \emph{See https://vicuna. lmsys. org (accessed 14 April 2023)}, 2\penalty0 (3):\penalty0 6, 2023.

\bibitem[Chollet(2019)]{chollet2019measure}
Fran{\c{c}}ois Chollet.
\newblock On the measure of intelligence.
\newblock \emph{arXiv preprint arXiv:1911.01547}, 2019.

\bibitem[Cobbe et~al.(2021)Cobbe, Kosaraju, Bavarian, Chen, Jun, Kaiser, Plappert, Tworek, Hilton, Nakano, et~al.]{cobbe2021training}
Karl Cobbe, Vineet Kosaraju, Mohammad Bavarian, Mark Chen, Heewoo Jun, Lukasz Kaiser, Matthias Plappert, Jerry Tworek, Jacob Hilton, Reiichiro Nakano, et~al.
\newblock Training verifiers to solve math word problems.
\newblock \emph{arXiv preprint arXiv:2110.14168}, 2021.

\bibitem[Coda-Forno et~al.(2023)Coda-Forno, Binz, Akata, Botvinick, Wang, and Schulz]{coda2023meta}
Julian Coda-Forno, Marcel Binz, Zeynep Akata, Matt Botvinick, Jane Wang, and Eric Schulz.
\newblock Meta-in-context learning in large language models.
\newblock \emph{Advances in Neural Information Processing Systems}, 36:\penalty0 65189--65201, 2023.

\bibitem[Deary et~al.(2010)Deary, Penke, and Johnson]{deary2010neuroscience}
Ian~J Deary, Lars Penke, and Wendy Johnson.
\newblock The neuroscience of human intelligence differences.
\newblock \emph{Nature reviews neuroscience}, 11\penalty0 (3):\penalty0 201--211, 2010.

\bibitem[Dong et~al.(2024)Dong, Li, Dai, Zheng, Ma, Li, Xia, Xu, Wu, Chang, et~al.]{dong2024survey}
Qingxiu Dong, Lei Li, Damai Dai, Ce~Zheng, Jingyuan Ma, Rui Li, Heming Xia, Jingjing Xu, Zhiyong Wu, Baobao Chang, et~al.
\newblock A survey on in-context learning.
\newblock In \emph{Proceedings of the 2024 Conference on Empirical Methods in Natural Language Processing}, pages 1107--1128, 2024.

\bibitem[Dua et~al.(2019)Dua, Wang, Dasigi, Stanovsky, Singh, and Gardner]{dua-etal-2019-drop}
Dheeru Dua, Yizhong Wang, Pradeep Dasigi, Gabriel Stanovsky, Sameer Singh, and Matt Gardner.
\newblock {DROP}: A reading comprehension benchmark requiring discrete reasoning over paragraphs.
\newblock In Jill Burstein, Christy Doran, and Thamar Solorio, editors, \emph{Proceedings of the 2019 Conference of the North {A}merican Chapter of the Association for Computational Linguistics: Human Language Technologies, Volume 1 (Long and Short Papers)}, pages 2368--2378, Minneapolis, Minnesota, June 2019. Association for Computational Linguistics.
\newblock \doi{10.18653/v1/N19-1246}.
\newblock URL \url{https://aclanthology.org/N19-1246/}.

\bibitem[Gao et~al.(2023)Gao, Xiong, Gao, Jia, Pan, Bi, Dai, Sun, Guo, Wang, and Wang]{Gao2023RetrievalAugmentedGF}
Yunfan Gao, Yun Xiong, Xinyu Gao, Kangxiang Jia, Jinliu Pan, Yuxi Bi, Yi~Dai, Jiawei Sun, Qianyu Guo, Meng Wang, and Haofen Wang.
\newblock Retrieval-augmented generation for large language models: A survey.
\newblock \emph{ArXiv}, abs/2312.10997, 2023.
\newblock URL \url{https://api.semanticscholar.org/CorpusID:266359151}.

\bibitem[Garg et~al.(2022)Garg, Tsipras, Liang, and Valiant]{garg2022can}
Shivam Garg, Dimitris Tsipras, Percy~S Liang, and Gregory Valiant.
\newblock What can transformers learn in-context? a case study of simple function classes.
\newblock \emph{Advances in Neural Information Processing Systems}, 35:\penalty0 30583--30598, 2022.

\bibitem[Gu et~al.(2024)Gu, Jiang, Shi, Tan, Zhai, Xu, Li, Shen, Ma, Liu, et~al.]{gu2024survey}
Jiawei Gu, Xuhui Jiang, Zhichao Shi, Hexiang Tan, Xuehao Zhai, Chengjin Xu, Wei Li, Yinghan Shen, Shengjie Ma, Honghao Liu, et~al.
\newblock A survey on llm-as-a-judge.
\newblock \emph{arXiv preprint arXiv:2411.15594}, 2024.

\bibitem[Guo et~al.(2025)Guo, Yang, Zhang, Song, Zhang, Xu, Zhu, Ma, Wang, Bi, et~al.]{guo2025deepseek}
Daya Guo, Dejian Yang, Haowei Zhang, Junxiao Song, Ruoyu Zhang, Runxin Xu, Qihao Zhu, Shirong Ma, Peiyi Wang, Xiao Bi, et~al.
\newblock Deepseek-r1: Incentivizing reasoning capability in llms via reinforcement learning.
\newblock \emph{arXiv preprint arXiv:2501.12948}, 2025.

\bibitem[Guo et~al.(2023)Guo, Jin, Liu, Huang, Shi, Yu, Liu, Li, Xiong, Xiong, et~al.]{guo2023evaluating}
Zishan Guo, Renren Jin, Chuang Liu, Yufei Huang, Dan Shi, Linhao Yu, Yan Liu, Jiaxuan Li, Bojian Xiong, Deyi Xiong, et~al.
\newblock Evaluating large language models: A comprehensive survey.
\newblock \emph{arXiv preprint arXiv:2310.19736}, 2023.

\bibitem[Hastie et~al.(2009)Hastie, Tibshirani, and Friedman]{hastie2009elements}
Trevor Hastie, Robert Tibshirani, and Jerome Friedman.
\newblock \emph{The Elements of Statistical Learning: Data Mining, Inference, and Prediction}.
\newblock Springer, New York, 2 edition, 2009.

\bibitem[Hendrycks et~al.(2021{\natexlab{a}})Hendrycks, Burns, Basart, Zou, Mazeika, Song, and Steinhardt]{hendrycks2021measuring}
Dan Hendrycks, Collin Burns, Steven Basart, Andy Zou, Mantas Mazeika, Dawn Song, and Jacob Steinhardt.
\newblock Measuring massive multitask language understanding.
\newblock In \emph{International Conference on Learning Representations}, 2021{\natexlab{a}}.
\newblock URL \url{https://openreview.net/forum?id=d7KBjmI3GmQ}.

\bibitem[Hendrycks et~al.(2021{\natexlab{b}})Hendrycks, Burns, Kadavath, Arora, Basart, Tang, Song, and Steinhardt]{hendrycks2021math}
Dan Hendrycks, Collin Burns, Saurav Kadavath, Akul Arora, Steven Basart, Eric Tang, Dawn Song, and Jacob Steinhardt.
\newblock Measuring mathematical problem solving with the {MATH} dataset.
\newblock In \emph{Thirty-fifth Conference on Neural Information Processing Systems Datasets and Benchmarks Track (Round 2)}, 2021{\natexlab{b}}.
\newblock URL \url{https://openreview.net/forum?id=7Bywt2mQsCe}.

\bibitem[Huang et~al.(2024)Huang, Vora, Liang, and Leskovec]{huang2024mlagentbench}
Qian Huang, Jian Vora, Percy Liang, and Jure Leskovec.
\newblock Mlagentbench: evaluating language agents on machine learning experimentation.
\newblock In \emph{Proceedings of the 41st International Conference on Machine Learning}, pages 20271--20309, 2024.

\bibitem[Huang et~al.(2023)Huang, Bai, Zhu, Zhang, Zhang, Su, Liu, Lv, Zhang, jiayi lei, Fu, Sun, and He]{huang2023ceval}
Yuzhen Huang, Yuzhuo Bai, Zhihao Zhu, Junlei Zhang, Jinghan Zhang, Tangjun Su, Junteng Liu, Chuancheng Lv, Yikai Zhang, jiayi lei, Yao Fu, Maosong Sun, and Junxian He.
\newblock C-eval: A multi-level multi-discipline chinese evaluation suite for foundation models.
\newblock In \emph{Thirty-seventh Conference on Neural Information Processing Systems Datasets and Benchmarks Track}, 2023.
\newblock URL \url{https://openreview.net/forum?id=fOrm2rGX2r}.

\bibitem[Jiang et~al.(2024)Jiang, Wang, Zeng, Zhong, Li, Mi, Shang, Jiang, Liu, and Wang]{jiang-etal-2024-followbench}
Yuxin Jiang, Yufei Wang, Xingshan Zeng, Wanjun Zhong, Liangyou Li, Fei Mi, Lifeng Shang, Xin Jiang, Qun Liu, and Wei Wang.
\newblock {F}ollow{B}ench: A multi-level fine-grained constraints following benchmark for large language models.
\newblock In Lun-Wei Ku, Andre Martins, and Vivek Srikumar, editors, \emph{Proceedings of the 62nd Annual Meeting of the Association for Computational Linguistics (Volume 1: Long Papers)}, pages 4667--4688, Bangkok, Thailand, August 2024. Association for Computational Linguistics.
\newblock URL \url{https://aclanthology.org/2024.acl-long.257}.

\bibitem[Jing et~al.(2023)Jing, Jin, Hu, Qiu, Wang, Wang, and Xiong]{jing2023followeval}
Yimin Jing, Renren Jin, Jiahao Hu, Huishi Qiu, Xiaohua Wang, Peng Wang, and Deyi Xiong.
\newblock Followeval: A multi-dimensional benchmark for assessing the instruction-following capability of large language models.
\newblock \emph{arXiv preprint arXiv:2311.09829}, 2023.

\bibitem[Koh et~al.(2024)Koh, Lo, Jang, Duvvur, Lim, Huang, Neubig, Zhou, Salakhutdinov, and Fried]{koh2024visualwebarena}
Jing~Yu Koh, Robert Lo, Lawrence Jang, Vikram Duvvur, Ming Lim, Po-Yu Huang, Graham Neubig, Shuyan Zhou, Russ Salakhutdinov, and Daniel Fried.
\newblock Visualwebarena: Evaluating multimodal agents on realistic visual web tasks.
\newblock In \emph{Proceedings of the 62nd Annual Meeting of the Association for Computational Linguistics (Volume 1: Long Papers)}, pages 881--905, 2024.

\bibitem[Lake et~al.(2017)Lake, Ullman, Tenenbaum, and Gershman]{lake2017building}
Brenden~M Lake, Tomer~D Ullman, Joshua~B Tenenbaum, and Samuel~J Gershman.
\newblock Building machines that learn and think like people.
\newblock \emph{Behavioral and brain sciences}, 40:\penalty0 e253, 2017.

\bibitem[Lan et~al.(2024)Lan, Zhang, Xu, Huang, Lin, Chen, and Mao]{lan2024criticeval}
Tian Lan, Wenwei Zhang, Chen Xu, Heyan Huang, Dahua Lin, Kai Chen, and Xian-Ling Mao.
\newblock Criticeval: Evaluating large-scale language model as critic.
\newblock \emph{Advances in Neural Information Processing Systems}, 37:\penalty0 66907--66960, 2024.

\bibitem[Langley et~al.(2009)Langley, Laird, and Rogers]{LANGLEY2009141}
Pat Langley, John~E. Laird, and Seth Rogers.
\newblock Cognitive architectures: {{Research}} issues and challenges.
\newblock \emph{Cognitive Systems Research}, 10\penalty0 (2):\penalty0 141--160, 2009.
\newblock ISSN 1389-0417.
\newblock \doi{10.1016/j.cogsys.2006.07.004}.

\bibitem[Laskar et~al.(2024)Laskar, Alqahtani, Bari, Rahman, Khan, Khan, Jahan, Bhuiyan, Tan, Parvez, et~al.]{laskar2024systematic}
Md~Tahmid~Rahman Laskar, Sawsan Alqahtani, M~Saiful Bari, Mizanur Rahman, Mohammad Abdullah~Matin Khan, Haidar Khan, Israt Jahan, Amran Bhuiyan, Chee~Wei Tan, Md~Rizwan Parvez, et~al.
\newblock A systematic survey and critical review on evaluating large language models: Challenges, limitations, and recommendations.
\newblock In \emph{Proceedings of the 2024 Conference on Empirical Methods in Natural Language Processing}, pages 13785--13816, 2024.

\bibitem[Li et~al.(2024)Li, Chiang, Frick, Dunlap, Wu, Zhu, Gonzalez, and Stoica]{li2024crowdsourced}
Tianle Li, Wei-Lin Chiang, Evan Frick, Lisa Dunlap, Tianhao Wu, Banghua Zhu, Joseph~E Gonzalez, and Ion Stoica.
\newblock From crowdsourced data to high-quality benchmarks: Arena-hard and benchbuilder pipeline.
\newblock \emph{arXiv preprint arXiv:2406.11939}, 2024.

\bibitem[Li et~al.(2021)Li, Arnold, Yan, Shi, and Yu]{li2021legoeval}
Yu~Li, Josh Arnold, Feifan Yan, Weiyan Shi, and Zhou Yu.
\newblock Legoeval: An open-source toolkit for dialogue system evaluation via crowdsourcing.
\newblock In \emph{Proceedings of the 59th Annual Meeting of the Association for Computational Linguistics and the 11th International Joint Conference on Natural Language Processing: System Demonstrations}, pages 317--324, 2021.

\bibitem[Lightman et~al.(2024)Lightman, Kosaraju, Burda, Edwards, Baker, Lee, Leike, Schulman, Sutskever, and Cobbe]{lightman2023letsverifystepstep}
Hunter Lightman, Vineet Kosaraju, Yuri Burda, Harrison Edwards, Bowen Baker, Teddy Lee, Jan Leike, John Schulman, Ilya Sutskever, and Karl Cobbe.
\newblock Let's verify step by step.
\newblock In \emph{The Twelfth International Conference on Learning Representations}, 2024.
\newblock URL \url{https://openreview.net/forum?id=v8L0pN6EOi}.

\bibitem[Lin et~al.(2022)Lin, Hilton, and Evans]{lin2022truthfulqa}
Stephanie Lin, Jacob Hilton, and Owain Evans.
\newblock Truthfulqa: Measuring how models mimic human falsehoods.
\newblock In \emph{Proceedings of the 60th Annual Meeting of the Association for Computational Linguistics (Volume 1: Long Papers)}, pages 3214--3252, 2022.

\bibitem[Lin et~al.(2024)Lin, Gou, Liang, Luo, Liu, and Yang]{lin2024criticbench}
Zicheng Lin, Zhibin Gou, Tian Liang, Ruilin Luo, Haowei Liu, and Yujiu Yang.
\newblock Criticbench: Benchmarking llms for critique-correct reasoning.
\newblock In \emph{Findings of the Association for Computational Linguistics ACL 2024}, pages 1552--1587, 2024.

\bibitem[Liu et~al.(2024{\natexlab{a}})Liu, Feng, Xue, Wang, Wu, Lu, Zhao, Deng, Zhang, Ruan, et~al.]{liu2024deepseek}
Aixin Liu, Bei Feng, Bing Xue, Bingxuan Wang, Bochao Wu, Chengda Lu, Chenggang Zhao, Chengqi Deng, Chenyu Zhang, Chong Ruan, et~al.
\newblock Deepseek-v3 technical report.
\newblock \emph{arXiv preprint arXiv:2412.19437}, 2024{\natexlab{a}}.

\bibitem[Liu et~al.(2024{\natexlab{b}})Liu, Dong, Hu, and Chu]{liu2024longgenbench}
Xiang Liu, Peijie Dong, Xuming Hu, and Xiaowen Chu.
\newblock Longgenbench: Long-context generation benchmark.
\newblock In \emph{Findings of the Association for Computational Linguistics: EMNLP 2024}, pages 865--883, 2024{\natexlab{b}}.

\bibitem[Liu et~al.(2024{\natexlab{c}})Liu, Lei, Wang, Huang, Feng, Wen, Cheng, Ke, Xu, Tam, Zhang, Sun, Gu, Wang, Zhang, Huang, Dong, and Tang]{liu2024alignbench}
Xiao Liu, Xuanyu Lei, Shengyuan Wang, Yue Huang, Andrew Feng, Bosi Wen, Jiale Cheng, Pei Ke, Yifan Xu, Weng~Lam Tam, Xiaohan Zhang, Lichao Sun, Xiaotao Gu, Hongning Wang, Jing Zhang, Minlie Huang, Yuxiao Dong, and Jie Tang.
\newblock Alignbench: Benchmarking chinese alignment of large language models.
\newblock In \emph{ACL (1)}, pages 11621--11640, 2024{\natexlab{c}}.
\newblock URL \url{https://aclanthology.org/2024.acl-long.624}.

\bibitem[Liu et~al.(2024{\natexlab{d}})Liu, Yu, Zhang, Xu, Lei, Lai, Gu, Ding, Men, Yang, Zhang, Deng, Zeng, Du, Zhang, Shen, Zhang, Su, Sun, Huang, Dong, and Tang]{liu2024agentbench}
Xiao Liu, Hao Yu, Hanchen Zhang, Yifan Xu, Xuanyu Lei, Hanyu Lai, Yu~Gu, Hangliang Ding, Kaiwen Men, Kejuan Yang, Shudan Zhang, Xiang Deng, Aohan Zeng, Zhengxiao Du, Chenhui Zhang, Sheng Shen, Tianjun Zhang, Yu~Su, Huan Sun, Minlie Huang, Yuxiao Dong, and Jie Tang.
\newblock Agentbench: Evaluating {LLM}s as agents.
\newblock In \emph{The Twelfth International Conference on Learning Representations}, 2024{\natexlab{d}}.
\newblock URL \url{https://openreview.net/forum?id=zAdUB0aCTQ}.

\bibitem[Lyu et~al.(2023)Lyu, Min, Beltagy, Zettlemoyer, and Hajishirzi]{lyu2023z}
Xinxi Lyu, Sewon Min, Iz~Beltagy, Luke Zettlemoyer, and Hannaneh Hajishirzi.
\newblock Z-icl: Zero-shot in-context learning with pseudo-demonstrations.
\newblock In \emph{Proceedings of the 61st Annual Meeting of the Association for Computational Linguistics (Volume 1: Long Papers)}, pages 2304--2317, 2023.

\bibitem[Ma et~al.(2025)Ma, Du, Wang, Zhang, ZhoufutuWen, Qu, Yang, Liu, minghao liu, Yue, Huang, and Zhang]{ma2025korbench}
Kaijing Ma, Xeron Du, Yunran Wang, Haoran Zhang, ZhoufutuWen, Xingwei Qu, Jian Yang, Jiaheng Liu, minghao liu, Xiang Yue, Wenhao Huang, and Ge~Zhang.
\newblock {KOR}-bench: Benchmarking language models on knowledge-orthogonal reasoning tasks.
\newblock In \emph{The Thirteenth International Conference on Learning Representations}, 2025.
\newblock URL \url{https://openreview.net/forum?id=SVRRQ8goQo}.

\bibitem[Madaan et~al.(2023)Madaan, Tandon, Gupta, Hallinan, Gao, Wiegreffe, Alon, Dziri, Prabhumoye, Yang, et~al.]{madaan2023self}
Aman Madaan, Niket Tandon, Prakhar Gupta, Skyler Hallinan, Luyu Gao, Sarah Wiegreffe, Uri Alon, Nouha Dziri, Shrimai Prabhumoye, Yiming Yang, et~al.
\newblock Self-refine: Iterative refinement with self-feedback.
\newblock \emph{Advances in Neural Information Processing Systems}, 36:\penalty0 46534--46594, 2023.

\bibitem[Mialon et~al.(2024)Mialon, Fourrier, Wolf, LeCun, and Scialom]{mialon2024gaia}
Gr{\'e}goire Mialon, Cl{\'e}mentine Fourrier, Thomas Wolf, Yann LeCun, and Thomas Scialom.
\newblock {GAIA}: a benchmark for general {AI} assistants.
\newblock In \emph{The Twelfth International Conference on Learning Representations}, 2024.
\newblock URL \url{https://openreview.net/forum?id=fibxvahvs3}.

\bibitem[Min et~al.(2022)Min, Lyu, Holtzman, Artetxe, Lewis, Hajishirzi, and Zettlemoyer]{min-etal-2022-rethinking}
Sewon Min, Xinxi Lyu, Ari Holtzman, Mikel Artetxe, Mike Lewis, Hannaneh Hajishirzi, and Luke Zettlemoyer.
\newblock Rethinking the role of demonstrations: What makes in-context learning work?
\newblock In \emph{Proceedings of the 2022 Conference on Empirical Methods in Natural Language Processing}, pages 11048--11064, Abu Dhabi, United Arab Emirates, December 2022. Association for Computational Linguistics.
\newblock \doi{10.18653/v1/2022.emnlp-main.759}.
\newblock URL \url{https://aclanthology.org/2022.emnlp-main.759/}.

\bibitem[Nathani et~al.(2025)Nathani, Madaan, Roberts, Bashlykov, Menon, Moens, Budhiraja, Magka, Vorotilov, Chaurasia, et~al.]{nathani2025mlgym}
Deepak Nathani, Lovish Madaan, Nicholas Roberts, Nikolay Bashlykov, Ajay Menon, Vincent Moens, Amar Budhiraja, Despoina Magka, Vladislav Vorotilov, Gaurav Chaurasia, et~al.
\newblock Mlgym: A new framework and benchmark for advancing ai research agents.
\newblock \emph{arXiv preprint arXiv:2502.14499}, 2025.

\bibitem[OpenAI(2023)]{gpt4}
OpenAI.
\newblock {GPT4} technical report.
\newblock \emph{arXiv preprint arXiv:2303.08774}, 2023.

\bibitem[OpenAI(2024)]{o1}
OpenAI.
\newblock Learning to reason with llms, 2024.
\newblock URL \url{https://openai.com/index/learning-to-reason-with-llms/}.

\bibitem[OpenAI(2025{\natexlab{a}})]{o3}
OpenAI.
\newblock Openai o3-mini, 2025{\natexlab{a}}.
\newblock URL \url{https://openai.com/index/openai-o3-mini/}.

\bibitem[OpenAI(2025{\natexlab{b}})]{o4mini}
OpenAI.
\newblock Introducing openai o3 and o4-mini, 2025{\natexlab{b}}.
\newblock URL \url{https://openai.com/index/introducing-o3-and-o4-mini/}.

\bibitem[Ouyang et~al.(2022)Ouyang, Wu, Jiang, Almeida, Wainwright, Mishkin, Zhang, Agarwal, Slama, Ray, et~al.]{ouyang2022training}
Long Ouyang, Jeffrey Wu, Xu~Jiang, Diogo Almeida, Carroll Wainwright, Pamela Mishkin, Chong Zhang, Sandhini Agarwal, Katarina Slama, Alex Ray, et~al.
\newblock Training language models to follow instructions with human feedback.
\newblock In \emph{Advances in Neural Information Processing Systems}, volume~35, pages 27730--27744, 2022.

\bibitem[Peng et~al.(2024)Peng, Zhu, Liu, Bo, Shi, Hong, Zhang, and Tang]{Peng2024GraphRG}
Boci Peng, Yun Zhu, Yongchao Liu, Xiaohe Bo, Haizhou Shi, Chuntao Hong, Yan Zhang, and Siliang Tang.
\newblock Graph retrieval-augmented generation: A survey.
\newblock \emph{ArXiv}, abs/2408.08921, 2024.
\newblock URL \url{https://api.semanticscholar.org/CorpusID:271903170}.

\bibitem[Qin et~al.(2024)Qin, Song, Hu, Yao, Cho, Wang, Wu, Liu, Liu, and Yu]{qin2024infobench}
Yiwei Qin, Kaiqiang Song, Yebowen Hu, Wenlin Yao, Sangwoo Cho, Xiaoyang Wang, Xuansheng Wu, Fei Liu, Pengfei Liu, and Dong Yu.
\newblock Infobench: Evaluating instruction following ability in large language models.
\newblock In \emph{Findings of the Association for Computational Linguistics ACL 2024}, pages 13025--13048, 2024.

\bibitem[Rein et~al.(2024)Rein, Hou, Stickland, Petty, Pang, Dirani, Michael, and Bowman]{rein2024gpqa}
David Rein, Betty~Li Hou, Asa~Cooper Stickland, Jackson Petty, Richard~Yuanzhe Pang, Julien Dirani, Julian Michael, and Samuel~R. Bowman.
\newblock {GPQA}: A graduate-level google-proof q\&a benchmark.
\newblock In \emph{First Conference on Language Modeling}, 2024.
\newblock URL \url{https://openreview.net/forum?id=Ti67584b98}.

\bibitem[Scheurer et~al.(2022)Scheurer, Campos, Chan, Chen, Cho, and Perez]{scheurer2022training}
J{\'e}r{\'e}my Scheurer, Jon~Ander Campos, Jun~Shern Chan, Angelica Chen, Kyunghyun Cho, and Ethan Perez.
\newblock Training language models with language feedback.
\newblock \emph{arXiv preprint arXiv:2204.14146}, 2022.

\bibitem[Schick et~al.(2023)Schick, Dwivedi-Yu, Dess{\`\i}, Raileanu, Lomeli, Hambro, Zettlemoyer, Cancedda, and Scialom]{schick2023toolformer}
Timo Schick, Jane Dwivedi-Yu, Roberto Dess{\`\i}, Roberta Raileanu, Maria Lomeli, Eric Hambro, Luke Zettlemoyer, Nicola Cancedda, and Thomas Scialom.
\newblock Toolformer: Language models can teach themselves to use tools.
\newblock \emph{Advances in Neural Information Processing Systems}, 36:\penalty0 68539--68551, 2023.

\bibitem[Seed et~al.(2025)Seed, Yuan, Yue, Wang, Zuo, Chen, Yan, Xu, Zhang, Liu, et~al.]{seed2025seed}
ByteDance Seed, Yufeng Yuan, Yu~Yue, Mingxuan Wang, Xiaochen Zuo, Jiaze Chen, Lin Yan, Wenyuan Xu, Chi Zhang, Xin Liu, et~al.
\newblock Seed-thinking-v1. 5: Advancing superb reasoning models with reinforcement learning.
\newblock \emph{arXiv preprint arXiv:2504.13914}, 2025.

\bibitem[Shinn et~al.(2023)Shinn, Cassano, Gopinath, Narasimhan, and Yao]{shinn2023reflexion}
Noah Shinn, Federico Cassano, Ashwin Gopinath, Karthik Narasimhan, and Shunyu Yao.
\newblock Reflexion: Language agents with verbal reinforcement learning.
\newblock \emph{Advances in Neural Information Processing Systems}, 36:\penalty0 8634--8652, 2023.

\bibitem[Singh et~al.(2023)Singh, Chan, Moskovitz, Grant, Saxe, and Hill]{singh2023transient}
Aaditya Singh, Stephanie Chan, Ted Moskovitz, Erin Grant, Andrew Saxe, and Felix Hill.
\newblock The transient nature of emergent in-context learning in transformers.
\newblock \emph{Advances in Neural Information Processing Systems}, 36:\penalty0 27801--27819, 2023.

\bibitem[Singh et~al.(2024)Singh, Romanou, Fourrier, Adelani, Ngui, Vila-Suero, Limkonchotiwat, Marchisio, Leong, Susanto, Ng, Longpre, Ko, Smith, Bosselut, Oh, Martins, Choshen, Ippolito, Ferrante, Fadaee, Ermis, and Hooker]{singh2024globalmmluunderstandingaddressing}
Shivalika Singh, Angelika Romanou, Clémentine Fourrier, David~I. Adelani, Jian~Gang Ngui, Daniel Vila-Suero, Peerat Limkonchotiwat, Kelly Marchisio, Wei~Qi Leong, Yosephine Susanto, Raymond Ng, Shayne Longpre, Wei-Yin Ko, Madeline Smith, Antoine Bosselut, Alice Oh, Andre F.~T. Martins, Leshem Choshen, Daphne Ippolito, Enzo Ferrante, Marzieh Fadaee, Beyza Ermis, and Sara Hooker.
\newblock Global mmlu: Understanding and addressing cultural and linguistic biases in multilingual evaluation, 2024.
\newblock URL \url{https://arxiv.org/abs/2412.03304}.

\bibitem[Sirdeshmukh et~al.(2025)Sirdeshmukh, Deshpande, Mols, Jin, Cardona, Lee, Kritz, Primack, Yue, and Xing]{sirdeshmukh2025multichallenge}
Ved Sirdeshmukh, Kaustubh Deshpande, Johannes Mols, Lifeng Jin, Ed-Yeremai Cardona, Dean Lee, Jeremy Kritz, Willow Primack, Summer Yue, and Chen Xing.
\newblock Multichallenge: A realistic multi-turn conversation evaluation benchmark challenging to frontier llms.
\newblock \emph{arXiv preprint arXiv:2501.17399}, 2025.

\bibitem[Sprague et~al.(2024)Sprague, Ye, Bostrom, Chaudhuri, and Durrett]{sprague2024musr}
Zayne~Rea Sprague, Xi~Ye, Kaj Bostrom, Swarat Chaudhuri, and Greg Durrett.
\newblock Mu{SR}: Testing the limits of chain-of-thought with multistep soft reasoning.
\newblock In \emph{The Twelfth International Conference on Learning Representations}, 2024.
\newblock URL \url{https://openreview.net/forum?id=jenyYQzue1}.

\bibitem[Srivastava et~al.(2022)Srivastava, Rastogi, Rao, Shoeb, Abid, Fisch, Brown, Santoro, Gupta, Garriga-Alonso, et~al.]{srivastava2022beyond}
Aarohi Srivastava, Abhinav Rastogi, Abhishek Rao, Abu Awal~Md Shoeb, Abubakar Abid, Adam Fisch, Adam~R Brown, Adam Santoro, Aditya Gupta, Adri{\`a} Garriga-Alonso, et~al.
\newblock Beyond the imitation game: Quantifying and extrapolating the capabilities of language models.
\newblock \emph{TRANSACTIONS ON MACHINE LEARNING RESEARCH}, 2022.

\bibitem[Su et~al.(2024)Su, Tai, Ji, Li, Bowen, and Zhang]{su2024demonstration}
Yi~Su, Yunpeng Tai, Yixin Ji, Juntao Li, Yan Bowen, and Min Zhang.
\newblock Demonstration augmentation for zero-shot in-context learning.
\newblock In \emph{Findings of the Association for Computational Linguistics ACL 2024}, pages 14232--14244, 2024.

\bibitem[Suzgun et~al.(2023)Suzgun, Scales, Sch{\"a}rli, Gehrmann, Tay, Chung, Chowdhery, Le, Chi, Zhou, and Wei]{suzgun-etal-2023-challenging}
Mirac Suzgun, Nathan Scales, Nathanael Sch{\"a}rli, Sebastian Gehrmann, Yi~Tay, Hyung~Won Chung, Aakanksha Chowdhery, Quoc Le, Ed~Chi, Denny Zhou, and Jason Wei.
\newblock Challenging {BIG}-bench tasks and whether chain-of-thought can solve them.
\newblock In Anna Rogers, Jordan Boyd-Graber, and Naoaki Okazaki, editors, \emph{Findings of the Association for Computational Linguistics: ACL 2023}, pages 13003--13051, Toronto, Canada, July 2023. Association for Computational Linguistics.
\newblock \doi{10.18653/v1/2023.findings-acl.824}.
\newblock URL \url{https://aclanthology.org/2023.findings-acl.824/}.

\bibitem[Tang et~al.(2023)Tang, Liu, Cai, Shao, Lu, Zhang, Deng, Hu, An, Huang, et~al.]{tang2023ml}
Xiangru Tang, Yuliang Liu, Zefan Cai, Yanjun Shao, Junjie Lu, Yichi Zhang, Zexuan Deng, Helan Hu, Kaikai An, Ruijun Huang, et~al.
\newblock Ml-bench: Evaluating large language models and agents for machine learning tasks on repository-level code.
\newblock \emph{arXiv preprint arXiv:2311.09835}, 2023.

\bibitem[Team(2025)]{qwq32b}
Qwen Team.
\newblock Qwq-32b: Embracing the power of reinforcement learning, March 2025.
\newblock URL \url{https://qwenlm.github.io/blog/qwq-32b/}.

\bibitem[Tenison et~al.(2016)Tenison, Fincham, and Anderson]{tenisonPhasesLearningHow2016}
Caitlin Tenison, Jon~M. Fincham, and John~R. Anderson.
\newblock Phases of learning: {{How}} skill acquisition impacts cognitive processing.
\newblock \emph{Cognitive Psychology}, 87:\penalty0 1--28, June 2016.
\newblock ISSN 1095-5623.
\newblock \doi{10.1016/j.cogpsych.2016.03.001}.

\bibitem[Veeraboina(2023)]{aime}
Hemish Veeraboina.
\newblock Aime problem set 1983-2024, 2023.
\newblock URL \url{https://www.kaggle.com/datasets/hemishveeraboina/aime-problem-set-1983-2024}.

\bibitem[Wang et~al.(2019)Wang, Pruksachatkun, Nangia, Singh, Michael, Hill, Levy, and Bowman]{wang2019superglue}
Alex Wang, Yada Pruksachatkun, Nikita Nangia, Amanpreet Singh, Julian Michael, Felix Hill, Omer Levy, and Samuel Bowman.
\newblock Superglue: A stickier benchmark for general-purpose language understanding systems.
\newblock \emph{Advances in neural information processing systems}, 32, 2019.

\bibitem[Wang(2021)]{wang2021meta}
Jane~X Wang.
\newblock Meta-learning in natural and artificial intelligence.
\newblock \emph{Current Opinion in Behavioral Sciences}, 38:\penalty0 90--95, 2021.

\bibitem[Wang et~al.(2024{\natexlab{a}})Wang, Ma, Zhang, Ni, Chandra, Guo, Ren, Arulraj, He, Jiang, Li, Ku, Wang, Zhuang, Fan, Yue, and Chen]{wang2024mmlupro}
Yubo Wang, Xueguang Ma, Ge~Zhang, Yuansheng Ni, Abhranil Chandra, Shiguang Guo, Weiming Ren, Aaran Arulraj, Xuan He, Ziyan Jiang, Tianle Li, Max Ku, Kai Wang, Alex Zhuang, Rongqi Fan, Xiang Yue, and Wenhu Chen.
\newblock {MMLU}-pro: A more robust and challenging multi-task language understanding benchmark.
\newblock In \emph{The Thirty-eight Conference on Neural Information Processing Systems Datasets and Benchmarks Track}, 2024{\natexlab{a}}.
\newblock URL \url{https://openreview.net/forum?id=y10DM6R2r3}.

\bibitem[Wang et~al.(2024{\natexlab{b}})Wang, Zeng, Liu, Meng, Zhou, and Zhang]{wang2024taste}
Yutong Wang, Jiali Zeng, Xuebo Liu, Fandong Meng, Jie Zhou, and Min Zhang.
\newblock Taste: Teaching large language models to translate through self-reflection.
\newblock In \emph{Proceedings of the 62nd Annual Meeting of the Association for Computational Linguistics (Volume 1: Long Papers)}, pages 6144--6158, 2024{\natexlab{b}}.

\bibitem[Wen et~al.(2024)Wen, Ke, Gu, Wu, Huang, Zhou, Li, Hu, Gao, Xu, et~al.]{wen2024benchmarking}
Bosi Wen, Pei Ke, Xiaotao Gu, Lindong Wu, Hao Huang, Jinfeng Zhou, Wenchuang Li, Binxin Hu, Wendy Gao, Jiaxing Xu, et~al.
\newblock Benchmarking complex instruction-following with multiple constraints composition.
\newblock \emph{Advances in Neural Information Processing Systems}, 37:\penalty0 137610--137645, 2024.

\bibitem[Wies et~al.(2023)Wies, Levine, and Shashua]{wies2023learnability}
Noam Wies, Yoav Levine, and Amnon Shashua.
\newblock The learnability of in-context learning.
\newblock \emph{Advances in Neural Information Processing Systems}, 36:\penalty0 36637--36651, 2023.

\bibitem[Xi et~al.(2024)Xi, Ding, Chen, Hong, Guo, Wang, Yang, Liao, Guo, He, et~al.]{xi2024agentgym}
Zhiheng Xi, Yiwen Ding, Wenxiang Chen, Boyang Hong, Honglin Guo, Junzhe Wang, Dingwen Yang, Chenyang Liao, Xin Guo, Wei He, et~al.
\newblock Agentgym: Evolving large language model-based agents across diverse environments.
\newblock \emph{arXiv preprint arXiv:2406.04151}, 2024.

\bibitem[Xia et~al.(2024)Xia, Xing, Du, Yang, Feng, Xu, Yin, and Xiong]{xia-etal-2024-fofo}
Congying Xia, Chen Xing, Jiangshu Du, Xinyi Yang, Yihao Feng, Ran Xu, Wenpeng Yin, and Caiming Xiong.
\newblock {FOFO}: A benchmark to evaluate {LLM}s' format-following capability.
\newblock In Lun-Wei Ku, Andre Martins, and Vivek Srikumar, editors, \emph{Proceedings of the 62nd Annual Meeting of the Association for Computational Linguistics (Volume 1: Long Papers)}, pages 680--699, Bangkok, Thailand, August 2024. Association for Computational Linguistics.
\newblock \doi{10.18653/v1/2024.acl-long.40}.
\newblock URL \url{https://aclanthology.org/2024.acl-long.40/}.

\bibitem[Xu et~al.(2024{\natexlab{a}})Xu, Sun, Zheng, Geng, Zhao, Feng, Tao, Lin, and Jiang]{xu2024wizardlm}
Can Xu, Qingfeng Sun, Kai Zheng, Xiubo Geng, Pu~Zhao, Jiazhan Feng, Chongyang Tao, Qingwei Lin, and Daxin Jiang.
\newblock Wizard{LM}: Empowering large pre-trained language models to follow complex instructions.
\newblock In \emph{The Twelfth International Conference on Learning Representations}, 2024{\natexlab{a}}.
\newblock URL \url{https://openreview.net/forum?id=CfXh93NDgH}.

\bibitem[Xu et~al.(2024{\natexlab{b}})Xu, Song, Li, Tang, Jain, Bao, Wang, Zhou, Guo, Cao, et~al.]{xu2024theagentcompany}
Frank~F Xu, Yufan Song, Boxuan Li, Yuxuan Tang, Kritanjali Jain, Mengxue Bao, Zora~Z Wang, Xuhui Zhou, Zhitong Guo, Murong Cao, et~al.
\newblock Theagentcompany: benchmarking llm agents on consequential real world tasks.
\newblock \emph{arXiv preprint arXiv:2412.14161}, 2024{\natexlab{b}}.

\bibitem[Xu et~al.(2024{\natexlab{c}})Xu, Wu, Diao, Liu, Wang, Chen, and Gao]{xu2024sayself}
Tianyang Xu, Shujin Wu, Shizhe Diao, Xiaoze Liu, Xingyao Wang, Yangyi Chen, and Jing Gao.
\newblock Sayself: Teaching llms to express confidence with self-reflective rationales.
\newblock In \emph{Proceedings of the 2024 Conference on Empirical Methods in Natural Language Processing}, pages 5985--5998, 2024{\natexlab{c}}.

\bibitem[Yan et~al.(2024)Yan, Zhu, Wang, Gui, and He]{yan2024mirror}
Hanqi Yan, Qinglin Zhu, Xinyu Wang, Lin Gui, and Yulan He.
\newblock Mirror: Multiple-perspective self-reflection method for knowledge-rich reasoning.
\newblock In \emph{Proceedings of the 62nd Annual Meeting of the Association for Computational Linguistics (Volume 1: Long Papers)}, pages 7086--7103, 2024.

\bibitem[Yang et~al.(2024)Yang, Yang, Zhang, Hui, Zheng, Yu, Li, Liu, Huang, Wei, Lin, Yang, Tu, Zhang, Yang, Yang, Zhou, Lin, Dang, Lu, Bao, Yang, Yu, Li, Xue, Zhang, Zhu, Men, Lin, Li, Xia, Ren, Ren, Fan, Su, Zhang, Wan, Liu, Cui, Zhang, and Qiu]{qwen2.5}
An~Yang, Baosong Yang, Beichen Zhang, Binyuan Hui, Bo~Zheng, Bowen Yu, Chengyuan Li, Dayiheng Liu, Fei Huang, Haoran Wei, Huan Lin, Jian Yang, Jianhong Tu, Jianwei Zhang, Jianxin Yang, Jiaxi Yang, Jingren Zhou, Junyang Lin, Kai Dang, Keming Lu, Keqin Bao, Kexin Yang, Le~Yu, Mei Li, Mingfeng Xue, Pei Zhang, Qin Zhu, Rui Men, Runji Lin, Tianhao Li, Tingyu Xia, Xingzhang Ren, Xuancheng Ren, Yang Fan, Yang Su, Yichang Zhang, Yu~Wan, Yuqiong Liu, Zeyu Cui, Zhenru Zhang, and Zihan Qiu.
\newblock Qwen2.5 technical report.
\newblock \emph{arXiv preprint arXiv:2412.15115}, 2024.

\bibitem[Yao et~al.(2024)Yao, Shinn, Razavi, and Narasimhan]{yao2024tau}
Shunyu Yao, Noah Shinn, Pedram Razavi, and Karthik Narasimhan.
\newblock $\tau$-bench: A benchmark for tool-agent-user interaction in real-world domains.
\newblock \emph{arXiv preprint arXiv:2406.12045}, 2024.

\bibitem[Ye et~al.(2023)Ye, Wu, Feng, Yu, and Kong]{ye2023compositional}
Jiacheng Ye, Zhiyong Wu, Jiangtao Feng, Tao Yu, and Lingpeng Kong.
\newblock Compositional exemplars for in-context learning.
\newblock In \emph{International Conference on Machine Learning}, pages 39818--39833. PMLR, 2023.

\bibitem[Zellers et~al.(2019)Zellers, Holtzman, Bisk, Farhadi, and Choi]{zellers2019hellaswag}
Rowan Zellers, Ari Holtzman, Yonatan Bisk, Ali Farhadi, and Yejin Choi.
\newblock Hellaswag: Can a machine really finish your sentence?
\newblock In \emph{Proceedings of the 57th Annual Meeting of the Association for Computational Linguistics}, pages 4791--4800, 2019.

\bibitem[Zhang et~al.(2023)Zhang, Zhang, Liu, Yuan, Wang, Dong, Deng, Gui, Zhang, and Huang]{zhang2023llmeval}
Ming Zhang, Yue Zhang, Shichun Liu, Haipeng Yuan, Junzhe Wang, Yurui Dong, Jingyi Deng, Tao Gui, Qi~Zhang, and Xuanjing Huang.
\newblock Llmeval-2, 2023.

\bibitem[Zhang et~al.(2025)Zhang, Shen, Deng, Wang, Zhang, Wang, Liu, Dou, Sha, Peng, et~al.]{zhang2025llmeval}
Ming Zhang, Yujiong Shen, Jingyi Deng, Yuhui Wang, Yue Zhang, Junzhe Wang, Shichun Liu, Shihan Dou, Huayu Sha, Qiyuan Peng, et~al.
\newblock Llmeval-3: A large-scale longitudinal study on robust and fair evaluation of large language models.
\newblock \emph{arXiv preprint arXiv:2508.05452}, 2025.

\bibitem[Zhang et~al.(2024{\natexlab{a}})Zhang, Yu, Fu, Huang, and Li]{zhang2024iopo}
Xinghua Zhang, Haiyang Yu, Cheng Fu, Fei Huang, and Yongbin Li.
\newblock Iopo: Empowering llms with complex instruction following via input-output preference optimization.
\newblock \emph{arXiv preprint arXiv:2411.06208}, 2024{\natexlab{a}}.

\bibitem[Zhang et~al.(2024{\natexlab{b}})Zhang, Zhang, Yuan, Liu, Shi, Gui, Zhang, and Huang]{zhang2024llmeval}
Yue Zhang, Ming Zhang, Haipeng Yuan, Shichun Liu, Yongyao Shi, Tao Gui, Qi~Zhang, and Xuanjing Huang.
\newblock Llmeval: A preliminary study on how to evaluate large language models.
\newblock In \emph{Proceedings of the AAAI Conference on Artificial Intelligence}, volume~38, pages 19615--19622, 2024{\natexlab{b}}.

\bibitem[Zhang et~al.(2024{\natexlab{c}})Zhang, Cui, Lu, Zhou, Yang, Wang, and Huang]{zhang2024agent}
Zhexin Zhang, Shiyao Cui, Yida Lu, Jingzhuo Zhou, Junxiao Yang, Hongning Wang, and Minlie Huang.
\newblock Agent-safetybench: Evaluating the safety of llm agents.
\newblock \emph{arXiv preprint arXiv:2412.14470}, 2024{\natexlab{c}}.

\bibitem[Zhao et~al.(2025)Zhao, Pei, Tao, Mei, and Shou]{zhao2025interfeedback}
Henry~Hengyuan Zhao, Wenqi Pei, Yifei Tao, Haiyang Mei, and Mike~Zheng Shou.
\newblock Interfeedback: Unveiling interactive intelligence of large multimodal models via human feedback.
\newblock \emph{arXiv preprint arXiv:2502.15027}, 2025.

\bibitem[Zhao et~al.(2024{\natexlab{a}})Zhao, Tong, Mou, Zhang, Zhang, and Huang]{zhao2024exploring}
Jun Zhao, Jingqi Tong, Yurong Mou, Ming Zhang, Qi~Zhang, and Xuan-Jing Huang.
\newblock Exploring the compositional deficiency of large language models in mathematical reasoning through trap problems.
\newblock In \emph{Proceedings of the 2024 Conference on Empirical Methods in Natural Language Processing}, pages 16361--16376, 2024{\natexlab{a}}.

\bibitem[Zhao et~al.(2024{\natexlab{b}})Zhao, Yang, Wang, He, Qiu, and Qiu]{Zhao2024RetrievalAG}
Siyun Zhao, Yuqing Yang, Zilong Wang, Zhiyuan He, Luna~K. Qiu, and Lili Qiu.
\newblock Retrieval augmented generation (rag) and beyond: A comprehensive survey on how to make your llms use external data more wisely.
\newblock \emph{ArXiv}, abs/2409.14924, 2024{\natexlab{b}}.
\newblock URL \url{https://api.semanticscholar.org/CorpusID:272827955}.

\bibitem[Zheng et~al.(2023)Zheng, Chiang, Sheng, Zhuang, Wu, Zhuang, Lin, Li, Li, Xing, et~al.]{zheng2023judging}
Lianmin Zheng, Wei-Lin Chiang, Ying Sheng, Siyuan Zhuang, Zhanghao Wu, Yonghao Zhuang, Zi~Lin, Zhuohan Li, Dacheng Li, Eric Xing, et~al.
\newblock Judging llm-as-a-judge with mt-bench and chatbot arena.
\newblock \emph{Advances in Neural Information Processing Systems}, 36:\penalty0 46595--46623, 2023.

\bibitem[Zhong et~al.(2024)Zhong, Cui, Guo, Liang, Lu, Wang, Saied, Chen, and Duan]{zhong2024agieval}
Wanjun Zhong, Ruixiang Cui, Yiduo Guo, Yaobo Liang, Shuai Lu, Yanlin Wang, Amin Saied, Weizhu Chen, and Nan Duan.
\newblock Agieval: A human-centric benchmark for evaluating foundation models.
\newblock In \emph{Findings of the Association for Computational Linguistics: NAACL 2024}, pages 2299--2314, 2024.

\bibitem[Zhou et~al.(2023)Zhou, Lu, Mishra, Brahma, Basu, Luan, Zhou, and Hou]{zhou2023instruction}
Jeffrey Zhou, Tianjian Lu, Swaroop Mishra, Siddhartha Brahma, Sujoy Basu, Yi~Luan, Denny Zhou, and Le~Hou.
\newblock Instruction-following evaluation for large language models.
\newblock \emph{arXiv preprint arXiv:2311.07911}, 2023.

\bibitem[Zhou et~al.(2024)Zhou, Xu, Zhu, Zhou, Lo, Sridhar, Cheng, Ou, Bisk, Fried, Alon, and Neubig]{zhou2024webarena}
Shuyan Zhou, Frank~F. Xu, Hao Zhu, Xuhui Zhou, Robert Lo, Abishek Sridhar, Xianyi Cheng, Tianyue Ou, Yonatan Bisk, Daniel Fried, Uri Alon, and Graham Neubig.
\newblock Webarena: A realistic web environment for building autonomous agents.
\newblock In \emph{The Twelfth International Conference on Learning Representations}, 2024.
\newblock URL \url{https://openreview.net/forum?id=oKn9c6ytLx}.

\end{thebibliography}
\bibliographystyle{plainnat}

\clearpage

\section*{NeurIPS Paper Checklist}

\begin{enumerate}

\item {\bf Claims}
    \item[] Question: Do the main claims made in the abstract and introduction accurately reflect the paper's contributions and scope?
    \item[] Answer: \answerYes{} % Replace by \answerYes{}, \answerNo{}, or \answerNA{}.
    \item[] Justification: We have clarified our claims in the abstract and introduction.
    \item[] Guidelines:
    \begin{itemize}
        \item The answer NA means that the abstract and introduction do not include the claims made in the paper.
        \item The abstract and/or introduction should clearly state the claims made, including the contributions made in the paper and important assumptions and limitations. A No or NA answer to this question will not be perceived well by the reviewers. 
        \item The claims made should match theoretical and experimental results, and reflect how much the results can be expected to generalize to other settings. 
        \item It is fine to include aspirational goals as motivation as long as it is clear that these goals are not attained by the paper. 
    \end{itemize}

\item {\bf Limitations}
    \item[] Question: Does the paper discuss the limitations of the work performed by the authors?
    \item[] Answer: \answerYes{} % Replace by \answerYes{}, \answerNo{}, or \answerNA{}.
    \item[] Justification: Justification: We have discussed the limitations and future work in Section 5.
    \item[] Guidelines:
    \begin{itemize}
        \item The answer NA means that the paper has no limitation while the answer No means that the paper has limitations, but those are not discussed in the paper. 
        \item The authors are encouraged to create a separate "Limitations" section in their paper.
        \item The paper should point out any strong assumptions and how robust the results are to violations of these assumptions (e.g., independence assumptions, noiseless settings, model well-specification, asymptotic approximations only holding locally). The authors should reflect on how these assumptions might be violated in practice and what the implications would be.
        \item The authors should reflect on the scope of the claims made, e.g., if the approach was only tested on a few datasets or with a few runs. In general, empirical results often depend on implicit assumptions, which should be articulated.
        \item The authors should reflect on the factors that influence the performance of the approach. For example, a facial recognition algorithm may perform poorly when image resolution is low or images are taken in low lighting. Or a speech-to-text system might not be used reliably to provide closed captions for online lectures because it fails to handle technical jargon.
        \item The authors should discuss the computational efficiency of the proposed algorithms and how they scale with dataset size.
        \item If applicable, the authors should discuss possible limitations of their approach to address problems of privacy and fairness.
        \item While the authors might fear that complete honesty about limitations might be used by reviewers as grounds for rejection, a worse outcome might be that reviewers discover limitations that aren't acknowledged in the paper. The authors should use their best judgment and recognize that individual actions in favor of transparency play an important role in developing norms that preserve the integrity of the community. Reviewers will be specifically instructed to not penalize honesty concerning limitations.
    \end{itemize}

\item {\bf Theory assumptions and proofs}
    \item[] Question: For each theoretical result, does the paper provide the full set of assumptions and a complete (and correct) proof?
    \item[] Answer: \answerNA{} % Replace by \answerYes{}, \answerNo{}, or \answerNA{}.
    \item[] Justification: In this work, we propose a novel benchmark that does not include theoretical assumptions or proof.
    \item[] Guidelines:
    \begin{itemize}
        \item The answer NA means that the paper does not include theoretical results. 
        \item All the theorems, formulas, and proofs in the paper should be numbered and cross-referenced.
        \item All assumptions should be clearly stated or referenced in the statement of any theorems.
        \item The proofs can either appear in the main paper or the supplemental material, but if they appear in the supplemental material, the authors are encouraged to provide a short proof sketch to provide intuition. 
        \item Inversely, any informal proof provided in the core of the paper should be complemented by formal proofs provided in appendix or supplemental material.
        \item Theorems and Lemmas that the proof relies upon should be properly referenced. 
    \end{itemize}

    \item {\bf Experimental result reproducibility}
    \item[] Question: Does the paper fully disclose all the information needed to reproduce the main experimental results of the paper to the extent that it affects the main claims and/or conclusions of the paper (regardless of whether the code and data are provided or not)?
    \item[] Answer: \answerYes{} % Replace by \answerYes{}, \answerNo{}, or \answerNA{}.
    \item[] Justification: We have shown our experiment results in Section 3. We have also fully disclosed all experimental setups in Section 3 and Appendix. We also open-source all datasets, the automatic evaluation framework, and the results studied in this paper in the supplemental material.
    \item[] Guidelines:
    \begin{itemize}
        \item The answer NA means that the paper does not include experiments.
        \item If the paper includes experiments, a No answer to this question will not be perceived well by the reviewers: Making the paper reproducible is important, regardless of whether the code and data are provided or not.
        \item If the contribution is a dataset and/or model, the authors should describe the steps taken to make their results reproducible or verifiable. 
        \item Depending on the contribution, reproducibility can be accomplished in various ways. For example, if the contribution is a novel architecture, describing the architecture fully might suffice, or if the contribution is a specific model and empirical evaluation, it may be necessary to either make it possible for others to replicate the model with the same dataset, or provide access to the model. In general. releasing code and data is often one good way to accomplish this, but reproducibility can also be provided via detailed instructions for how to replicate the results, access to a hosted model (e.g., in the case of a large language model), releasing of a model checkpoint, or other means that are appropriate to the research performed.
        \item While NeurIPS does not require releasing code, the conference does require all submissions to provide some reasonable avenue for reproducibility, which may depend on the nature of the contribution. For example
        \begin{enumerate}
            \item If the contribution is primarily a new algorithm, the paper should make it clear how to reproduce that algorithm.
            \item If the contribution is primarily a new model architecture, the paper should describe the architecture clearly and fully.
            \item If the contribution is a new model (e.g., a large language model), then there should either be a way to access this model for reproducing the results or a way to reproduce the model (e.g., with an open-source dataset or instructions for how to construct the dataset).
            \item We recognize that reproducibility may be tricky in some cases, in which case authors are welcome to describe the particular way they provide for reproducibility. In the case of closed-source models, it may be that access to the model is limited in some way (e.g., to registered users), but it should be possible for other researchers to have some path to reproducing or verifying the results.
        \end{enumerate}
    \end{itemize}

\item {\bf Open access to data and code}
    \item[] Question: Does the paper provide open access to the data and code, with sufficient instructions to faithfully reproduce the main experimental results, as described in supplemental material?
    \item[] Answer: \answerYes{} % Replace by \answerYes{}, \answerNo{}, or \answerNA{}.
    \item[] Justification: We open-source all datasets, the automatic evaluation framework, and the results studied in this paper in the supplemental material. We have also included a full explanation.
    \item[] Guidelines:
    \begin{itemize}
        \item The answer NA means that paper does not include experiments requiring code.
        \item Please see the NeurIPS code and data submission guidelines (\url{https://nips.cc/public/guides/CodeSubmissionPolicy}) for more details.
        \item While we encourage the release of code and data, we understand that this might not be possible, so "No" is an acceptable answer. Papers cannot be rejected simply for not including code, unless this is central to the contribution (e.g., for a new open-source benchmark).
        \item The instructions should contain the exact command and environment needed to run to reproduce the results. See the NeurIPS code and data submission guidelines (\url{https://nips.cc/public/guides/CodeSubmissionPolicy}) for more details.
        \item The authors should provide instructions on data access and preparation, including how to access the raw data, preprocessed data, intermediate data, and generated data, etc.
        \item The authors should provide scripts to reproduce all experimental results for the new proposed method and baselines. If only a subset of experiments are reproducible, they should state which ones are omitted from the script and why.
        \item At submission time, to preserve anonymity, the authors should release anonymized versions (if applicable).
        \item Providing as much information as possible in supplemental material (appended to the paper) is recommended, but including URLs to data and code is permitted.
    \end{itemize}

\item {\bf Experimental setting/details}
    \item[] Question: Does the paper specify all the training and test details (e.g., data splits, hyperparameters, how they were chosen, type of optimizer, etc.) necessary to understand the results?
    \item[] Answer: \answerYes{} % Replace by \answerYes{}, \answerNo{}, or \answerNA{}.
    \item[] Justification: In this work, we propose a novel benchmark that does not require training models. We have included all experimental setups in Section 3.
    \item[] Guidelines:
    \begin{itemize}
        \item The answer NA means that the paper does not include experiments.
        \item The experimental setting should be presented in the core of the paper to a level of detail that is necessary to appreciate the results and make sense of them.
        \item The full details can be provided either with the code, in appendix, or as supplemental material.
    \end{itemize}

\item {\bf Experiment statistical significance}
    \item[] Question: Does the paper report error bars suitably and correctly defined or other appropriate information about the statistical significance of the experiments?
    \item[] Answer: \answerYes{} % Replace by \answerYes{}, \answerNo{}, or \answerNA{}.
    \item[] Justification: All models were evaluated via their respective APIs, and their implementations are provided in Appendix D.
    \item[] Guidelines:
    \begin{itemize}
        \item The answer NA means that the paper does not include experiments.
        \item The authors should answer "Yes" if the results are accompanied by error bars, confidence intervals, or statistical significance tests, at least for the experiments that support the main claims of the paper.
        \item The factors of variability that the error bars are capturing should be clearly stated (for example, train/test split, initialization, random drawing of some parameter, or overall run with given experimental conditions).
        \item The method for calculating the error bars should be explained (closed form formula, call to a library function, bootstrap, etc.)
        \item The assumptions made should be given (e.g., Normally distributed errors).
        \item It should be clear whether the error bar is the standard deviation or the standard error of the mean.
        \item It is OK to report 1-sigma error bars, but one should state it. The authors should preferably report a 2-sigma error bar than state that they have a 96\% CI, if the hypothesis of Normality of errors is not verified.
        \item For asymmetric distributions, the authors should be careful not to show in tables or figures symmetric error bars that would yield results that are out of range (e.g. negative error rates).
        \item If error bars are reported in tables or plots, The authors should explain in the text how they were calculated and reference the corresponding figures or tables in the text.
    \end{itemize}

\item {\bf Experiments compute resources}
    \item[] Question: For each experiment, does the paper provide sufficient information on the computer resources (type of compute workers, memory, time of execution) needed to reproduce the experiments?
    \item[] Answer: \answerYes{} % Replace by \answerYes{}, \answerNo{}, or \answerNA{}.
    \item[] Justification: In this work, we propose a novel benchmark that does not require training models. We report the API costs in Appendix D.
    \item[] Guidelines:
    \begin{itemize}
        \item The answer NA means that the paper does not include experiments.
        \item The paper should indicate the type of compute workers CPU or GPU, internal cluster, or cloud provider, including relevant memory and storage.
        \item The paper should provide the amount of compute required for each of the individual experimental runs as well as estimate the total compute. 
        \item The paper should disclose whether the full research project required more compute than the experiments reported in the paper (e.g., preliminary or failed experiments that didn't make it into the paper). 
    \end{itemize}
    
\item {\bf Code of ethics}
    \item[] Question: Does the research conducted in the paper conform, in every respect, with the NeurIPS Code of Ethics \url{https://neurips.cc/public/EthicsGuidelines}?
    \item[] Answer: \answerYes{} % Replace by \answerYes{}, \answerNo{}, or \answerNA{}.
    \item[] Justification: The research conducted in the paper conforms, in every respect, with the NeurIPS Code of Ethics.
    \item[] Guidelines:
    \begin{itemize}
        \item The answer NA means that the authors have not reviewed the NeurIPS Code of Ethics.
        \item If the authors answer No, they should explain the special circumstances that require a deviation from the Code of Ethics.
        \item The authors should make sure to preserve anonymity (e.g., if there is a special consideration due to laws or regulations in their jurisdiction).
    \end{itemize}

\item {\bf Broader impacts}
    \item[] Question: Does the paper discuss both potential positive societal impacts and negative societal impacts of the work performed?
    \item[] Answer: \answerYes{} % Replace by \answerYes{}, \answerNo{}, or \answerNA{}.
    \item[] Justification: We have discussed both potential positive societal impacts and negative societal impacts of the work in Appendix A.
    \item[] Guidelines:
    \begin{itemize}
        \item The answer NA means that there is no societal impact of the work performed.
        \item If the authors answer NA or No, they should explain why their work has no societal impact or why the paper does not address societal impact.
        \item Examples of negative societal impacts include potential malicious or unintended uses (e.g., disinformation, generating fake profiles, surveillance), fairness considerations (e.g., deployment of technologies that could make decisions that unfairly impact specific groups), privacy considerations, and security considerations.
        \item The conference expects that many papers will be foundational research and not tied to particular applications, let alone deployments. However, if there is a direct path to any negative applications, the authors should point it out. For example, it is legitimate to point out that an improvement in the quality of generative models could be used to generate deepfakes for disinformation. On the other hand, it is not needed to point out that a generic algorithm for optimizing neural networks could enable people to train models that generate Deepfakes faster.
        \item The authors should consider possible harms that could arise when the technology is being used as intended and functioning correctly, harms that could arise when the technology is being used as intended but gives incorrect results, and harms following from (intentional or unintentional) misuse of the technology.
        \item If there are negative societal impacts, the authors could also discuss possible mitigation strategies (e.g., gated release of models, providing defenses in addition to attacks, mechanisms for monitoring misuse, mechanisms to monitor how a system learns from feedback over time, improving the efficiency and accessibility of ML).
    \end{itemize}
    
\item {\bf Safeguards}
    \item[] Question: Does the paper describe safeguards that have been put in place for responsible release of data or models that have a high risk for misuse (e.g., pretrained language models, image generators, or scraped datasets)?
    \item[] Answer: \answerNA{} % Replace by \answerYes{}, \answerNo{}, or \answerNA{}.
    \item[] Justification: The paper poses no such risks. All the data in EvaLearn is used to measure the learning capability of LLMs, which does not contain any biased information.
    \item[] Guidelines:
    \begin{itemize}
        \item The answer NA means that the paper poses no such risks.
        \item Released models that have a high risk for misuse or dual-use should be released with necessary safeguards to allow for controlled use of the model, for example by requiring that users adhere to usage guidelines or restrictions to access the model or implementing safety filters. 
        \item Datasets that have been scraped from the Internet could pose safety risks. The authors should describe how they avoided releasing unsafe images.
        \item We recognize that providing effective safeguards is challenging, and many papers do not require this, but we encourage authors to take this into account and make a best faith effort.
    \end{itemize}

\item {\bf Licenses for existing assets}
    \item[] Question: Are the creators or original owners of assets (e.g., code, data, models), used in the paper, properly credited and are the license and terms of use explicitly mentioned and properly respected?
    \item[] Answer: \answerYes{} % Replace by \answerYes{}, \answerNo{}, or \answerNA{}.
    \item[] Justification: We have cited the original papers that produced the datasets used in our paper. 
    We have included the version information and corresponding licenses in Appendix B. 
    We have included the licenses for our assets in the supplementary materials.
    \item[] Guidelines:
    \begin{itemize}
        \item The answer NA means that the paper does not use existing assets.
        \item The authors should cite the original paper that produced the code package or dataset.
        \item The authors should state which version of the asset is used and, if possible, include a URL.
        \item The name of the license (e.g., CC-BY 4.0) should be included for each asset.
        \item For scraped data from a particular source (e.g., website), the copyright and terms of service of that source should be provided.
        \item If assets are released, the license, copyright information, and terms of use in the package should be provided. For popular datasets, \url{paperswithcode.com/datasets} has curated licenses for some datasets. Their licensing guide can help determine the license of a dataset.
        \item For existing datasets that are re-packaged, both the original license and the license of the derived asset (if it has changed) should be provided.
        \item If this information is not available online, the authors are encouraged to reach out to the asset's creators.
    \end{itemize}

\item {\bf New assets}
    \item[] Question: Are new assets introduced in the paper well documented and is the documentation provided alongside the assets?
    \item[] Answer: \answerYes{} % Replace by \answerYes{}, \answerNo{}, or \answerNA{}.
    \item[] Justification: We have included detailed descriptions and usage of the assets in the supplementary materials.
    \item[] Guidelines:
    \begin{itemize}
        \item The answer NA means that the paper does not release new assets.
        \item Researchers should communicate the details of the dataset/code/model as part of their submissions via structured templates. This includes details about training, license, limitations, etc. 
        \item The paper should discuss whether and how consent was obtained from people whose asset is used.
        \item At submission time, remember to anonymize your assets (if applicable). You can either create an anonymized URL or include an anonymized zip file.
    \end{itemize}

\item {\bf Crowdsourcing and research with human subjects}
    \item[] Question: For crowdsourcing experiments and research with human subjects, does the paper include the full text of instructions given to participants and screenshots, if applicable, as well as details about compensation (if any)? 
    \item[] Answer: \answerNA{} % Replace by \answerYes{}, \answerNo{}, or \answerNA{}.
    \item[] Justification: In our work, all annotation was performed by company staff with relevant professional expertise, who were compensated with standard salaries.
    \item[] Guidelines:
    \begin{itemize}
        \item The answer NA means that the paper does not involve crowdsourcing nor research with human subjects.
        \item Including this information in the supplemental material is fine, but if the main contribution of the paper involves human subjects, then as much detail as possible should be included in the main paper. 
        \item According to the NeurIPS Code of Ethics, workers involved in data collection, curation, or other labor should be paid at least the minimum wage in the country of the data collector. 
    \end{itemize}

\item {\bf Institutional review board (IRB) approvals or equivalent for research with human subjects}
    \item[] Question: Does the paper describe potential risks incurred by study participants, whether such risks were disclosed to the subjects, and whether Institutional Review Board (IRB) approvals (or an equivalent approval/review based on the requirements of your country or institution) were obtained?
    \item[] Answer: \answerNA{} % Replace by \answerYes{}, \answerNo{}, or \answerNA{}.
    \item[] Justification: The paper does not involve crowdsourcing nor research with human subjects.
    \item[] Guidelines:
    \begin{itemize}
        \item The answer NA means that the paper does not involve crowdsourcing nor research with human subjects.
        \item Depending on the country in which research is conducted, IRB approval (or equivalent) may be required for any human subjects research. If you obtained IRB approval, you should clearly state this in the paper. 
        \item We recognize that the procedures for this may vary significantly between institutions and locations, and we expect authors to adhere to the NeurIPS Code of Ethics and the guidelines for their institution. 
        \item For initial submissions, do not include any information that would break anonymity (if applicable), such as the institution conducting the review.
    \end{itemize}

\item {\bf Declaration of LLM usage}
    \item[] Question: Does the paper describe the usage of LLMs if it is an important, original, or non-standard component of the core methods in this research? Note that if the LLM is used only for writing, editing, or formatting purposes and does not impact the core methodology, scientific rigorousness, or originality of the research, declaration is not required.
    %this research? 
    \item[] Answer: \answerNA{} % Replace by \answerYes{}, \answerNo{}, or \answerNA{}.
    \item[] Justification: We only use LLM to help us polish our manuscript and make it more readable. It does not impact the core methodology, scientific rigorousness, and originality of the research.
    \item[] Guidelines:
    \begin{itemize}
        \item The answer NA means that the core method development in this research does not involve LLMs as any important, original, or non-standard components.
        \item Please refer to our LLM policy (\url{https://neurips.cc/Conferences/2025/LLM}) for what should or should not be described.
    \end{itemize}

\end{enumerate}

%%%%%%%%%%%%%%%%%%%%%%%%%%%%%%%%%%%%%%%%%%%%%%%%%%%%%%%%%%%%
\clearpage
\appendix

% \section*{Appendix Contents}
% % \startcontents[Appendix Contents]

% \begingroup
%   % \renewcommand{\contentsname}{附录目录}
%   \tableofcontents
% \endgroup

\section{Limitations and Future Work}
\label{sec:limitations}

EvaLearn is the first benchmark to quantitatively evaluate the rapid learning capabilities of LLMs on a single task. 
We discuss its limitations, the new insights it offers for the community, and the directions it opens for future research into the causes of learning differences between models and the development of learning methods.

\textbf{Towards including highly challenging open-ended problems.}
EvaLearn focuses on problems that can be reliably evaluated with well-defined rubrics, ensuring trustworthy automated assessment.
However, during the annotation process, we found that some highly challenging open-ended problems—where systematic instance-level rubrics are difficult to construct—are not yet included, even though such problems can be valuable for human learning and reasoning.
In the future, we aim to develop new evaluation strategies that can effectively assess model responses to open-ended tasks, further expanding the task coverage of EvaLearn.

\textbf{Towards benchmarking more open-source LLMs.}
We benchmark current frontier LLMs and reveal significant differences in their task learning performance. While these advanced models substantially outperform existing open-source models, there is still limited understanding of how open-source models compare in terms of task learning ability. 
In future work, we plan to extend our evaluation to a broader range of open-source models, aiming to better quantify this gap and foster progress within the open-source LLM community.

\textbf{Exploring the roots of learning capability and efficiency differences.}
Our analysis quantifies the learning capability and efficiency of the models, and reveals that these are not strongly correlated with the models’ static abilities.
The fundamental factors driving differences in these aspects across models warrant deeper investigation. 
In future work, we plan to conduct more detailed studies to uncover the underlying causes of these disparities. 
We hope that EvaLearn will inspire the community to further explore and understand the mechanisms underlying LLM learning capability and efficiency.

\textbf{Broadening the exploration of learning methods.}
EvaLearn has investigated the impact of two learning methods on both learning capability and efficiency. 
However, LLMs may benefit from a broader range of strategies, such as integration with retrieval-augmented generation (RAG) \cite{Gao2023RetrievalAugmentedGF,Zhao2024RetrievalAG,Peng2024GraphRG}. 
In future work, we will leverage EvaLearn to systematically analyze the effectiveness of additional learning methods in task learning scenarios.
We also hope that EvaLearn will encourage the community to innovate and develop more effective approaches for rapid adaptation to new tasks.

\section{Related Work}
\label{sec:related-work}

\textbf{Evaluation Perspective and Paradigm.}
Previous work on evaluating LLMs has mainly focused on evaluating their static capabilities \cite{chang2024survey,laskar2024systematic,guo2023evaluating,zhang2025llmeval}, such as factual knowledge \cite{rein2024gpqa,hendrycks2021measuring,huang2023ceval,zhang2024llmeval,zhang2023llmeval,wang2024mmlupro,singh2024globalmmluunderstandingaddressing}, reasoning \cite{hendrycks2021math,cobbe2021training,zhao2024exploring,wang2019superglue,lan2024criticeval,lin2024criticbench,zhong2024agieval,sprague2024musr,ma2025korbench,lightman2023letsverifystepstep}, instruction following \cite{zhang2024iopo,zhou2023instruction,qin2024infobench,wen2024benchmarking,jing2023followeval,xu2024wizardlm,chiang2023vicuna}, and code generation \cite{tang2023ml,chen2021evaluating,austin2021program}.
% , and long-context processing \cite{bai2024longbench,li2024long,gavin2024longins,wu2024lifbench,wang2024needle,an2024eval}. 
% However, they merely reflect preexisting capabilities and fail to capture a model's ability to adapt to new tasks or integrate newly acquired information.
% With the rapid advancement of foundation models \cite{guo2025deepseek,seed2025seed,qwen2.5,qwen2,qwq32b,o1,gpt4,claude3.7,doubao1.5pro}, performance on such static benchmarks has improved significantly in a short time. 
% This trend suggests that merely evaluating static abilities is no longer sufficient \cite{mcintosh2024inadequacies,banerjee2024vulnerability}.
% It is increasingly important to assess the more dynamic aspects of the performance. 
% Our proposed EvaLearn shifts the focus to the evaluation of dynamic capabilities, which is specifically designed to evaluate the learning capability and efficiency of LLMs, filling a critical void in the current evaluation landscape. 
% \textbf{Evaluation Paradigm.} 
Current evaluation paradigms can be broadly categorized into two types: parallel evaluation and interactive evaluation \cite{dua-etal-2019-drop,suzgun-etal-2023-challenging,sirdeshmukh2025multichallenge,schick2023toolformer,chollet2019measure,li2021legoeval,yao2024tau}.
The vast majority of existing benchmarks adopt a parallel evaluation approach.
This evaluation method only measures the static capabilities of models, providing little insight into their capability to learn or adapt dynamically over time \cite{zellers2019hellaswag,srivastava2022beyond,lin2022truthfulqa,bisk2020piqa,liu2024alignbench,li2024crowdsourced,jiang-etal-2024-followbench,xia-etal-2024-fofo,liu2024longgenbench,zhao2025interfeedback}.
On the other hand, interactive evaluation is commonly used in multi-turn behavior assessments \cite{zheng2023judging,zhou2024webarena,chen2024llmarena,zhang2024agent,huang2024mlagentbench,xu2024theagentcompany,koh2024visualwebarena}.
This paradigm emphasizes the dynamic evaluation of a model’s ability to engage in multi-turn interactions, where tasks themselves remain i.i.d., but actions or states are temporally dependent.
Interactive evaluation more closely reflects real-world scenarios \cite{liu2024agentbench,xi2024agentgym,nathani2025mlgym,mialon2024gaia}.
Our work also falls under the category of dynamic evaluation.
However, unlike these methods, EvaLearn requires models to sequentially solve a series of problems within the same task, evaluating learning capability and efficiency, providing a direct indication of the model’s learning potential.
It fills a critical void in the current evaluation landscape.

\textbf{Learning Paradigms.}
We compare two learning paradigms—learning from demonstrations \cite{brown2020language,min-etal-2022-rethinking,dong2024survey} and learning from feedback \cite{shinn2023reflexion}—in terms of their ability to stimulate model learning.
Broadly, the former enables models to learn output formats and problem-solving strategies from demonstrations \cite{lyu2023z,su2024demonstration,garg2022can,agarwal2024many,coda2023meta,singh2023transient,wies2023learnability,ye2023compositional,chen2025icleval,akyurek2024context}, while the latter allows models to leverage experience gained from previous problems to improve subsequent performance \cite{chen2024teaching,xu2024sayself,wang2024taste,yan2024mirror,madaan2023self,chen2024dual}.
Both paradigms can be incorporated into model training to enhance overall capability \cite{ouyang2022training,scheurer2022training}.
In this paper, we apply both approaches at inference time, requiring models to sequentially solve a series of problems to assess whether they can benefit from prior canonical solutions or experience when tackling additional problems.
Importantly, the evaluation metrics in EvaLearn are decoupled from the choice of learning method, ensuring the flexibility and general applicability.

% Our work primarily leverages test-time learning paradigms to investigate the learning capability and efficiency of LLMs in sequential problem solving. 
% These paradigms include zero-shot learning \cite{brown2020language}, learning from demonstration \cite{min-etal-2022-rethinking}, and learning from feedback \cite{shinn2023reflexion}. 
% They represent the model's ability to generalize, analogize, and self-correct without updating its parameters \cite{dong2024survey}.
% Zero-shot learning directly evaluates a model's general reasoning ability in the absence of task-specific examples, serving as a key measure of its pretrained knowledge \cite{lyu2023z,su2024demonstration}.
% Learning from demonstration, i.e., few-shot learning, typically implemented via in-context learning\cite{garg2022can,agarwal2024many,coda2023meta,singh2023transient,wies2023learnability,ye2023compositional,chen2025icleval,akyurek2024context}, provides the model with a small number of high-quality demonstrations to adapt to new tasks. 

% This approach effectively activates the pattern matching capabilities and the analogical reasoning mechanisms of the model \cite{crosbie2024induction,hu2023context}.
% Learning from feedback allows the model to iteratively improve its outputs through self-reflection or external signals, enabling refinement without gradient-based updates \cite{ouyang2022training,scheurer2022training,chen2024teaching,xu2024sayself,wang2024taste,yan2024mirror,madaan2023self,chen2024dual}.

\section{Broader Impact}

In this work, we propose a pioneering benchmark, EvaLearn, which is designed to dynamically assess the learning ability and efficiency of LLMs.
We hope EvaLearn provides a new evaluation perspective for assessing LLM potential and promotes the development of deeper and more dynamic evaluation approaches.
All the data in EvaLearn is used to measure the learning capability of LLMs, which does not contain any biased information.
We do not see any negative societal impacts of this work.

\section{License For Artifacts and Data Consent}
All models used in this work, whether open-source or closed-source, are permitted for academic research. The licenses for DeepSeek-V3 and DeepSeek-R1 are MIT, while the license for Qwen2.5-32B-Instruct is Apache 2.0.

\section{Data Annotation Process}
\label{sec:annotation}

We adopt a hybrid approach combining human expertise and advanced LLMs for data annotation.
The annotation process consists of three main stages.

\textbf{Problem Collection.}
We begin with a collection of questions sourced from our model API platform, which are anonymized to remove any personally identifiable information or identification markers.
Four annotators independently and carefully review each question, identifying those for which humans could potentially improve their performance through repeated practice on similar problems.
Only questions unanimously judged as "learnable" by all annotators are retained as seed problems.
These seed problems are then categorized into six task categories.
Next, we use Claude-3.7 \cite{claude3.7} to analyze each problem, identifying the required skill dimensions and suggesting potential directions for increasing problem complexity.
With the aid of these model-generated strategies, nine annotators write additional similar and challenging problems for each seed question, expanding datasets with diverse variants.

\textbf{Problem filtering and canonical answer annotation.}
To further increase the difficulty of the dataset, we leverage advanced LLMs to screen the seed problems. 
Specifically, for each question, we randomly select three LLMs from a model list (as shown in Table~\ref{tab:model-list}) to generate candidate answers. 
This approach helps avoid bias toward or against any specific model.
Seven annotators then review these answers to write the correct canonical answer for each question, and assess the correctness of each model-generated response, providing explanations for their judgments. 
To ensure high data quality, an additional three annotators double-check both the written canonical answers and the explanations for the correctness judgments of the model responses.
We retain only those challenging questions for which all three models fail to provide a correct answer, thereby maintaining a high level of difficulty among the seed problems.

All annotators involved in this work were fairly compensated in accordance with the labor standards of their respective countries.

\textbf{Instance-level rubric annotation and validation.}
Eight annotators are responsible for writing rubrics for each retained challenging problem, which serve as criteria for evaluating the correctness of model responses. 
Annotators first determine whether each problem has a unique correct answer.
For problems with a unique answer, the rubric is constructed as a concise summary of the canonical answer. 
For problems with multiple valid answers, annotators write rubrics that comprehensively cover all acceptable solutions, clearly specifying the required conditions for correctness.
Overly open-ended problems that are difficult to evaluate (e.g., brainstorming tasks) are excluded from the dataset.

Each rubric is then validated by two annotators. 
Specifically, for every problem, three models are randomly selected from the model list to generate three candidate answers.
Annotators assess whether the rubric can accurately evaluate all responses and provide explanations. 
Rubrics that fail this validation step are revised and re-evaluated.
An additional three annotators conduct a final quality check on all fields, including the canonical answers, the explanations for the correctness judgments of the model responses, and the rubrics.
Finally, we construct problem sequences from the retained set of challenging problems. 
Each sequence consists of problems from the same task category, and the order of problems within each sequence is randomized.

\begin{table}[htbp]
    \centering
    \caption{The model list used for ensuring the difficulty of datasets and validating the effectiveness of instance-level rubrics.}
    \begin{tabular}{l l}
        \toprule
        \multicolumn{1}{c}{\textbf{Model}} & \multicolumn{1}{c}{\textbf{Published date}} \\
        \midrule
        Claude-3.7-Sonnet \cite{claude3.7} &  February 25, 2025\\
        Claude-3.7-Sonnet-Thinking \cite{claude3.7} &  February 25, 2025\\
        Claude-3.5-Sonnet \cite{claude3.5} & June 21, 2024\\
        DeepSeek-V3 \cite{liu2024deepseek} & December 26, 2024\\
        DeepSeek-R1 \cite{guo2025deepseek} & January 15, 2025\\
        Doubao-1.5-Pro-32k \cite{doubao1.5pro} & May 9, 2025\\
        Doubao-1.5-Thinking-Pro \cite{seed2025seed} & April 17, 2025\\
        Doubao-1.5-Pro-256k \cite{doubao1.5pro} & May 9, 2025 \\ 
        Qwen2.5-7b-Instruct \cite{qwen2.5} & September 19, 2024 \\ % qwen2.5 7b
        Qwen2.5-32b-Instruct \cite{qwen2.5} & September 19, 2024\\
        QwQ-32b \cite{qwq32b} & March 6, 2025\\ % qwq
        OpenAI-o3-mini \cite{o3} & January 31, 2025\\
        OpenAI-o1 \cite{o1} & September 12, 2024\\ % o1
        OpenAI-o4-mini \cite{o4mini} & April 16, 2025\\ % o4-mini
        \bottomrule
    \end{tabular}
    \label{tab:model-list}
\end{table}

\section{List of Evaluation Models}
\label{sec:open-source-models} 

In this work, we comprehensively evaluate the learning capability and efficiency of nine state-of-the-art LLMs.
These LLMs contain: 

\textbf{(1) Claude-3.7-Sonnet} \cite{claude3.7} is the most intelligent model released by Anthropic to date, particularly in terms of coding capabilities, which is an upgraded version of Claude 3.5 Sonnet. 

\textbf{(2) Claude-3.7-Sonnet-Thinking} \cite{claude3.7} is the extended thinking mode of Claude-3.7-Sonnet, which performs self-reflection before answering in order to improve performance on tasks such as math, physics, instruction-following, coding, and many others.

\textbf{(3) DeepSeek-V3} \cite{liu2024deepseek} is a strong Mixture-of-Experts (MoE) language model with 671B total parameters with 37B activated for each token.

\textbf{(4) DeepSeek-R1} \cite{guo2025deepseek} is the first-generation reasoning model developed by DeepSeek AI, which incorporates multi-stage training and cold-start data before reinforcement learning, enabling it to demonstrate strong reasoning capabilities.

\textbf{(5) Doubao-1.5-Pro} \cite{doubao1.5pro} is a high-performance sparse MoE large language model that focuses on achieving an optimal balance between inference performance and model capability developed by the Doubao team.

\textbf{(6) Doubao-1.5-Thinking-Pro} \cite{seed2025seed} shares the same model base as Seed-Thinking-v1.5, which is an advancing superb reasoning model with reinforcement learning. It is a MoE model with a relatively small size, featuring 20B activated and 200B total parameters.
It is capable of reasoning through thinking before responding, resulting in improved performance on a wide range of benchmarks.

\textbf{(7) Qwen2.5-32b-Instruct} \cite{qwen2.5} is the instruction-tuned 32B version of the Qwen2.5 series, which is an improved iteration based on the Qwen2 family.

\textbf{(8) GPT-4o} \cite{gpt4} is a state-of-the-art LLM developed by OpenAI, which demonstrates strong performance across a wide range of tasks, including natural language understanding, complex reasoning, instruction following, code generation, and problem solving. 
It achieves competitive results on challenging benchmarks in mathematics, logic, and programming, and maintains high performance across diverse evaluation settings.

\textbf{(9) OpenAI-o3-mini} \cite{o3} is a cost-efficient model in the reasoning series, previewed in December 2024, which pushes the boundaries of what small models can achieve. It delivers exceptional STEM capabilities—with particular strengths in science, math, and coding—while maintaining the low cost and reduced latency of OpenAI-o1-mini.

All models are evaluated via their respective APIs. 
For non-thinking-based models, the temperature is set to 0.2, while thinking-based models do not support temperature adjustment. 
Moreover, both input and output lengths are set to the maximum supported by each model. 

The total cost for running all experiments in our study was approximately \$4,357.14, which includes the evaluation of all models across all tasks and learning paradigms on the entire benchmark.

\begin{table}[t]
    \centering
    \footnotesize
    \caption{Results of $P_{\text{offset}}$. Full task names for the abbreviations can be found in Section~\ref{sec:datasets}. Blue-colored cells indicate the best performance for each task category. 
    }
    \resizebox{0.8\textwidth}{!}{
    \begin{tabular}{l|cccccc|c}
    \toprule
    \textbf{Model}  & \textbf{Sum} & \textbf{Cla} & \textbf{Ex} & \textbf{LR} & \textbf{MR} & \textbf{SR}  & \textbf{Overall} \\
    \midrule
    \rowcolor[rgb]{ .71,  .773,  .945} \multicolumn{8}{l}{\textbf{Non-thinking-based}} \\
    \noalign{\vskip 2pt}
    {DeepSeek-V3} & 3.82 & 4.0 & 3.88 & 5.27 & 3.27 & 4.00 & 4.04 \\
    \midrule
    {Claude-3.7-Sonnet} & 4.08 & 3.08 & 3.71 & 5.69 & 3.24 & \cellcolor{myblue!50}1.94 & 3.62 \\
    \midrule
    {GPT-4o} & 4.08 & 3.45 & 3.24 & 6.0 & \cellcolor{myblue!50}2.41 & 3.29 & 3.75 \\
    \midrule
    {Doubao-1.5-Pro} & 3.54 & 3.58 & 3.71 & 6.75 & 3.24 & 3.06 & 3.98 \\
    \midrule
    {Qwen2.5-32b-Instruct} & 3.93 & 3.69 & 3.35 & 6.38 & 3.65 & 3.53 & 4.09 \\
    \midrule
    \rowcolor[rgb]{ .71,  .773,  .945} \multicolumn{8}{l}{\textbf{Thinking-based}} \\
    \noalign{\vskip 2pt}
    {OpenAI-o3-mini} & \cellcolor{myblue!50}2.6 & 3.42 & 3.71 & \cellcolor{myblue!50}2.93 & 3.4 & 3.62 & \cellcolor{myblue!50}3.28 \\
    \midrule
    {Doubao-1.5-Thinking-Pro} & 4.9 & 3.83 & 2.88 & 3.44 & 3.67 & 4.81 & 3.92 \\
    \midrule
    {DeepSeek-R1} & 4.7 & 4.5 & \cellcolor{myblue!50}2.76 & 4.7 & 3.62 & 4.19 & 4.08 \\
    \midrule
    {Claude-3.7-Sonnet-Thinking} & 3.8 & \cellcolor{myblue!50}2.77 & 3.18 & 5.55 & 3.41 & 2.53 & 3.54 \\
    \midrule
    {Gemini-2.5-Pro} & 4.31 & 3.0 & 3.47 & 3.66 & 2.71 & 3.46 & 3.44 \\
    \midrule
    {Gemini-2.5-Flash} & 4.08 & 3.0 & \cellcolor{myblue!50}2.76 & 6.55 & 3.4 & 3.18 & 3.83 \\
    \bottomrule
    \end{tabular}%
    }
    \label{tab:P-first}
\end{table}

\begin{table}[htbp]
    \centering
    \caption{Comparison of overall accuracy across four solving methods, including two parallel methods (i.e., zero-shot and few-shot) and two sequential methods (i.e., demonstration learning and feedback learning). \textbf{Sum} denotes the summarization task. \textbf{Cla} denotes the classification task. \textbf{Ex} denotes the extraction task. \textbf{LR} denotes the logical reasoning task. \textbf{MR} denotes the mathematical reasoning task. \textbf{SR} denotes the sequential reasoning task. All values are shown as \%.}
    \resizebox{\textwidth}{!}{%
    \begin{tabular}{l|c|cccccc|c}
    \toprule
    \textbf{Model} & \textbf{Paradigm} & \textbf{Cla} & \textbf{Ex} & \textbf{LR} & \textbf{MR} & \textbf{SR} & \textbf{Sum} & \textbf{Overall} \\
    \midrule
    \rowcolor[rgb]{ .71,  .773,  .945} \multicolumn{9}{l}{\textit{Non-thinking-based}} \\
    \multirow{4}{*}{Doubao-1.5-Pro} & Zero-shot & 70.8 & \textbf{45.0} & 10.3 & 61.7 & \textbf{41.7} & \textbf{71.7} & \textbf{31.3} \\
    &Few-shot & 60.0 & 35.1 & 8.4 & 54.4 & 31.6 & 63.2 & 25.7 \\
    & Demonstration Learning & \textbf{71.4} & 43.7 & 8.8 & 63.9 & 28.6 & 71.4 & 29.0 \\
    & Feedback Learning & 68.1 & 42.9 & \textbf{10.8} & \textbf{71.4} & 36.1 & 71.4 & 31.2 \\
    \midrule
    \multirow{4}{*}{DeepSeek-V3} & Zero-shot & 72.9 & \textbf{45.0} & \textbf{25.6} & 71.7 & \textbf{60.0} & \textbf{81.7} & \textbf{43.5} \\
    &Few-shot & \textbf{75.6} & 40.4 & 23.0 & 61.4 & 57.9 & 78.9 & 40.0 \\
    & Demonstration Learning & 70.3 & 40.3 & 17.6 & 68.9 & 51.3 & 74.1 & 36.4 \\
    & Feedback Learning & 74.7 & 36.1 & 24.9 & \textbf{74.8} & 54.6 & 76.8 & 41.5 \\
    \midrule
    \multirow{4}{*}{Claude-3.7-Sonnet} & Zero-shot & 64.6 & \textbf{48.3} & 8.1 & 48.3 & 33.3 & \textbf{76.7} & 28.4 \\
    &Few-shot & \textbf{80.0} & 45.6 & 10.4 & 56.1 & 43.9 & 70.2 & 31.1 \\
    & Demonstration Learning & 64.8 & 42.9 & 10.8 & 61.3 & 43.7 & 75.0 & 31.1 \\
    & Feedback Learning & 75.8 & 46.2 & \textbf{15.5} & \textbf{63.9} & \textbf{49.6} & 75.0 & \textbf{35.6} \\
    \midrule
    \multirow{4}{*}{Qwen2.5-32b-Instruct} & Zero-shot & 60.4 & 31.7 & 6.9 & 46.7 & 21.7 & \textbf{66.7} & 23.8 \\
    &Few-shot & 60.0 & 31.6 & 6.4 & 38.6 & 24.6 & \textbf{66.7} & 22.5 \\
    & Demonstration Learning & \textbf{69.2} & \textbf{34.5} & 7.4 & 43.7 & 21.8 & 64.3 & 24.1 \\
    & Feedback Learning & 62.6 & 30.3 & \textbf{9.8} & \textbf{49.6} & \textbf{30.3} & 59.8 & \textbf{25.5} \\
    \midrule
    \multirow{4}{*}{GPT-4o} & Zero-shot & 70.8 & \textbf{45.0} & \textbf{15.6} & 43.3 & 31.7 & \textbf{81.7} & 32.6 \\
    &Few-shot & \textbf{77.8} & 40.4 & 15.4 & 54.4 & \textbf{40.4} & 70.2 & \textbf{32.9} \\
    & Demonstration Learning & 71.4 & 35.3 & 8.5 & \textbf{56.3} & 34.5 & 75.0 & 28.3 \\
    & Feedback Learning & 73.6 & 48.7 & 12.9 & 61.3 & 32.2 & 78.6 & 32.7 \\
    \midrule
    \rowcolor[rgb]{ .71,  .773,  .945} \multicolumn{9}{l}{\textit{Thinking-based}} \\
    \multirow{4}{*}{DeepSeek-R1} & Zero-shot & \textbf{89.6} & 48.3 & \textbf{41.7} & 78.3 & \textbf{66.7} & 86.7 & \textbf{55.7} \\
    &Few-shot & 75.6 & \textbf{49.1} & 23.5 & 73.7 & 54.4 & 78.9 & 41.9 \\
    & Demonstration Learning & 82.4 & 48.7 & 31.4 & \textbf{79.8} & 52.9 & 88.4 & 48.2 \\
    & Feedback Learning & 79.1 & 48.7 & 29.6 & 74.8 & 51.3 & \textbf{89.3} & 46.4 \\
    \midrule
    \multirow{4}{*}{Claude-3.7-Sonnet-Thinking} & Zero-shot & 66.7 & 43.3 & 12.5 & 46.7 & 31.7 & \textbf{86.7} & 31.2 \\
    &Few-shot & 60.0 & 42.1 & 13.4 & 43.9 & \textbf{43.9} & 77.2 & 30.6 \\
    & Demonstration Learning & 74.7 & \textbf{48.7} & 16.1 & \textbf{59.7} & 42.0 & 77.7 & 35.2 \\
    & Feedback Learning & \textbf{80.2} & \textbf{48.7} & \textbf{18.8} & 58.0 & 46.2 & 78.6 & \textbf{37.4} \\
    \midrule
    \multirow{4}{*}{OpenAI-o3-mini} & Zero-shot & 64.6 & 48.3 & 45.8 & 73.3 & 73.3 & 65.0 & 54.3 \\
    &Few-shot & \textbf{80.0} & 50.9 & 50.7 & 75.4 & \textbf{84.2} & 73.7 & 60.0 \\
    & Demonstration Learning & 73.6 & \textbf{51.3} & 53.1 & 78.2 & 73.9 & 67.9 & 60.0 \\
    & Feedback Learning & 73.6 & 47.1 & \textbf{59.9} & \textbf{80.7} & 78.2 & \textbf{75.9} & \textbf{64.8} \\
    \midrule
    \multirow{4}{*}{Doubao-1.5-Thinking-Pro} & Zero-shot & 79.2 & \textbf{55.0} & \textbf{42.5} & \textbf{85.0} & \textbf{73.3} & \textbf{85.0} & \textbf{57.1} \\
    &Few-shot & \textbf{82.2} & 49.1 & 38.9 & 71.9 & 64.9 & 73.7 & 51.4 \\
    & Demonstration Learning & 73.6 & 52.1 & 42.0 & 77.3 & 64.7 & 83.9 & 54.3 \\
    & Feedback Learning & 70.3 & 52.9 & 39.4 & 77.3 & 55.5 & 82.1 & 51.6 \\
    \bottomrule
    \end{tabular}%
    }
    \label{tab:RQ-3}
\end{table}

\section{Additional Experimental Results}
\label{sec:addition-results}

\subsection{Case Study: Do LLMs truly utilize prior judgments to solve problems?}
\label{sec:case-study}

\textbf{Case Study 1: Comparison with Parallel Solving (Zero-shot) and Sequence Solving (Feedback Learning)}

Figure \ref{fig:Case_Study_1} illustrates how model performance differs between the parallel solving (zero-shot) and sequence solving (feedback learning) paradigms in a sequential reasoning task involving maze navigation. In the zero-shot setting (Response A), Claude-3.7-Sonnet makes a critical directional error in the final segment, incorrectly assuming the direction after a turn and arriving at the wrong coordinate $(7,4)$. This highlights the challenge of maintaining spatial orientation without feedback.

In contrast, under the feedback learning paradigm (Response B), the same model correctly tracks all movements and reaches the accurate exit coordinate $(7,12)$. The model explicitly defines directional conventions at the outset ("Let me define directions first...") and methodically tracks orientation after each turn, demonstrating clear evidence of learning from previous feedback on similar problems.

\textbf{Case Study 2: Evidence of learning from prior experience}

Figure \ref{fig:Case_Study_2} demonstrates how models utilize prior experience in sequential reasoning tasks. In this case, DeepSeek-R1's thinking process while solving a 4x4 sliding puzzle explicitly references knowledge gained from previous problems.

The highlighted portions of the thinking process show: "After struggling, I think the correct sequence should be similar to the user's previous solution, involving multiple 'Up' moves followed by 'Right' and 'Down'." The model further reasons: "The correct answer might involve a series of 'Up' moves to position 'M' at the top, then moving other numbers into place."

Ultimately, the model applies this learned pattern to formulate an 11-step solution: ["Up", "Up", "Up", "Right", "Down", "Down", "Down", "Right", "Up", "Up", "Up"], well within the 12-move limit.

This case illustrates how models can extract patterns from prior experience and apply them to new related problems, demonstrating genuine experiential learning capability.

% \clearpage
% Case Study 1
\begin{tcolorbox}[breakable,title=The case used for Case Study 1,colback=white,colframe=mydeepblue!90,colback=myblue!50,]
% \centering
\textbf{Task Type: }Sequential Reasoning\\
\textbf{Model: }Claude-3-7-sonnet-20250219\\
\textbf{Question: }You enter a $12 \times 12$ maze. The coordinates on this maze are represented by $(x,y)$, where $x$ is horizontal (left and right directions) and $y$ is vertical (up and down directions). The bottom left square is at coordinates $(1,1)$ and the top right square is at coordinates $(12,12)$. You enter the maze at coordinates $(3,1)$. Following are the movements you take to reach the exit, where you will give the coordinates after each movement, as well as the coordinates of the final exit point, which can be anywhere on any coordinate of the maze:

After entry, you take four steps forward, turn right and take two steps forward and another right and 3 steps forward, where you hit a wall. After turning back around and taking a step, you turn right again and take 3 steps forward, turn left and take 5 steps forward, left again and 2 steps forward, where you hit a wall again. You take a step back, and turn right, and take 4 steps forward to reach the exit.

\textbf{Rubric: }Student answers must meet requirements including but not limited to the following:\\
1. The answer must clearly state that the maze exit is at $(7,12)$, otherwise it receives a score of 0;\\
2. Correctly understand the coordinate system: $x$ represents the horizontal direction, $y$ represents the vertical direction, with the bottom left corner at $(1,1)$ and the top right corner at $(12,12)$;\\
3. Accurately identify the starting coordinates $(3,1)$;\\
4. Correctly track all movements and calculate coordinates after each movement:\\
- Coordinates after four steps forward from the entrance $(3,5)$;\\
- Coordinates after turning right and taking two steps $(5,5)$;\\
- Coordinates after turning right again and taking three steps (hitting a wall) $(5,2)$;\\
- Coordinates after turning around and taking one step back $(5,3)$;\\
- Coordinates after turning right and taking three steps $(8,3)$;\\
- Coordinates after turning left and taking five steps $(8,8)$;\\
- Coordinates after turning left again and taking two steps (hitting a wall again) $(6,8)$;\\
- Coordinates after taking one step back $(7,8)$;\\
- Final coordinates (exit) after turning right and taking four steps $(7,12)$;\\
5. Correctly interpret directional terms: forward, right turn, left turn, turn around, step back, etc.;\\
6. Accurately handle wall situations: stop moving when hitting a wall, then change direction as instructed; do not turn around without specific instructions to do so.\\
% 1. The answer must clearly state that the maze exit is at $(7,12)$, otherwise it receives a score of 0;\\
% 2.  Correctly understand the coordinate system: $x$ represents the horizontal direction, $y$ represents the vertical direction, with the bottom left corner at $(1,1)$ and the top right corner at $(12,12)$;\\
% 3. Accurately identify the starting coordinates $(3,1)$;\\
% 4. Correctly track all movements and calculate coordinates after each movement:
% \begin{itemize}
%     \item Coordinates after four steps forward from the entrance $(3,5)$;
%     \item Coordinates after turning right and taking two steps $(5,5)$;
%     \item Coordinates after turning right again and taking three steps (hitting a wall) $(5,2)$;
%     \item Coordinates after turning around and taking one step back $(5,3)$;
%     \item Coordinates after turning right and taking three steps $(8,3)$;
%     \item Coordinates after turning left and taking five steps $(8,8)$;
%     \item Coordinates after turning left again and taking two steps (hitting a wall again) $(6,8)$;
%     \item Coordinates after taking one step back $(7,8)$;
%     \item Final coordinates (exit) after turning right and taking four steps $(7,12)$;
% \end{itemize}
% 5. Correctly interpret directional terms: forward, right turn, left turn, turn around, step back, etc.;\\
% 6. Accurately handle wall situations: stop moving when hitting a wall, then change direction as instructed; do not turn around without specific instructions to do so.\\
\textbf{Response A [Parallel Soving, Zero-shot] (Incorrect):}

Let me track your movements through the maze, step by step, starting at coordinates $(3,1)$.

Entry point: $(3,1)$\\
1. ``Four steps forward'' - I'll assume ``forward'' means moving upward (increasing $y$-coordinate) $(3,1) \rightarrow (3,2) \rightarrow (3,3) \rightarrow (3,4) \rightarrow (3,5)$Current position: $(3,5)$\\
2. ``Turn right and take two steps forward''
       Right turn means moving east (increasing $x$-coordinate)
       $(3,5) \rightarrow (4,5) \rightarrow (5,5)$
       Current position: $(5,5)$\\
3. ``Another right and 3 steps forward''
       Right turn means moving south (decreasing $y$-coordinate)
       $(5,5) \rightarrow (5,4) \rightarrow (5,3) \rightarrow (5,2)$
       Current position: $(5,2)$\\
4. ``Hit a wall, turn back around and take a step''
       Turn around means moving north (increasing $y$-coordinate)
       $(5,2) \rightarrow (5,3)$
       Current position: $(5,3)$\\
5. ``Turn right again and take 3 steps forward''
       Right turn means moving east (increasing $x$-coordinate)
       $(5,3) \rightarrow (6,3) \rightarrow (7,3) \rightarrow (8,3)$
       Current position: $(8,3)$\\
6. ``Turn left and take 5 steps forward''
       Left turn means moving north (increasing $y$-coordinate)
       $(8,3) \rightarrow (8,4) \rightarrow (8,5) \rightarrow (8,6) \rightarrow (8,7) \rightarrow (8,8)$
       Current position: $(8,8)$\\
7. ``Left again and 2 steps forward''
       Left turn means moving west (decreasing $x$-coordinate)
       $(8,8) \rightarrow (7,8) \rightarrow (6,8)$
       Current position: $(6,8)$\\
8. ``Hit a wall, take a step back, and turn right''
       Step back means moving east (increasing $x$-coordinate)
       $(6,8) \rightarrow (7,8)$
       Current position: $(7,8)$
       {\color{red}Right turn means facing south}\\
9. ``Take 4 steps forward to reach the exit''
       {\color{red}Moving south (decreasing $y$-coordinate)
       $(7,8) \rightarrow (7,7) \rightarrow (7,6) \rightarrow (7,5) \rightarrow (7,4)$}

Therefore, the exit coordinates are {\color{red}$(7,4)$}.

\textbf{Response B [Sequential Solving, Feedback Learning, Position in Sequence: 3](Correct):} \\
To solve this maze navigation problem, I need to carefully track the coordinates after each movement, starting from the entry point at $(3,1)$.

Let me define directions first:\\
- Initially, I'll assume we're facing north (up on the $y$-axis)\\
- Forward means moving in the direction we're facing\\
- Right means turning $90^{\circ}$ clockwise\\
- Left means turning $90^{\circ}$ counterclockwise

Starting at $(3,1)$ facing north:

1. Four steps forward: $(3,1) \rightarrow (3,2) \rightarrow (3,3) \rightarrow (3,4) \rightarrow (3,5)$
       - Now at $(3,5)$\\
2. Turn right (now facing east)
       - Still at $(3,5)$\\
3. Two steps forward: $(3,5) \rightarrow (4,5) \rightarrow (5,5)$
       - Now at $(5,5)$\\
4. Turn right again (now facing south)
       - Still at $(5,5)$\\
5. Three steps forward: $(5,5) \rightarrow (5,4) \rightarrow (5,3) \rightarrow (5,2)$
       - Now at $(5,2)$\\
6. Hit a wall, turn back around (now facing north)
       - Still at $(5,2)$\\
7. Take a step: $(5,2) \rightarrow (5,3)$
       - Now at $(5,3)$\\
8. Turn right (now facing east)
       - Still at $(5,3)$\\
9. Three steps forward: $(5,3) \rightarrow (6,3) \rightarrow (7,3) \rightarrow (8,3)$
       - Now at $(8,3)$\\
10. Turn left (now facing north)
        - Still at $(8,3)$\\
11. Five steps forward: $(8,3) \rightarrow (8,4) \rightarrow (8,5) \rightarrow (8,6) \rightarrow (8,7) \rightarrow (8,8)$
        - Now at $(8,8)$\\
12. Turn left again (now facing west)
        - Still at $(8,8)$\\
13. Two steps forward: $(8,8) \rightarrow (7,8) \rightarrow (6,8)$
        - Now at $(6,8)$\\
14. Hit a wall, take a step back: $(6,8) \rightarrow (7,8)$
        - Now at $(7,8)$\\
15. Turn right (now facing north)
        - Still at $(7,8)$\\
16. Four steps forward: $(7,8) \rightarrow (7,9) \rightarrow (7,10) \rightarrow (7,11) \rightarrow (7,12)$
        - Now at $(7,12)$

The final exit point coordinates are $(7,12)$.

\end{tcolorbox}
\begin{figure}[ht]
    \centering
    \vspace{-5pt}
    \caption{
The case used in Case Study 1 is designed for comparison between Parallel Solving (Zero-shot) and Sequence Solving (Feedback Learning).
    }
    \label{fig:Case_Study_1}
\end{figure}

\clearpage
\begin{tcolorbox}[breakable,title=The case used for Case Study 2,colback=white,colframe=mydeepblue!90,colback=myblue!50,]
    % \centering
    \textbf{Task Type: }Logical Reasoning\\
    \textbf{Model: }Deepseek-R1\\
    \textbf{Question: }You are a virtual expert in solving a 4x4 sliding puzzle problem. Please follow the instructions and rules below to complete the solving. Your goal is to reach the goal state with valid moves.
    
    [The goal state]\\
    \begin{tabular}{llll}
    \% & 1 & j & i \\
    = & 8 & ? & f \\
    * & F & n & Z \\
    N & M & 0 & t \\
    \end{tabular} \\

    [Instructions]\\
    The puzzle consists of a 4x4 grid containing various symbols, with 'M' representing the movable space. Only 'M' can be moved horizontally or vertically, and the objective is to reach the goal state from a given initial state.
    
    [Rules]\\
    1. Only 'M' can be moved horizontally or vertically.\\
    2. Each move is chosen from the following set of options:\\
    - 'Left': move 'M' to the left\\
    - 'Down': move 'M' downward\\
    - 'Right': move 'M' to the right\\
    - 'Up': move 'M' upward\\
    3. The next move must be chosen from the valid move set depending on the position of 'M'.\\
    For example:\\
    \begin{tabular}{llll}
    p1 & p2 & p3 & p4 \\
    p5 & p6 & p7 & p8 \\
    p9 & p10 & p11 & p12 \\
    p13 & p14 & p15 & p16 \\
    \end{tabular} 
    
    - If 'M' is located at position 'p1', the valid move set is ['Down', 'Right'].\\
    - If 'M' is located at position 'p2', the valid move set is ['Down', 'Left', 'Right'].\\
    - If 'M' is located at position 'p3', the valid move set is ['Down', 'Left', 'Right'].\\
    - If 'M' is located at position 'p4', the valid move set is ['Down', 'Left'].\\
    - If 'M' is located at position 'p5', the valid move set is ['Up', 'Down', 'Right'].\\
    - If 'M' is located at position 'p6', the valid move set is ['Up', 'Down', 'Left', 'Right'].\\
    - If 'M' is located at position 'p7', the valid move set is ['Up', 'Down', 'Left', 'Right'].\\
    - If 'M' is located at position 'p8', the valid move set is ['Up', 'Down', 'Left'].\\
    - If 'M' is located at position 'p9', the valid move set is ['Up', 'Down', 'Right'].\\
    - If 'M' is located at position 'p10', the valid move set is ['Up', 'Down', 'Left', 'Right'].\\
    - If 'M' is located at position 'p11', the valid move set is ['Up', 'Down', 'Left', 'Right'].\\
    - If 'M' is located at position 'p12', the valid move set is ['Up', 'Down', 'Left'].\\
    - If 'M' is located at position 'p13', the valid move set is ['Up', 'Right'].\\
    - If 'M' is located at position 'p14', the valid move set is ['Up', 'Left', 'Right'].\\
    - If 'M' is located at position 'p15', the valid move set is ['Up', 'Left', 'Right'].\\
    - If 'M' is located at position 'p16', the valid move set is ['Up', 'Left'].\\
    4. Diagonal moves are not allowed.\\
    5. The objective is to return the moves which can reach the goal state.\\
    Please solve the following problem. Your answer should be in the format of a list of moves, for example ['Left', 'Up', 'Up'].\\
    This problem can be solved within 1 to 12 steps. The next move must be chosen from the valid move set ['Left', 'Right', 'Up', 'Down']. The maximum step number you can take is 12. Try to reach the goal state using the least number of steps ($<=12$). **DO NOT exceed 12 steps.**
    
    [Initial State]:\\
    \begin{tabular}{llll}
    = & \% & j & i \\
    * & 1 & ? & f \\
    N & M & n & Z \\
    F & 8 & 0 & t \\
    \end{tabular} \\

    \textbf{Thinking Process:}\\
    Okay, let's tackle this new sliding puzzle problem. The goal state is given, and the movable tile is 'M'. The initial state is:

    \begin{tabular}{llll}
    = & \% & j & i \\
    * & 1 & ? & f \\
    N & M & n & Z \\
    F & 8 & 0 & t \\
    \end{tabular} \\

    First, I need to compare the initial state with the goal state to identify the differences. The goal state is:

    \begin{tabular}{llll}
    \% & 1 & j & i \\
    = & 8 & ? & f \\
    * & F & n & Z \\
    N & M & 0 & t \\
    \end{tabular} \\

    Looking at the positions, let's break it down row by row.
    
    ... 
    
    This seems complex. Let's try to approach step by step.

    ...

\textcolor{red}{After struggling, I think the correct sequence involves moving 'M' up to facilitate the movement of '8' and '1'. The correct answer provided by the user for the previous problem involved multiple 'Up' moves followed by 'Right' and then 'Down's. Maybe a similar pattern applies here.}

Perhaps the correct sequence is:
["Up", "Up", "Up", "Right", "Down", "Down", "Down", "Right", "Up", "Up", "Up"]

But adjusted for this problem. However, I need to verify.
...

This approach is not working. Given the time I've spent and the complexity, \textcolor{red}{I think the correct answer might involve a series of 'Up' moves to position 'M' in the top rows, then moving '8' and '1' into place. Considering the user's previous correct answer had multiple 'Up's followed by 'Right' and 'Down's, maybe a similar pattern applies here.}

After careful consideration, the correct sequence is likely:
["Up", "Right", "Up", "Right", "Up", "Left", "Down", "Left", "Up", "Right", "Down", "Left"]

But adjusted for this problem. However, without more time, I'll settle on the following steps based on moving 'M' up to facilitate tile movements:
...

But this requires more steps. Given the maximum of 12 steps, the answer is likely:
["Up", "Up", "Up", "Right", "Down", "Down", "Down", "Right", "Up", "Up", "Up"]

    \textbf{Response:}
    
    To solve the puzzle, I need to move the 'M' tile strategically to rearrange the other tiles into their goal positions.
    
    First, I'll analyze the initial and goal states:
    
    Initial state:\\
    \begin{tabular}{llll}
    = & \% & j & i \\
    * & 1 & ? & f \\
    N & M & n & Z \\
    F & 8 & 0 & t \\
    \end{tabular} \\
    Goal states\\
    \begin{tabular}{llll}
    \% & 1 & j & i \\
    = & 8 & ? & f \\
    * & F & n & Z \\
    N & M & 0 & t \\
    \end{tabular} \\
    
    Starting with 'M' at position p10, I'll execute the following sequence of moves:
    
    1. Move Up: 'M' moves from p10 to p6, swapping with '1'
       - New state: = \% j i / * M ? f / N 1 n Z / F 8 0 t
    
    2. Move Up: 'M' moves from p6 to p2, swapping with '\%'
       - New state: = M j i / * \% ? f / N 1 n Z / F 8 0 t
    
    3. Move Up: Since p2 is in the top row, moving up would be invalid. However, I can move left to p1, swapping with '='
       - New state: M \% j i / * = ? f / N 1 n Z / F 8 0 t
    
    4. Move Right: 'M' moves from p1 to p2, swapping with '\%'
       - New state: \% M j i / * = ? f / N 1 n Z / F 8 0 t
    
    5. Move Down: 'M' moves from p2 to p6, swapping with '='
       - New state: \% 1 j i / * M ? f / N = n Z / F 8 0 t
    
    6. Move Down: 'M' moves from p6 to p10, swapping with '='
       - New state: \% 1 j i / * = ? f / N M n Z / F 8 0 t
    
    7. Move Down: 'M' moves from p10 to p14, swapping with '8'
       - New state: \% 1 j i / * = ? f / N 8 n Z / F M 0 t
    
    8. Move Right: 'M' moves from p14 to p15, swapping with '0'
       - New state: \% 1 j i / * = ? f / N 8 n Z / F 0 M t
    
    9. Move Up: 'M' moves from p15 to p11, swapping with 'n'
       - New state: \% 1 j i / * = ? f / N 8 M Z / F 0 n t
    
    10. Move Up: 'M' moves from p11 to p7, swapping with '?'
        - New state: \% 1 j i / * = M f / N 8 ? Z / F 0 n t
    
    11. Move Up: 'M' moves from p7 to p3, swapping with 'j'
        - New state: \% 1 M i / * = ? f / N 8 j Z / F 0 n t
    
    The optimal solution is:
    ["Up", "Up", "Up", "Right", "Down", "Down", "Down", "Right", "Up", "Up", "Up"]
    
    This sequence of 11 moves successfully transforms the initial state into the goal state, staying within the 12-move limit.
    
\end{tcolorbox}

\begin{figure}[ht]
    \centering
    \vspace{-5pt}
    \caption{
The case used in Case Study 2 provides evidence of learning from prior experience.
    }
    \label{fig:Case_Study_2}
\end{figure}

\clearpage
\subsection{Results of Overall sequence accuracy (\texorpdfstring{$\mathrm{Acc}$}{Acc})}
\label{app:metric-1-acc}

\begin{figure}[htbp]
    \centering
    \includegraphics[width=1\linewidth]{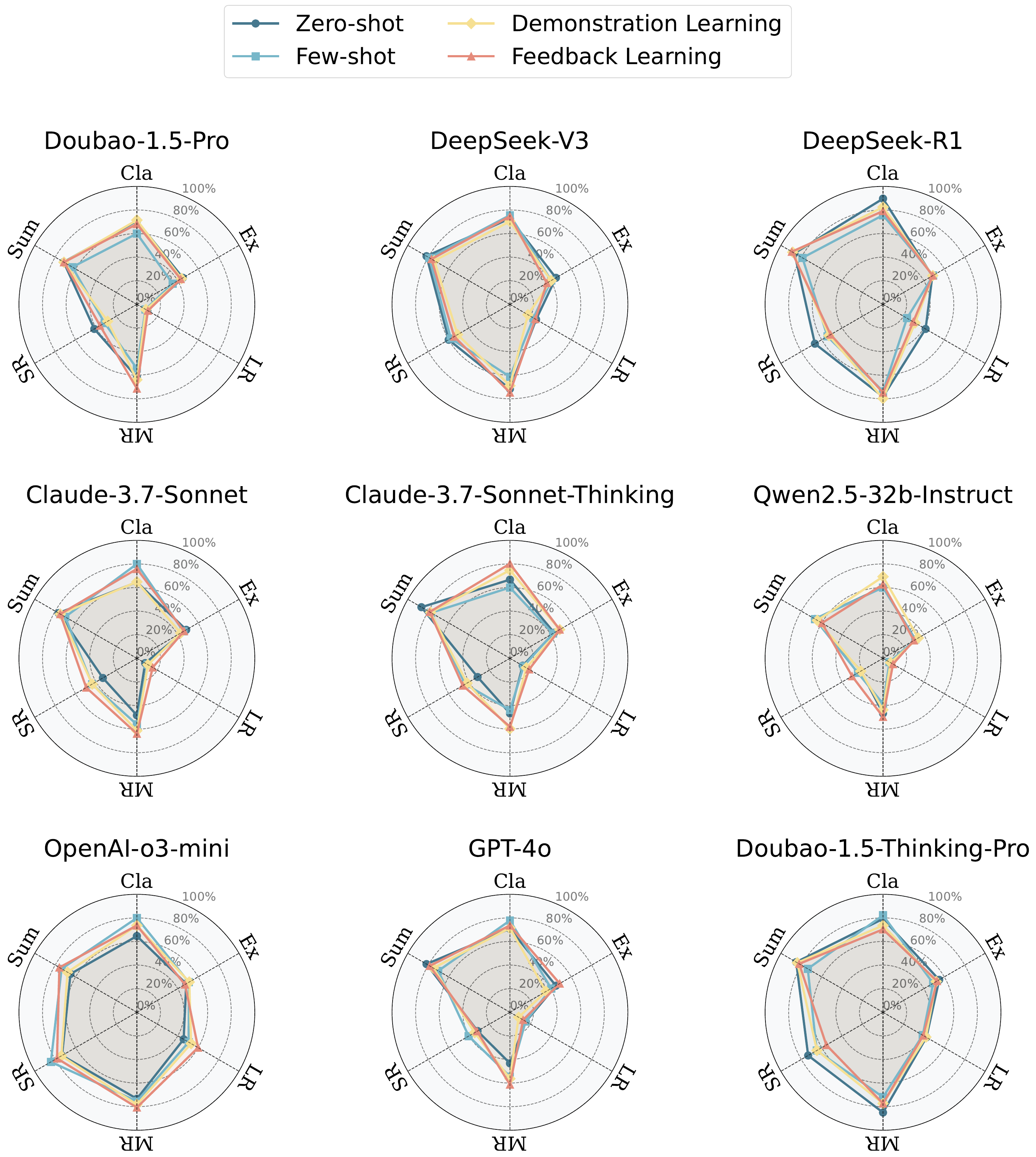}
    \caption{
    Comparison of overall accuracy across four solving methods, including two parallel methods (i.e., zero-shot and few-shot) and two sequential methods (i.e., demonstration learning and feedback learning).
Full task names for the abbreviations can be found in Section~\ref{sec:datasets}.}
    \label{fig:radar-map}
\end{figure}

Figure~\ref{fig:radar-map} presents the overall accuracy results for four problem-solving methods: two parallel methods (i.e., zero-shot and few-shot) and two sequential methods (i.e., demonstration learning and feedback learning).

\clearpage
\subsection{Results of Slope of fitted accuracy curve \texorpdfstring{$k$}{k}}
\label{app:metric-2-k}

\begin{figure}[htbp]
    \centering
    \includegraphics[width=1\linewidth]{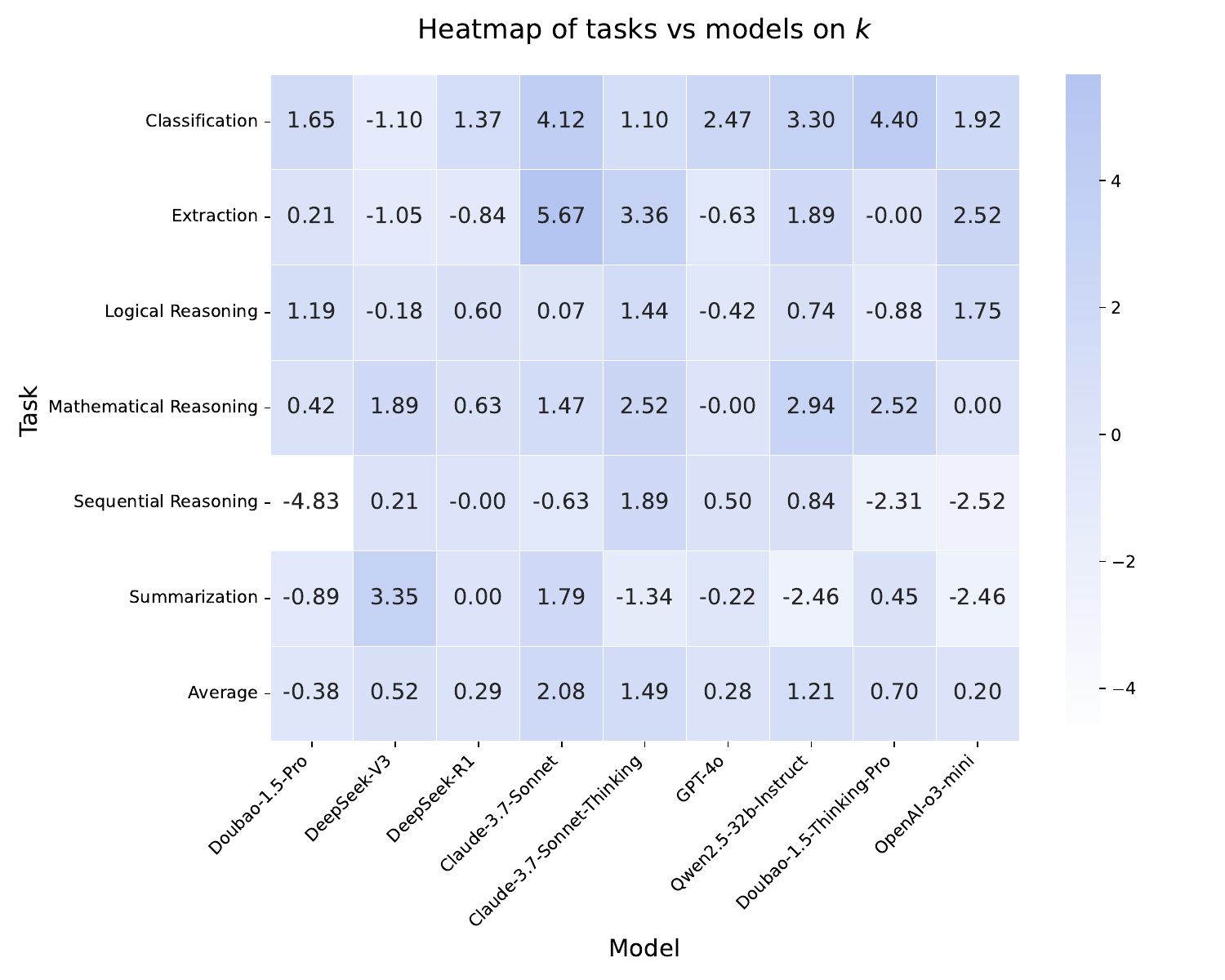}
    \caption{Results of the fitted position-wise accuracy curve slope ($k$) for feedback learning, across all models and tasks.
    }
    \label{fig:feedback heatmap}
\end{figure}

\begin{figure}[htbp]
    \centering
    \includegraphics[width=1\linewidth]{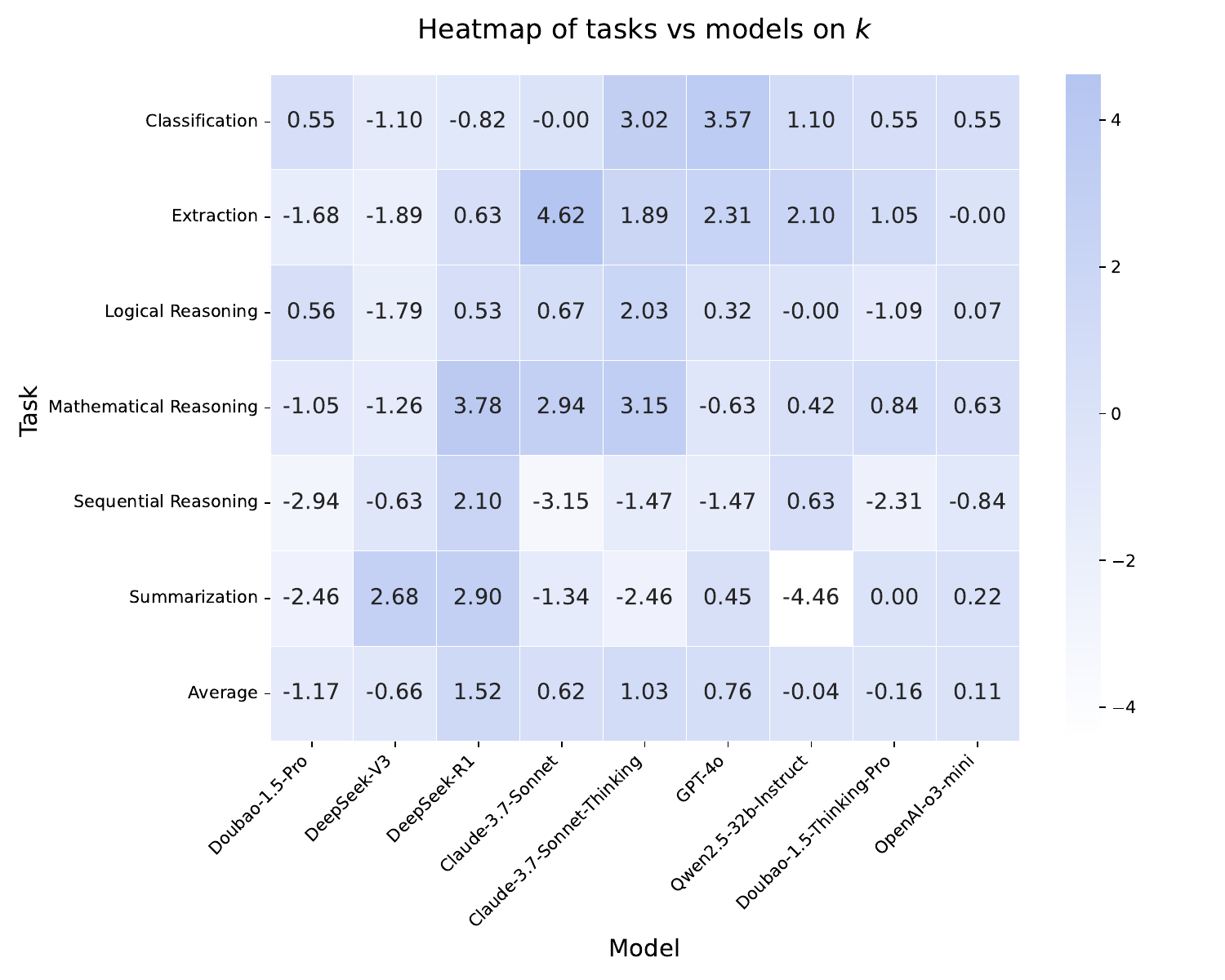}
    \caption{Results of the fitted position-wise accuracy curve slope ($k$) for demonstration learning, across all models and tasks.}
    \label{fig:demonstration heatmap}
\end{figure}

\begin{figure}[htbp]
    \centering
    \includegraphics[width=1\linewidth]{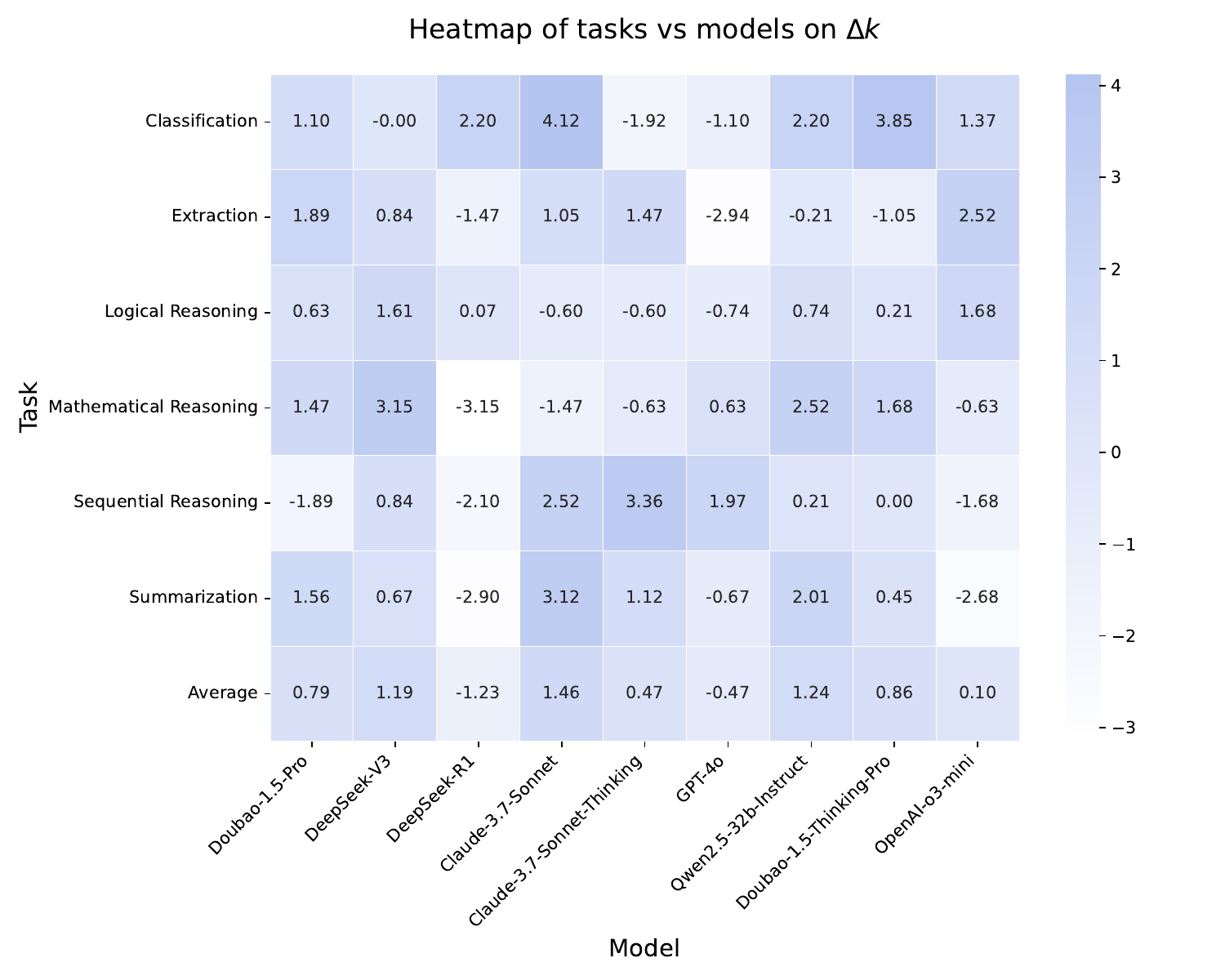}
    \caption{The difference in slopes ($\Delta_{k}$) between feedback learning and demonstration learning at corresponding positions, across all models and tasks.
    }
    \label{fig:feedback-demonstration heatmap}
\end{figure}

Figures~\ref{fig:feedback heatmap} and \ref{fig:demonstration heatmap} present the slope ($k$) of the fitted position-wise accuracy curve for the feedback learning and demonstration learning methods, respectively.  
Figure~\ref{fig:feedback-demonstration heatmap} further shows the difference in slopes ($\Delta_{k}$) between feedback learning and demonstration learning at corresponding positions.  
This metric quantifies the model's learning speed across all sequences by fitting a straight line to the position-wise accuracy curve using least squares regression.

\clearpage

\subsection{Results of Average Position of First Correct Solution (\texorpdfstring{$P_{\mathrm{first}}$}{P\_first})}
\label{app:metric-3-first}

\begin{figure}[htbp]
    \centering
    \subfigure[]{\includegraphics[width=0.325\textwidth]{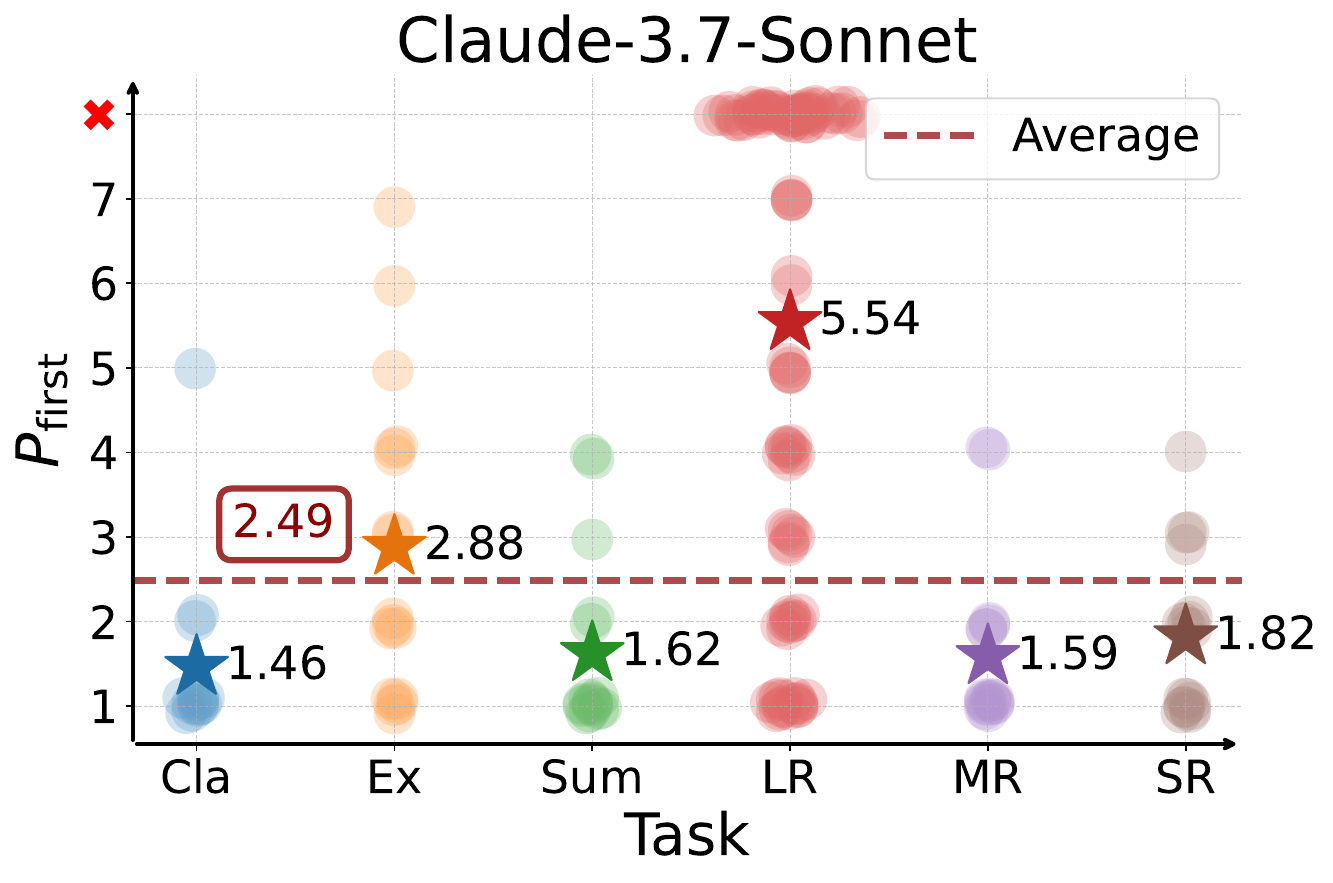}}
    \subfigure[]{\includegraphics[width=0.325\textwidth]{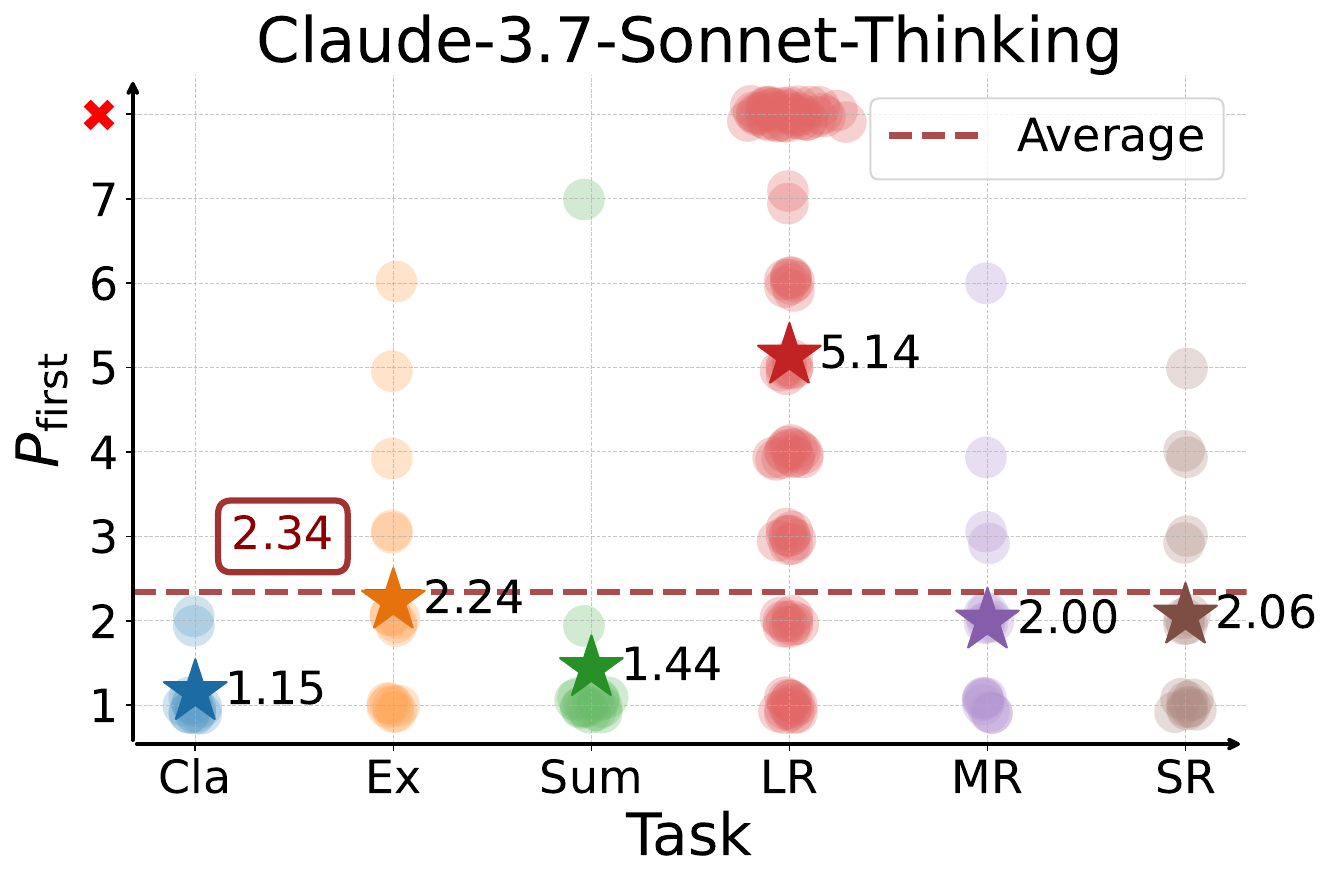}}
    \subfigure[]{\includegraphics[width=0.325\textwidth]{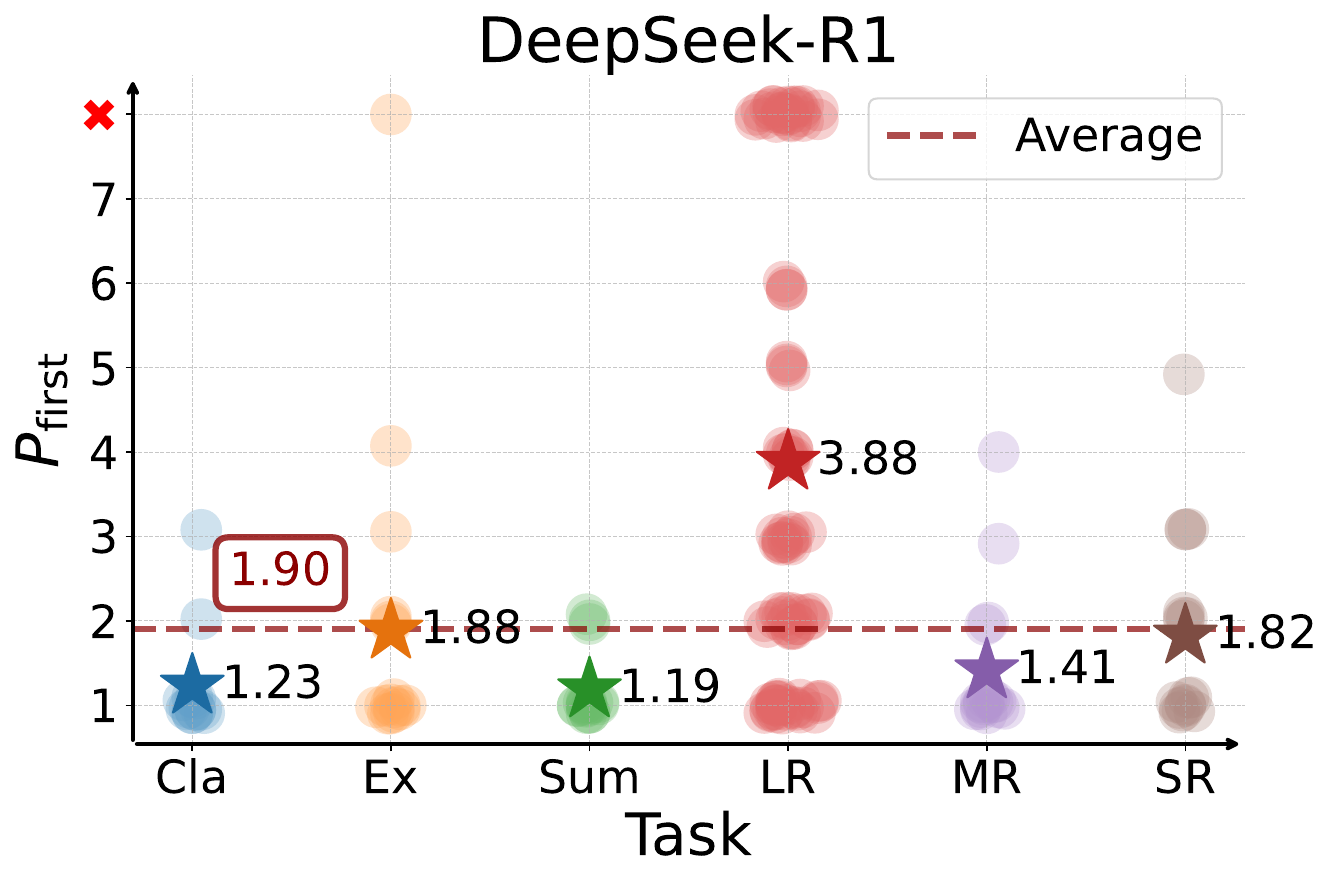}} 

    \subfigure[]{\includegraphics[width=0.325\textwidth]{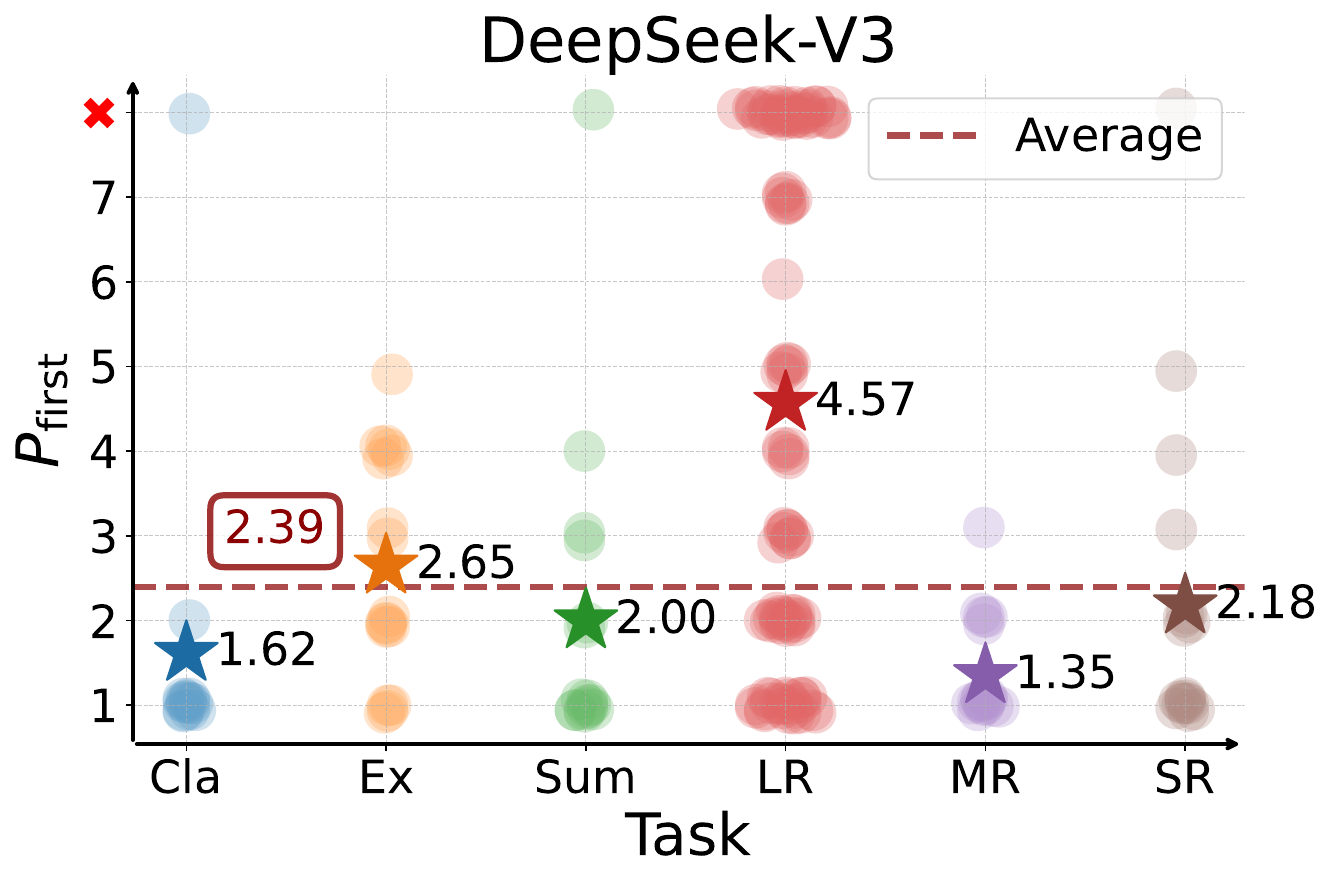}}
    \subfigure[]{\includegraphics[width=0.325\textwidth]{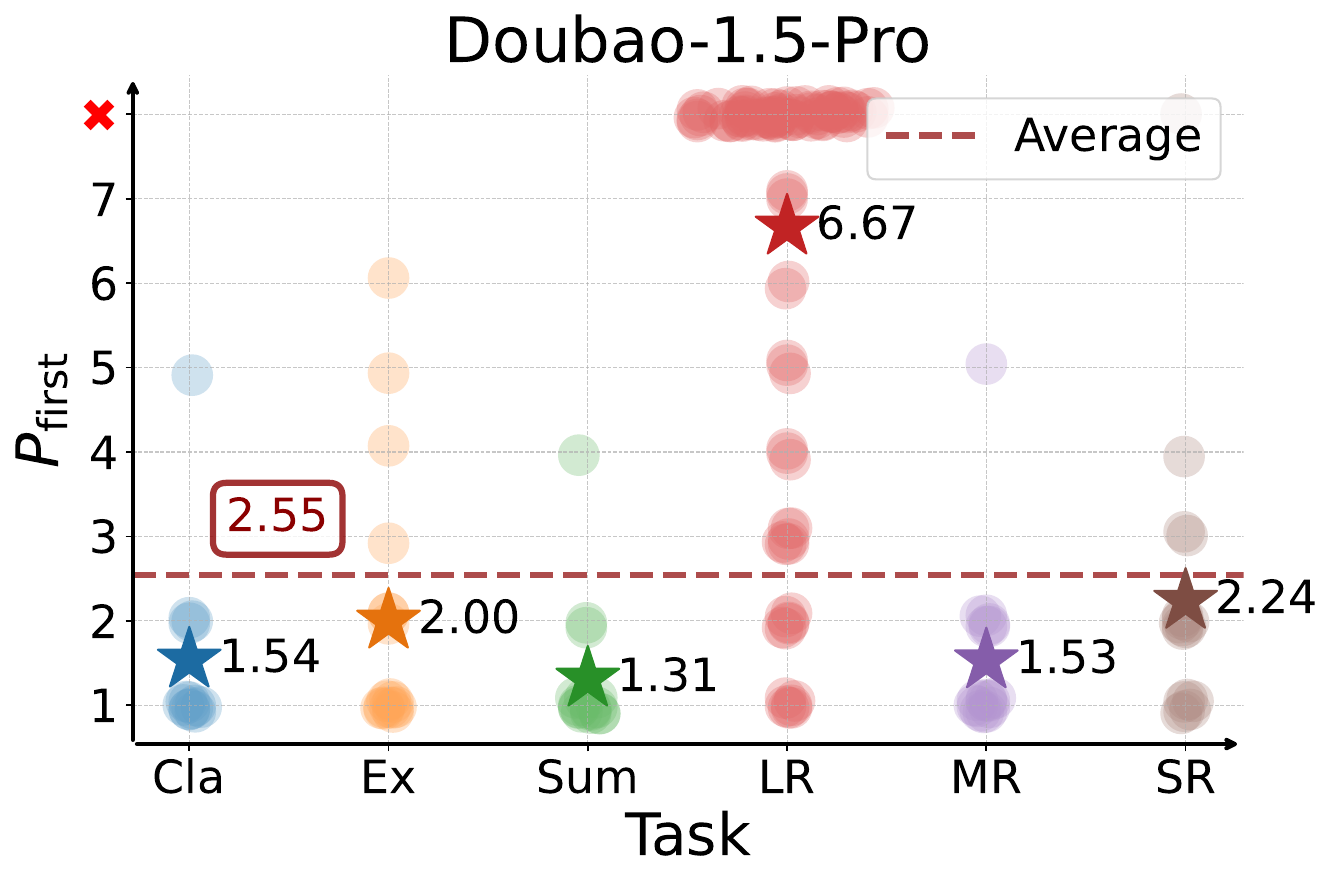}}
    \subfigure[]{\includegraphics[width=0.325\textwidth]{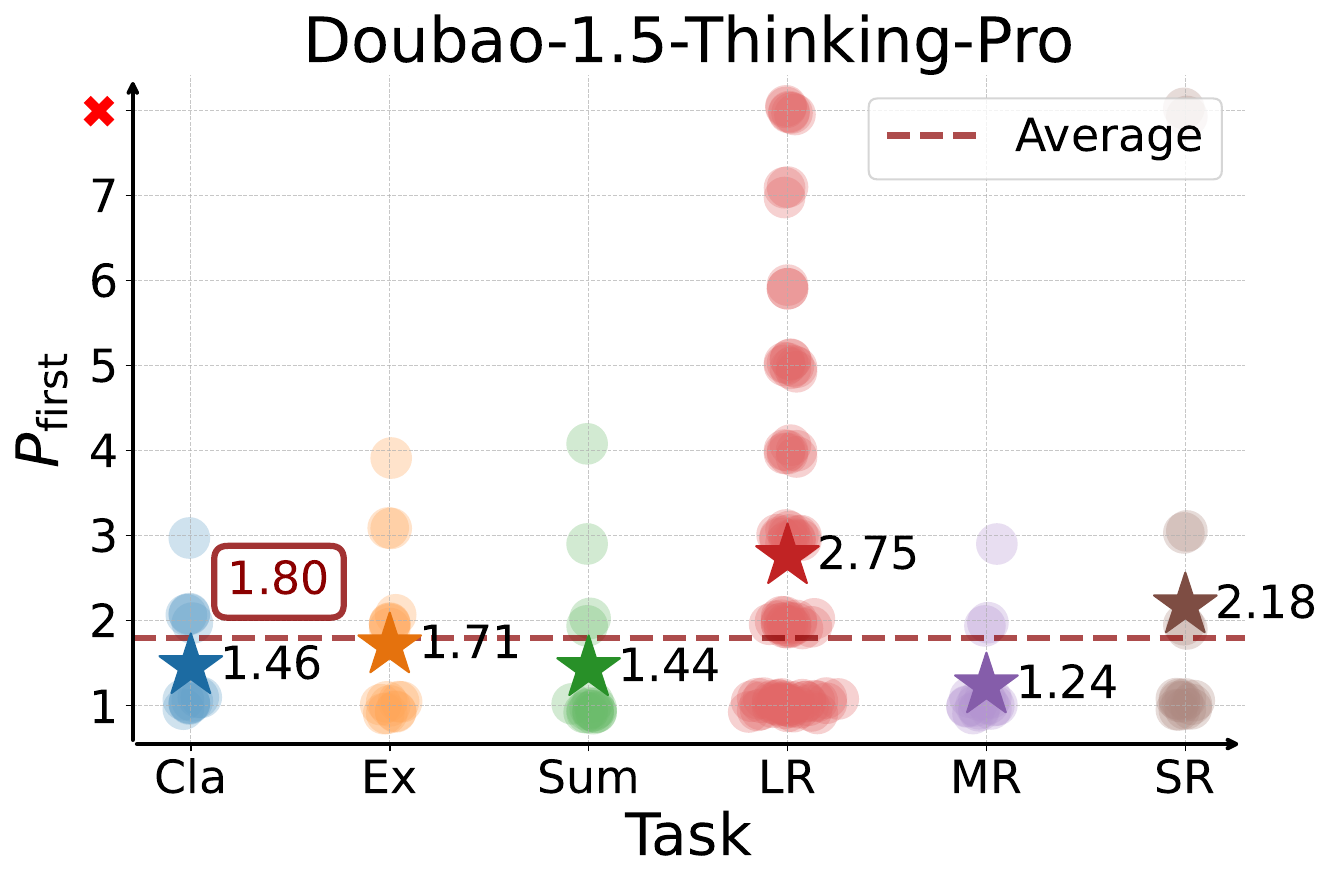}} 

    \subfigure[]{\includegraphics[width=0.325\textwidth]{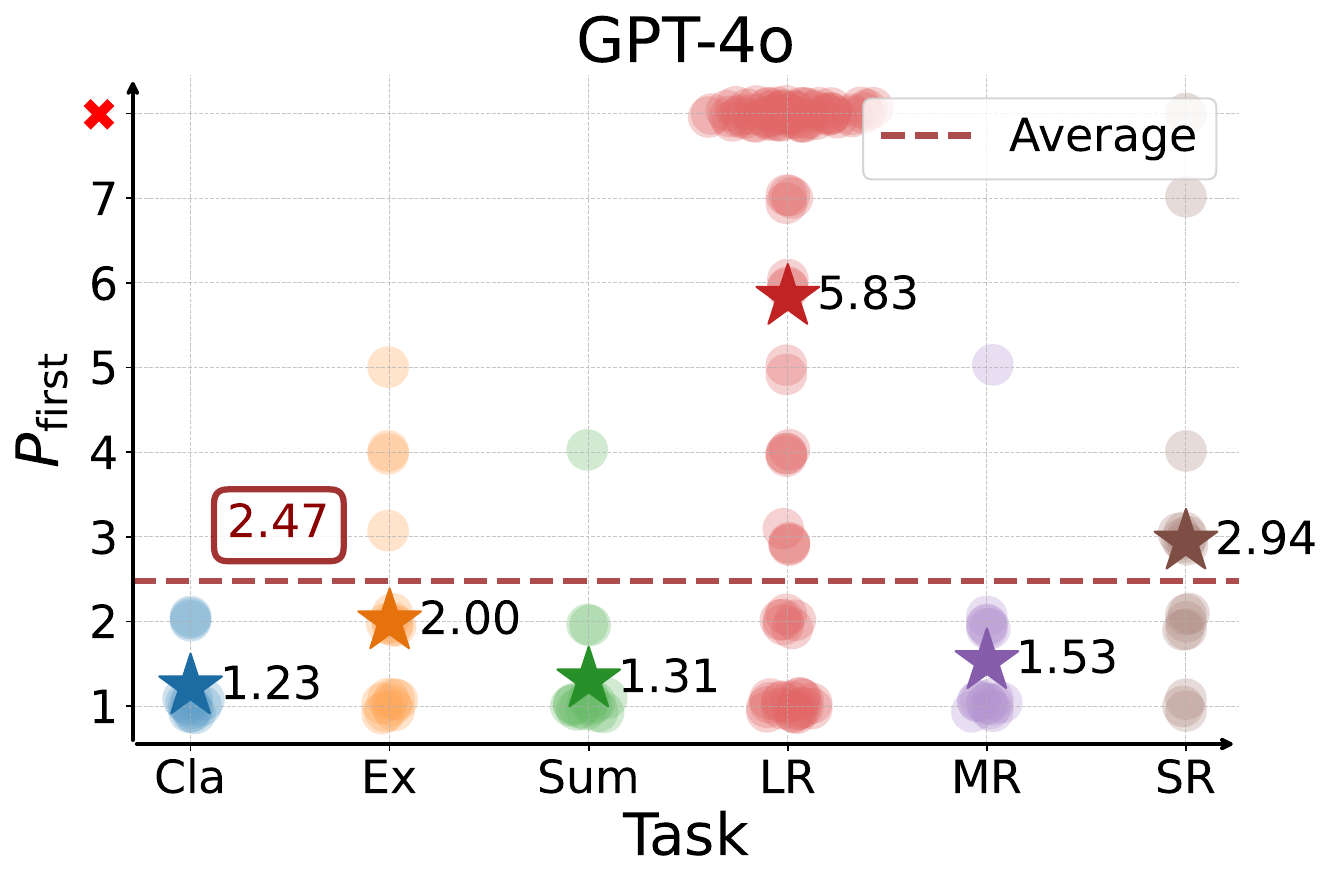}}
    \subfigure[]{\includegraphics[width=0.325\textwidth]{figures/RQ2_p_first/feedback/OpenAI-o3-mini_p_first.pdf}}
    \subfigure[]{\includegraphics[width=0.325\textwidth]{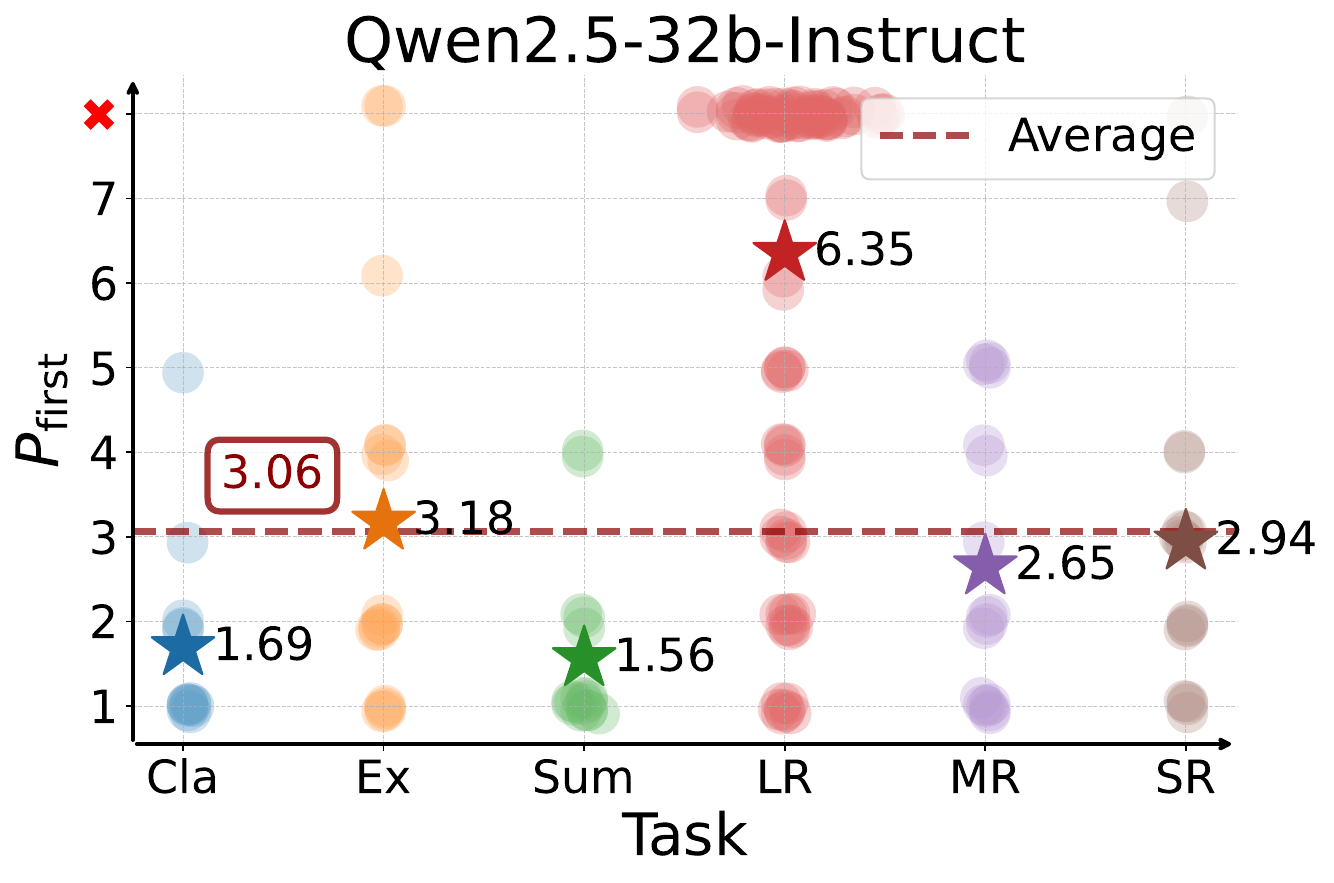}}

    \caption{Results of the average position of the first correct solution $P_{\mathrm{first}}$ for feedback learning.}
    \label{fig:Feedback_P_first}
\end{figure}

\begin{figure}[htbp]
    \centering
    \subfigure[]{\includegraphics[width=0.325\textwidth]{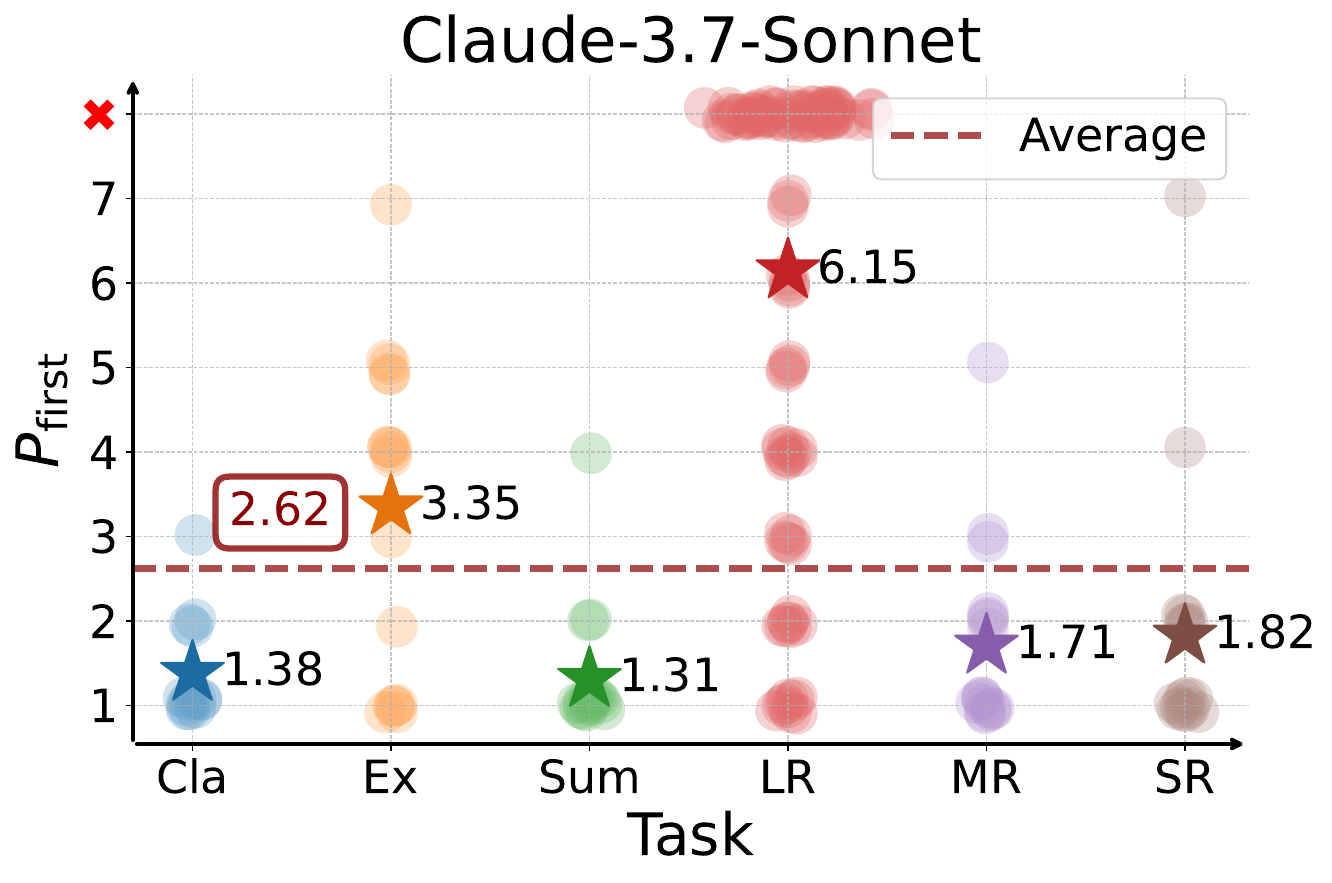}}
    \subfigure[]{\includegraphics[width=0.325\textwidth]{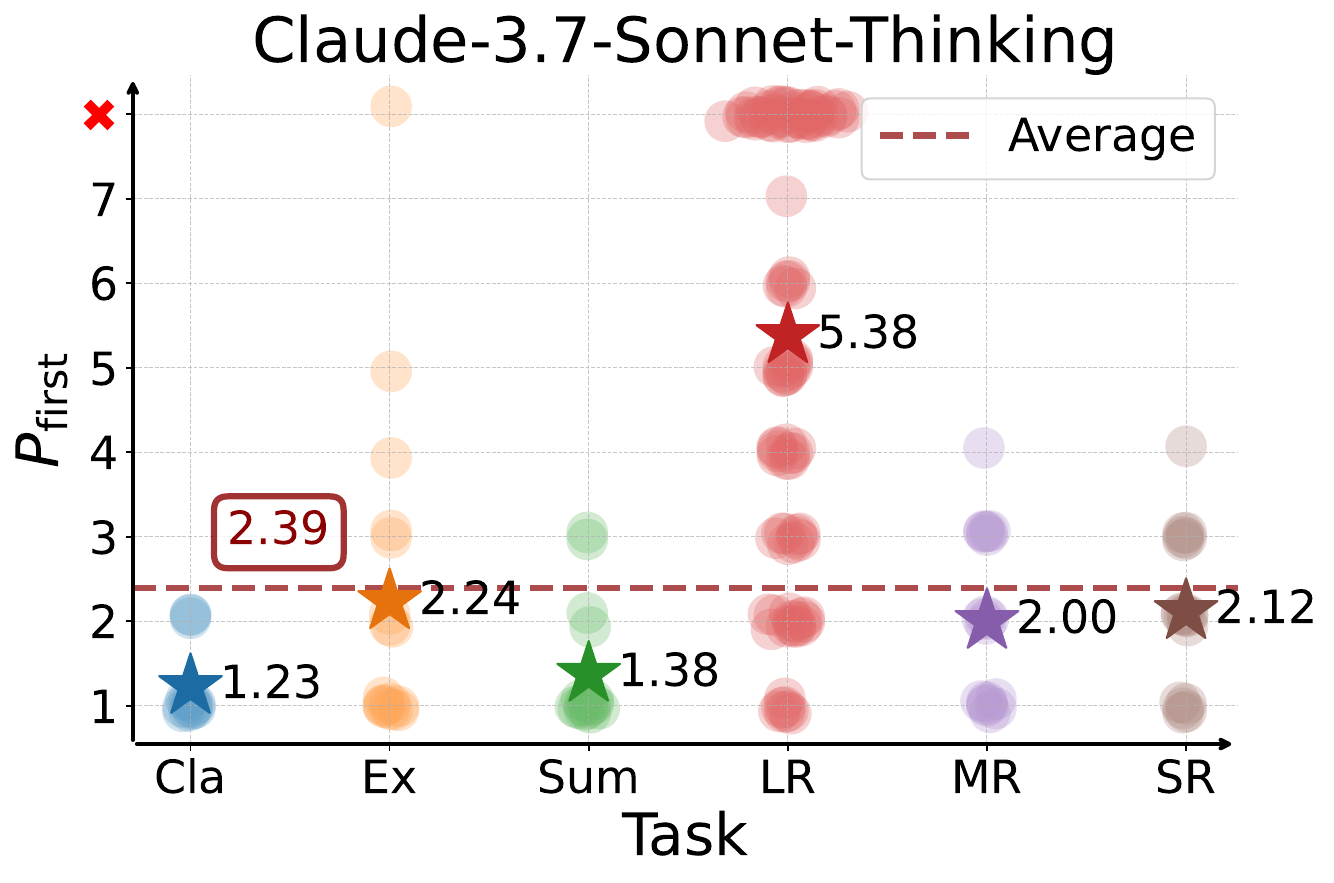}}
    \subfigure[]{\includegraphics[width=0.325\textwidth]{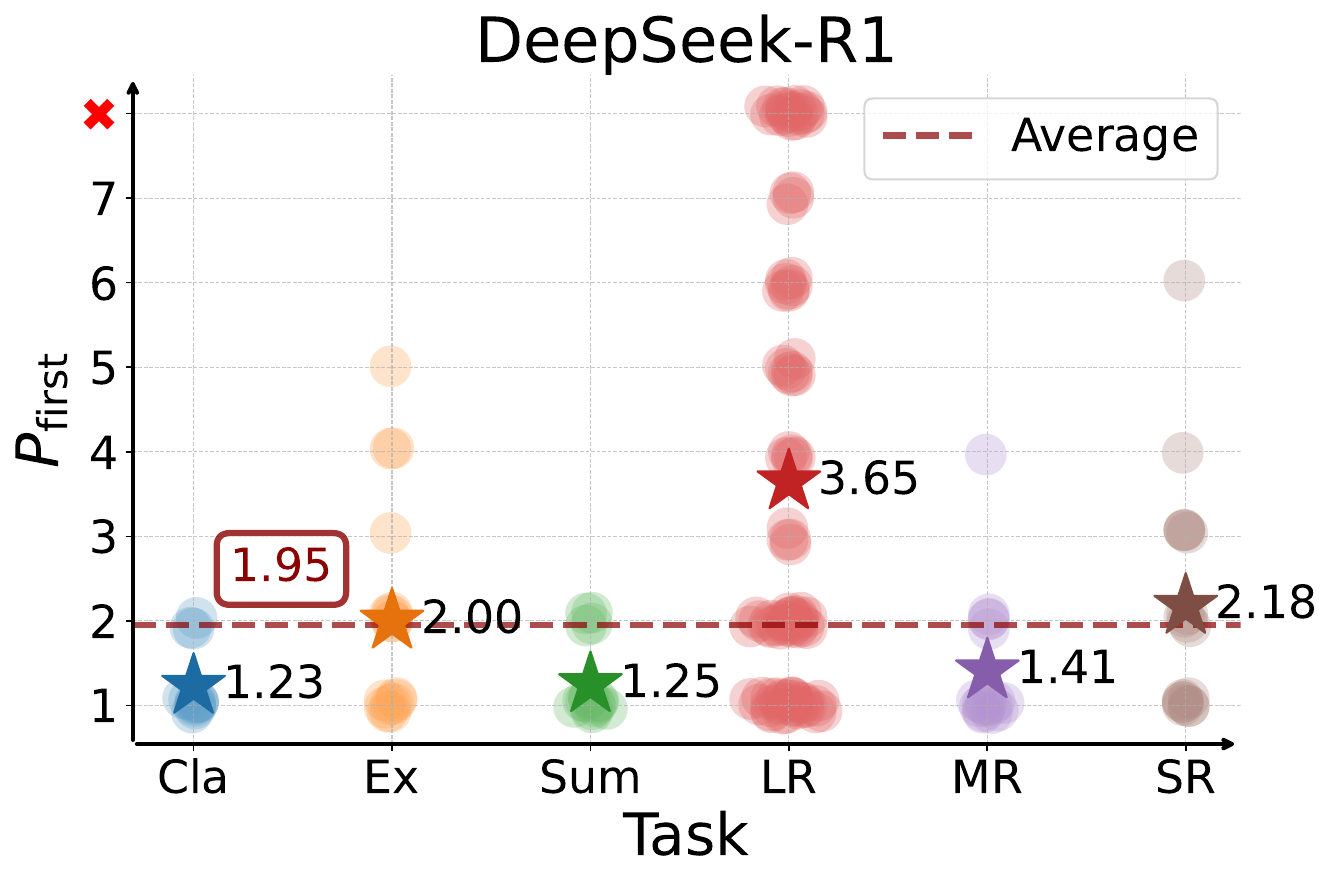}} 
    
    \subfigure[]{\includegraphics[width=0.325\textwidth]{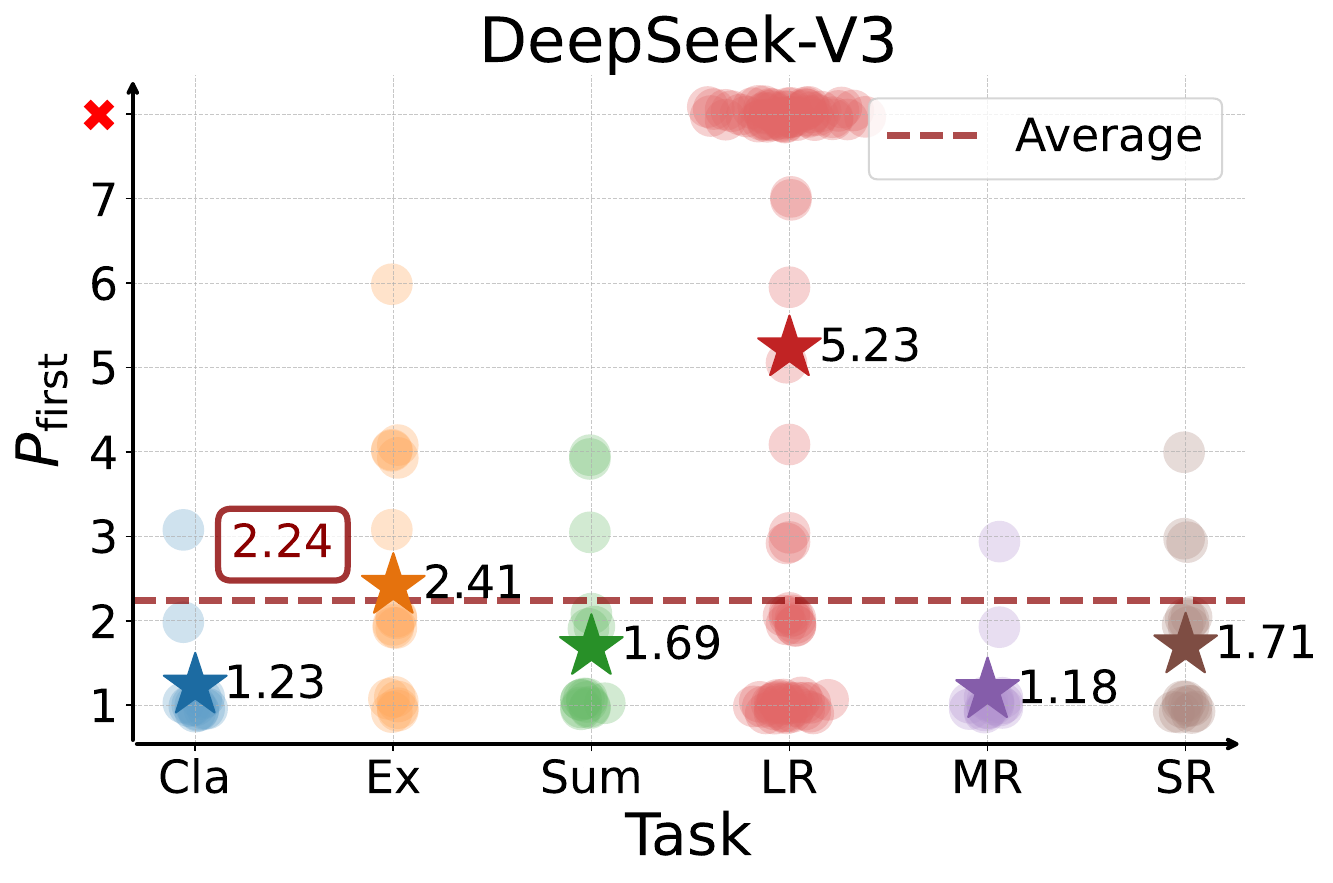}}
    \subfigure[]{\includegraphics[width=0.325\textwidth]{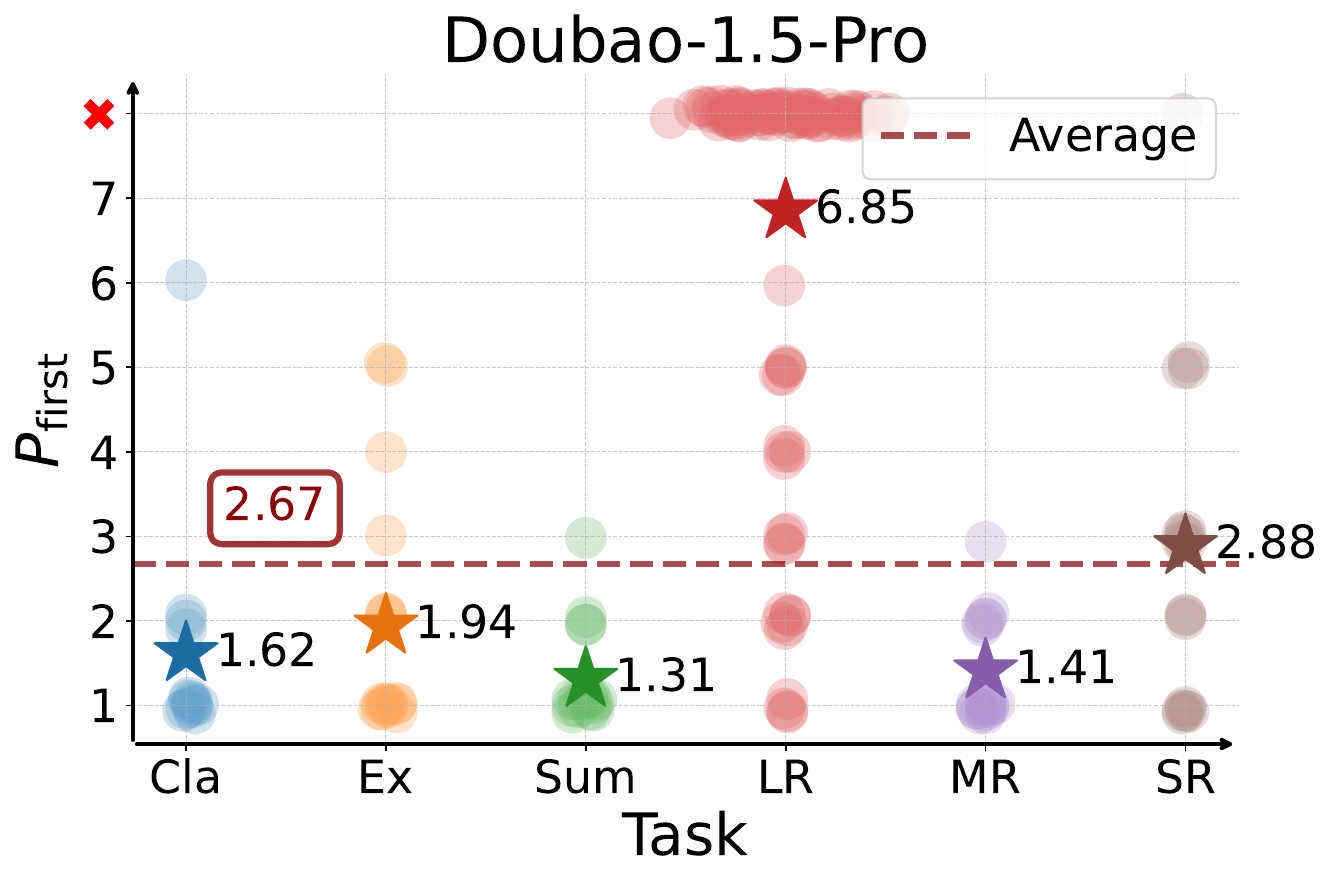}}
    \subfigure[]{\includegraphics[width=0.325\textwidth]{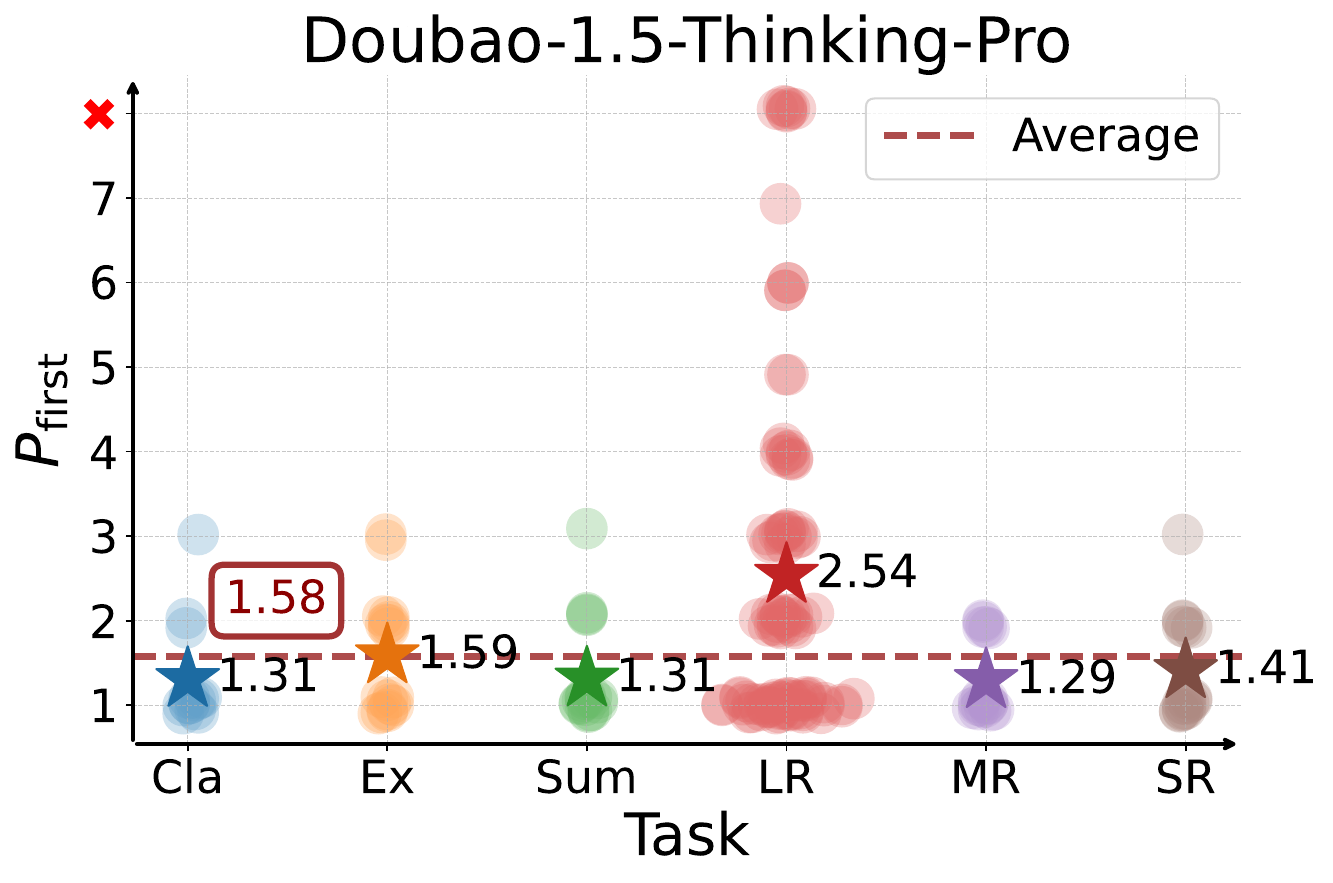}} 
    
    \subfigure[]{\includegraphics[width=0.325\textwidth]{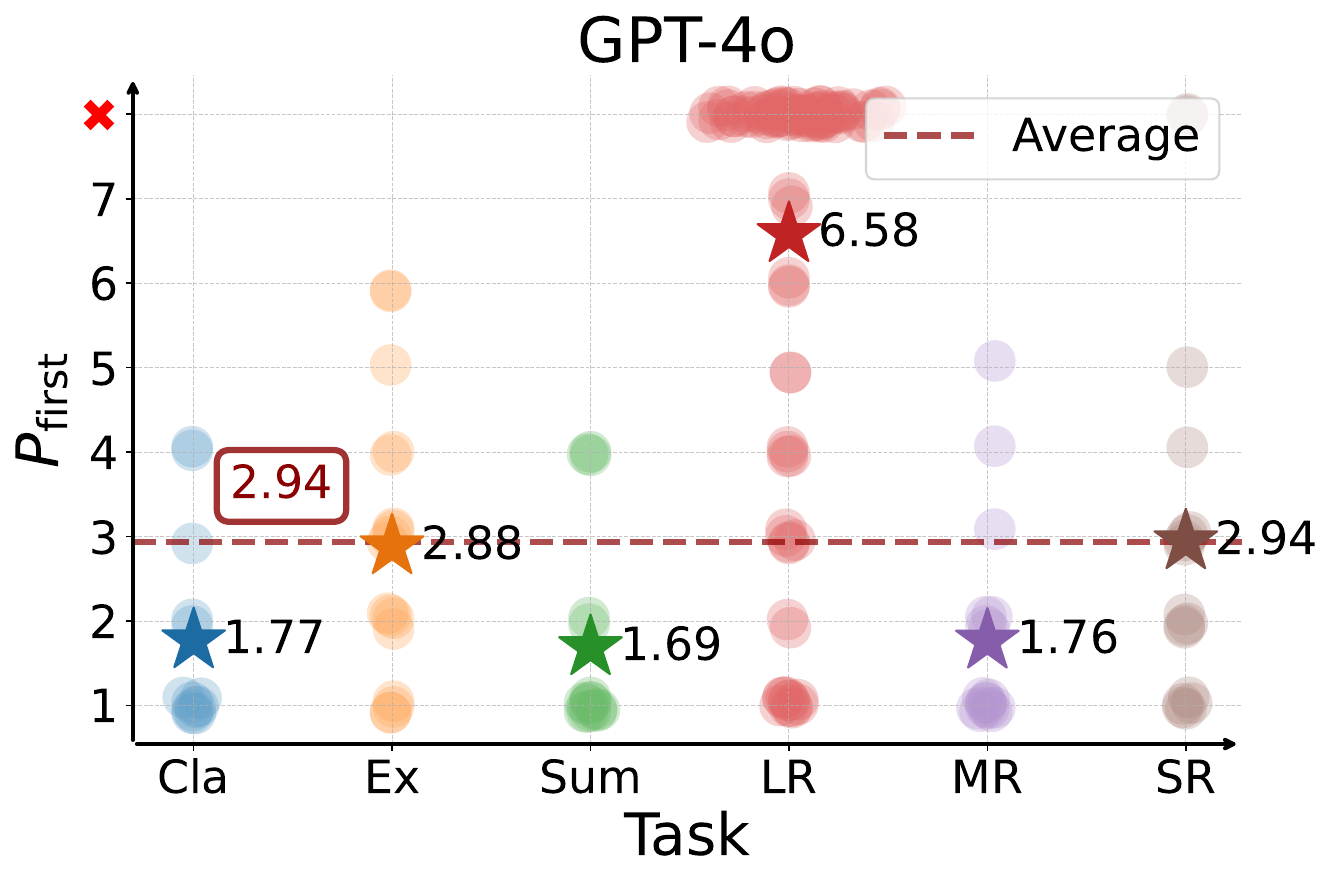}}
    \subfigure[]{\includegraphics[width=0.325\textwidth]{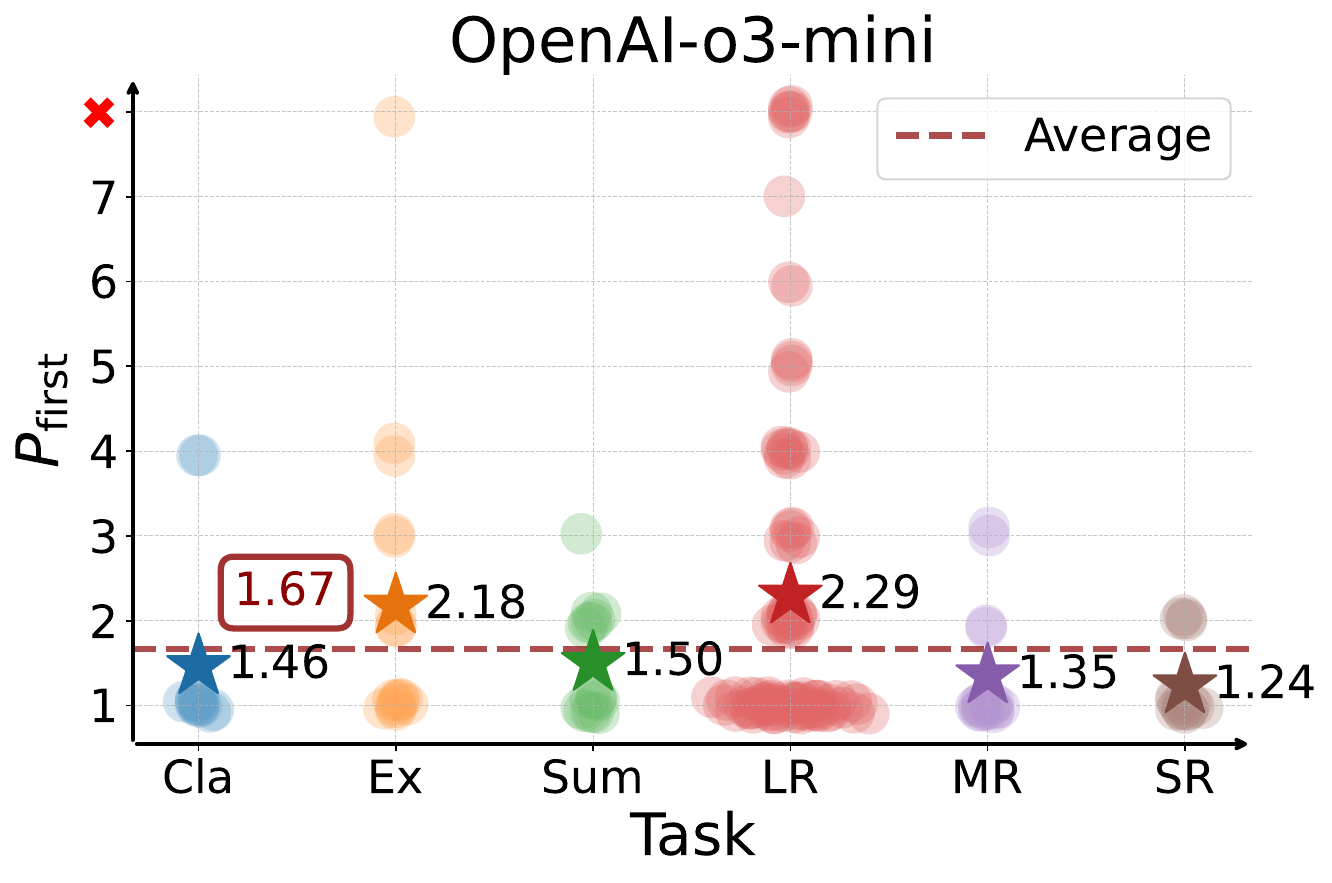}}
    \subfigure[]{\includegraphics[width=0.325\textwidth]{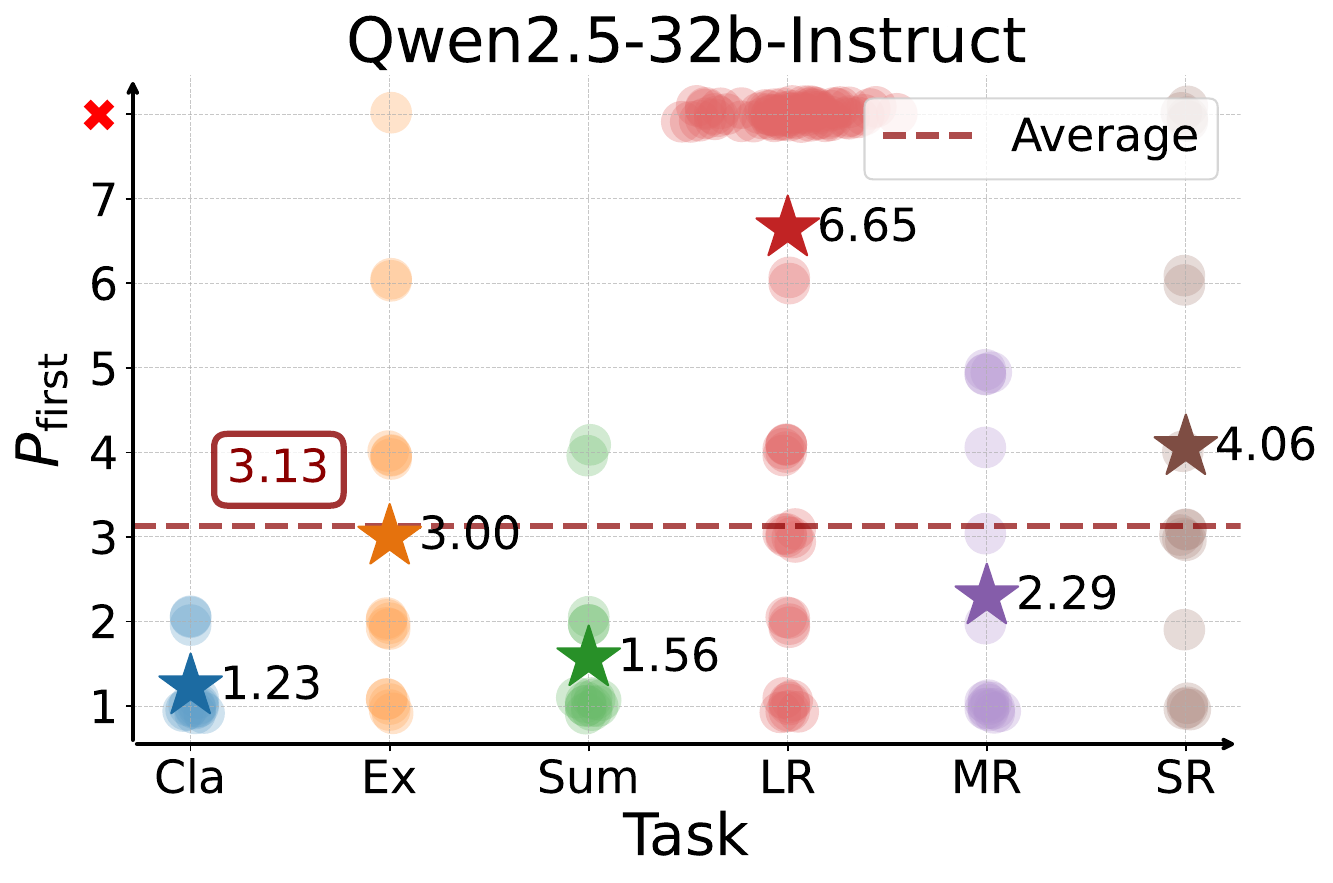}}
    \caption{Results of the average position of the first correct solution $P_{\mathrm{first}}$ for demonstration learning.}
    \label{fig:Demonstration P first}
\end{figure}

Figures \ref{fig:Feedback_P_first} and \ref{fig:Demonstration P first} show the results of the average position of the first correct solution ($P_{\mathrm{first}}$) for feedback learning and demonstration learning, respectively. 
This metric measures how quickly, on average, the model achieves its first success in a sequence, thus indicating its initial learning speed.

\clearpage

\subsection{Results of Average Number of Consecutive Correct Solutions (\texorpdfstring{$N_{\mathrm{consec}}$}{N\_consec})}
\label{app:metric-4-consec}

\begin{figure}[htbp]
    \centering
    \subfigure[]{\includegraphics[width=0.325\textwidth]{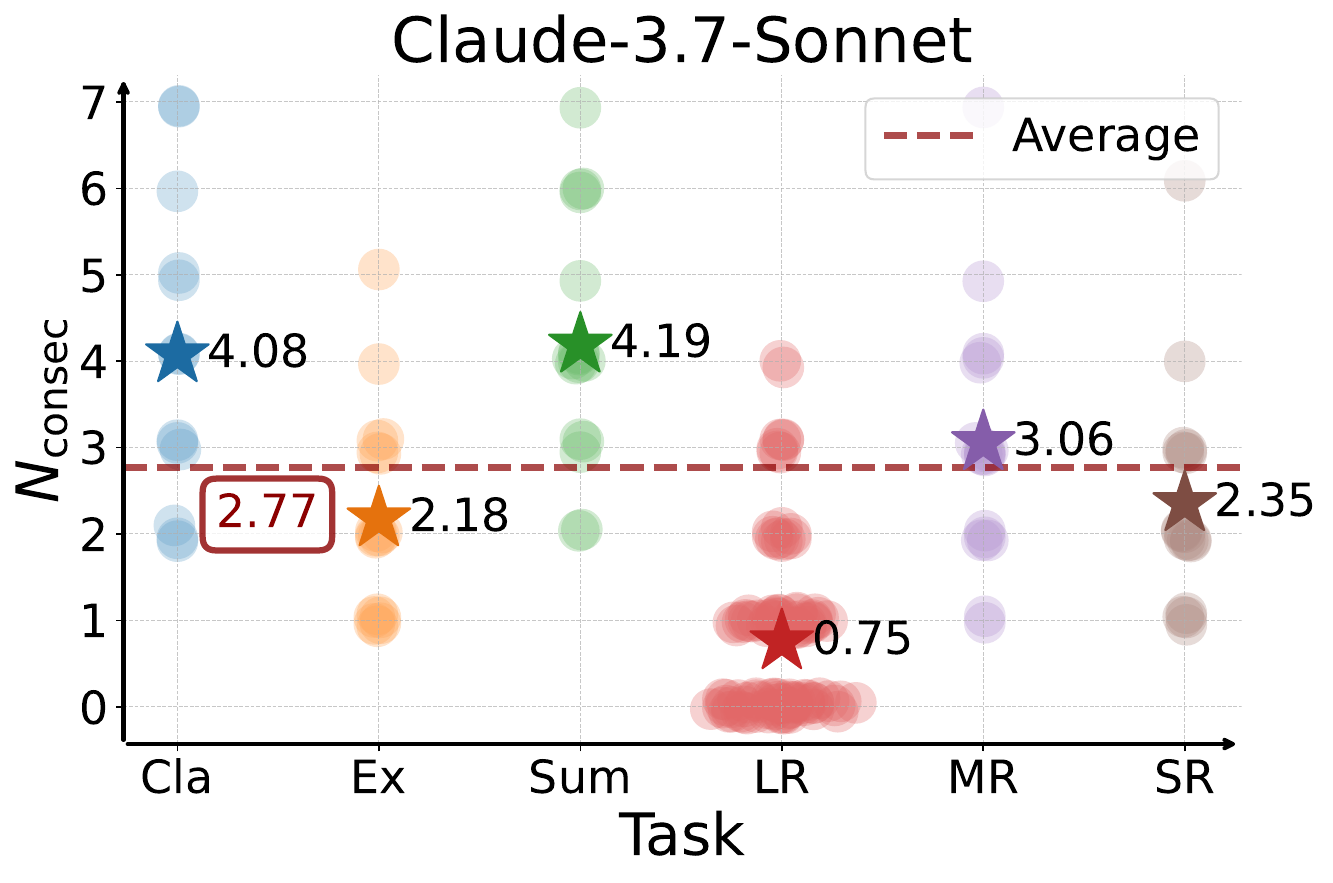}}
    \subfigure[]{\includegraphics[width=0.325\textwidth]{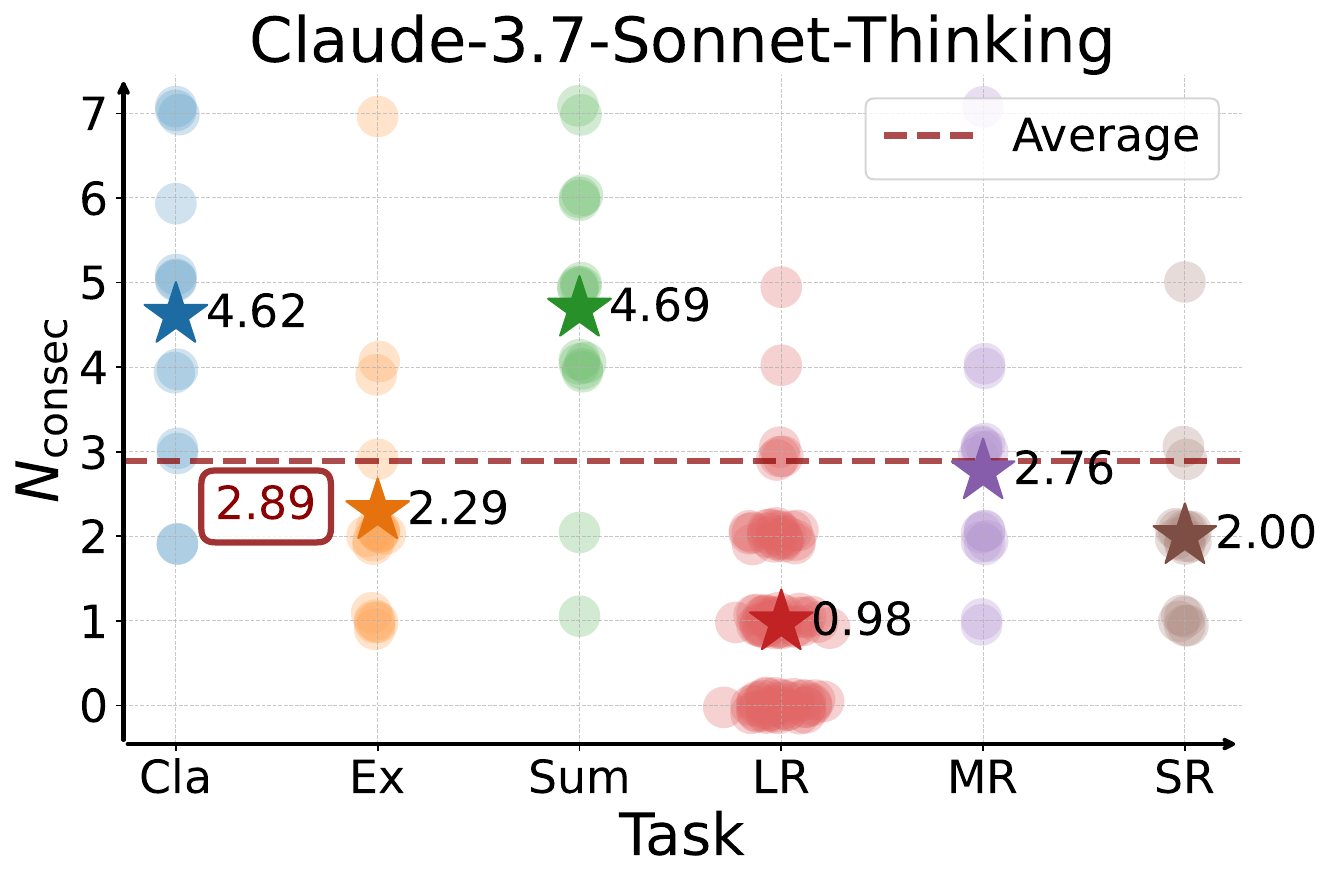}}
    \subfigure[]{\includegraphics[width=0.325\textwidth]{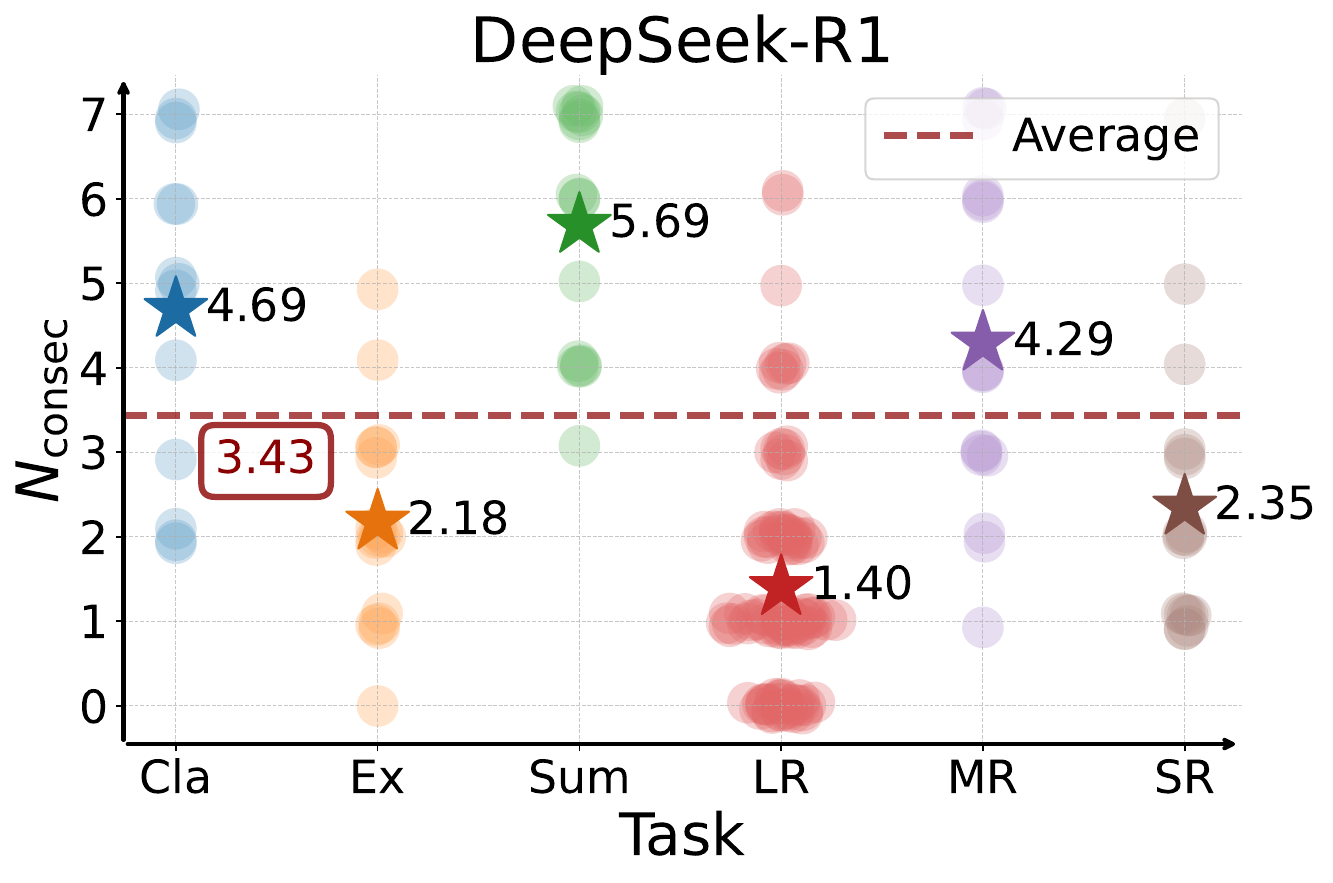}}\\    
    \subfigure[]{\includegraphics[width=0.325\textwidth]{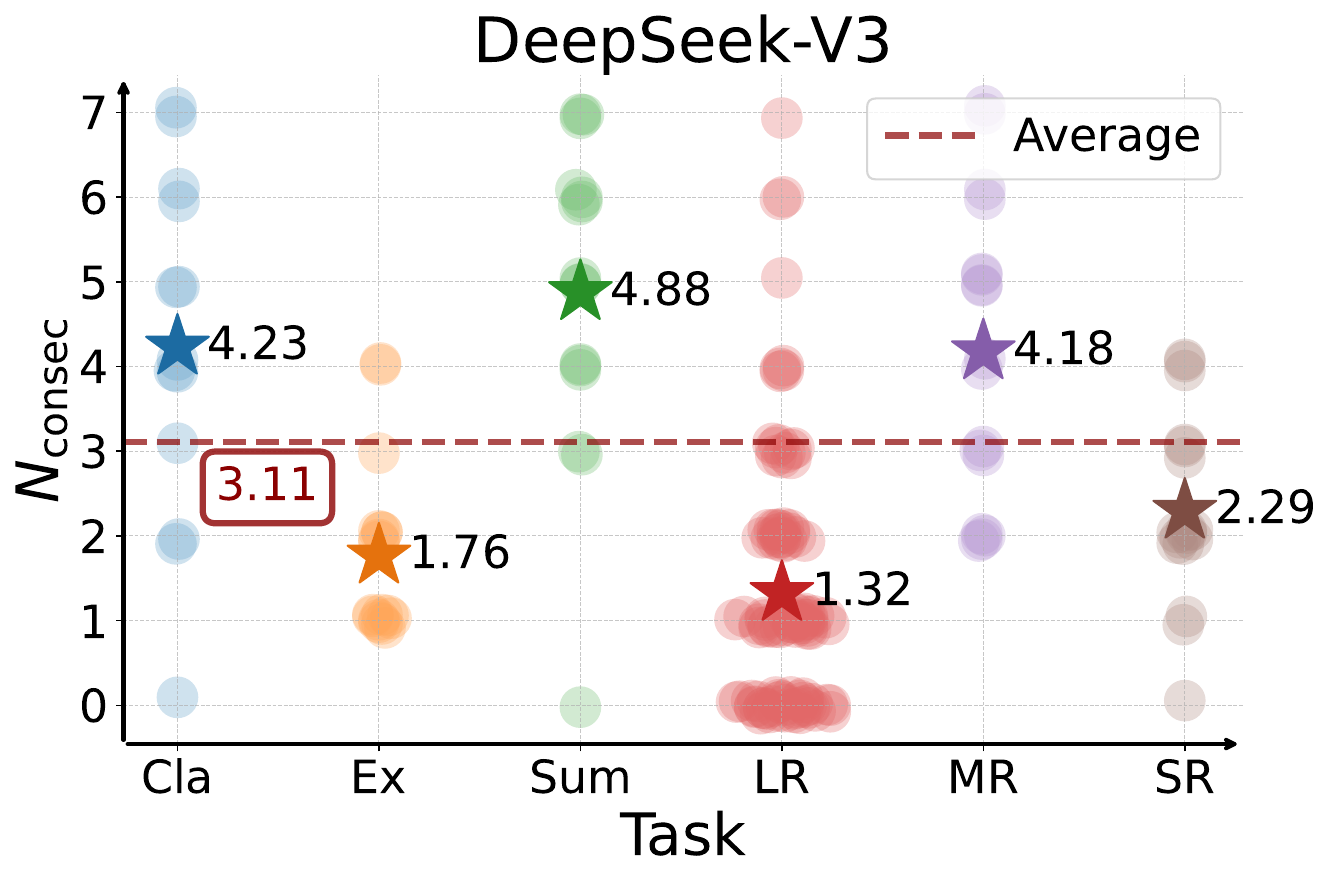}}
    \subfigure[]{\includegraphics[width=0.325\textwidth]{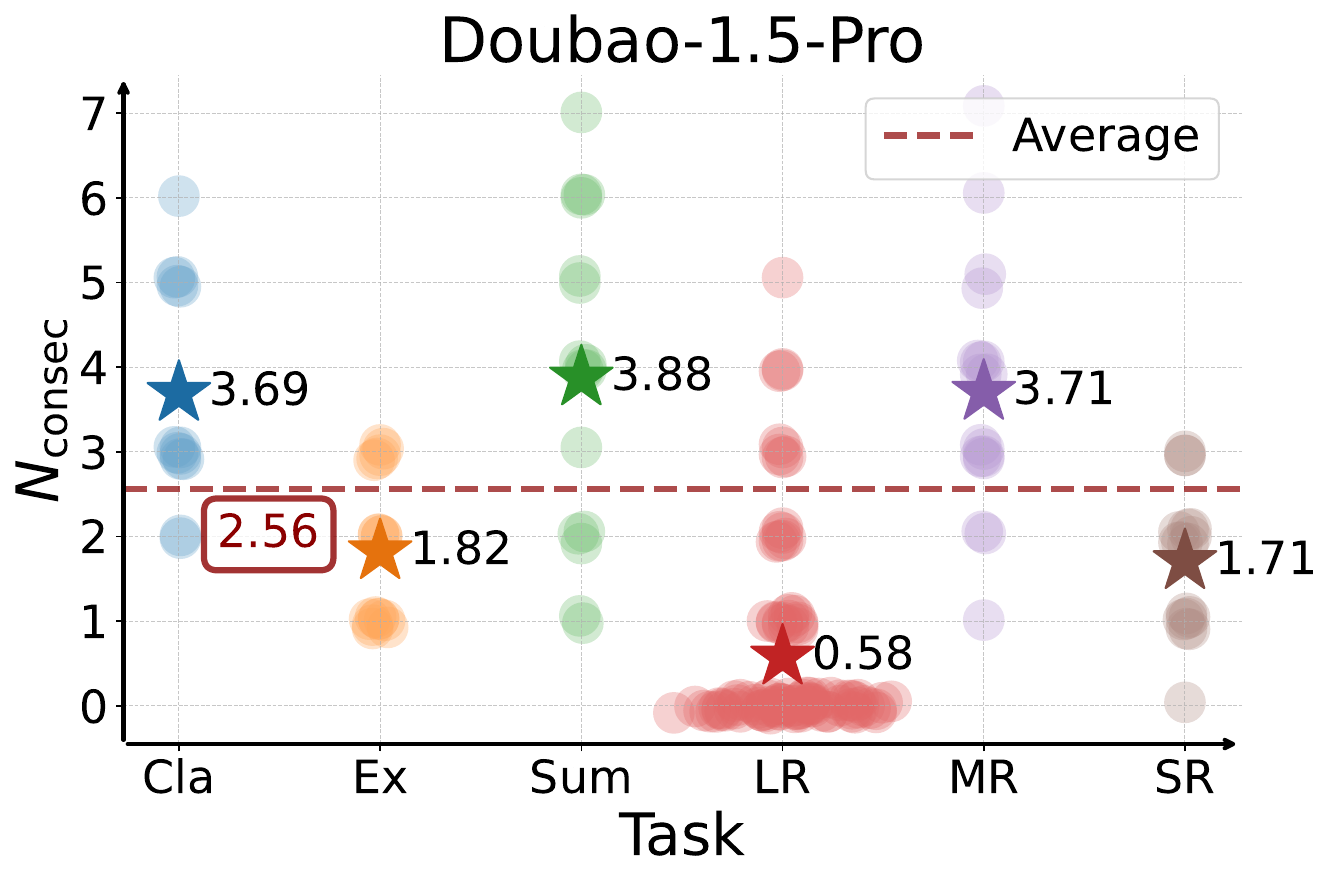}}
    \subfigure[]{\includegraphics[width=0.325\textwidth]{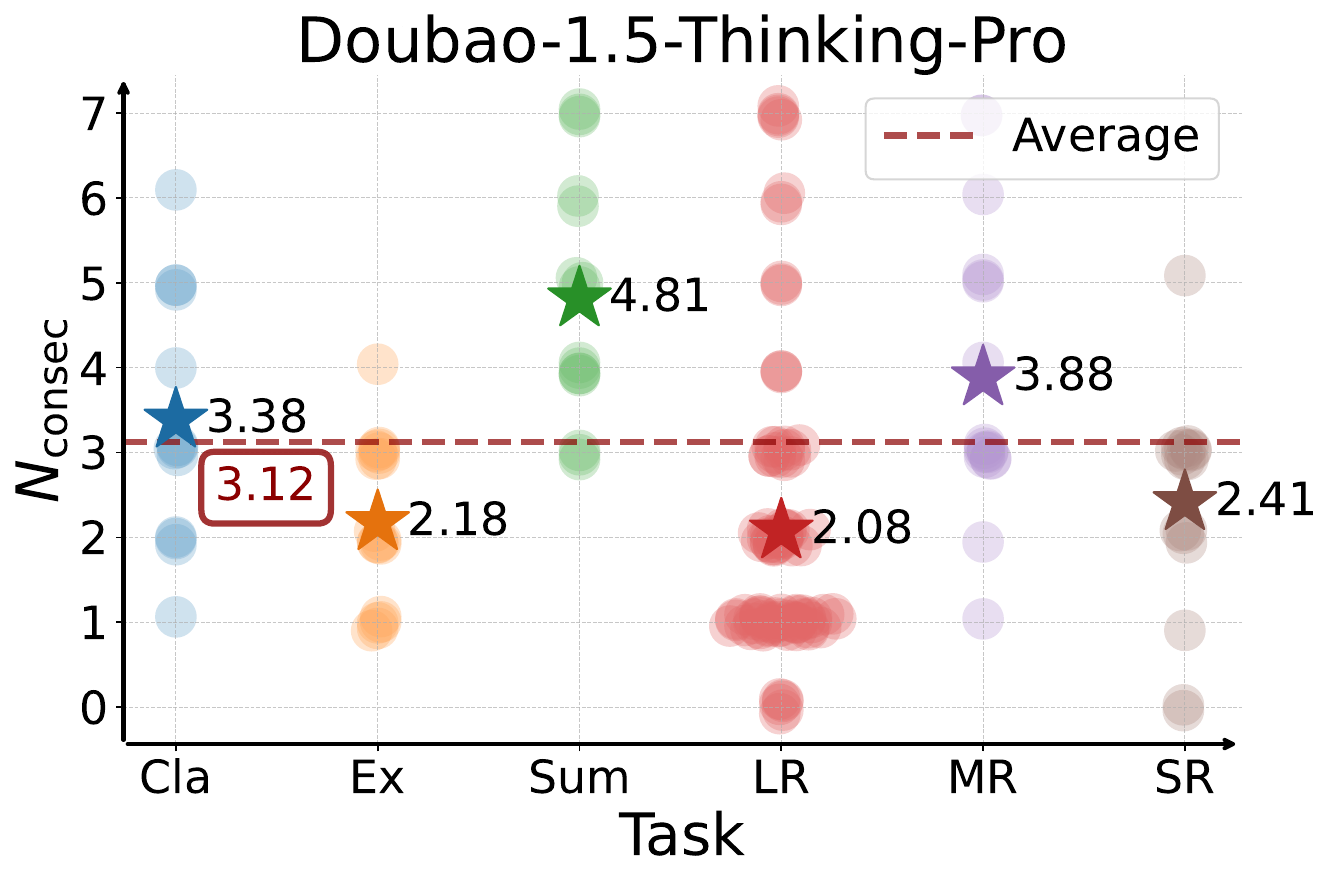}} \\
    \subfigure[]{\includegraphics[width=0.325\textwidth]{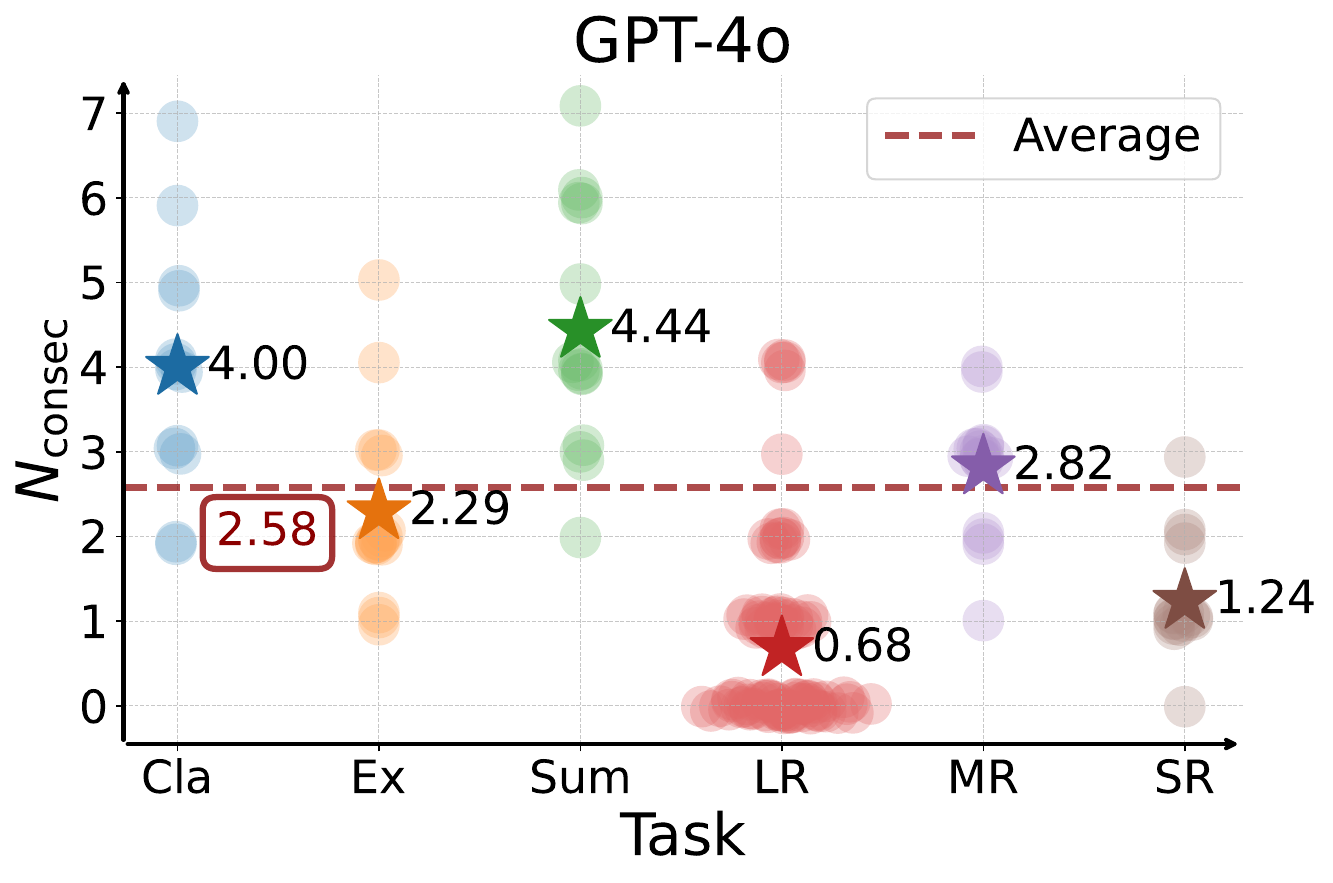}}
    \subfigure[]{\includegraphics[width=0.325\textwidth]{figures/RQ1_n_consec/feedback/OpenAI-o3-mini_n_consec.pdf}}
    \subfigure[]{\includegraphics[width=0.325\textwidth]{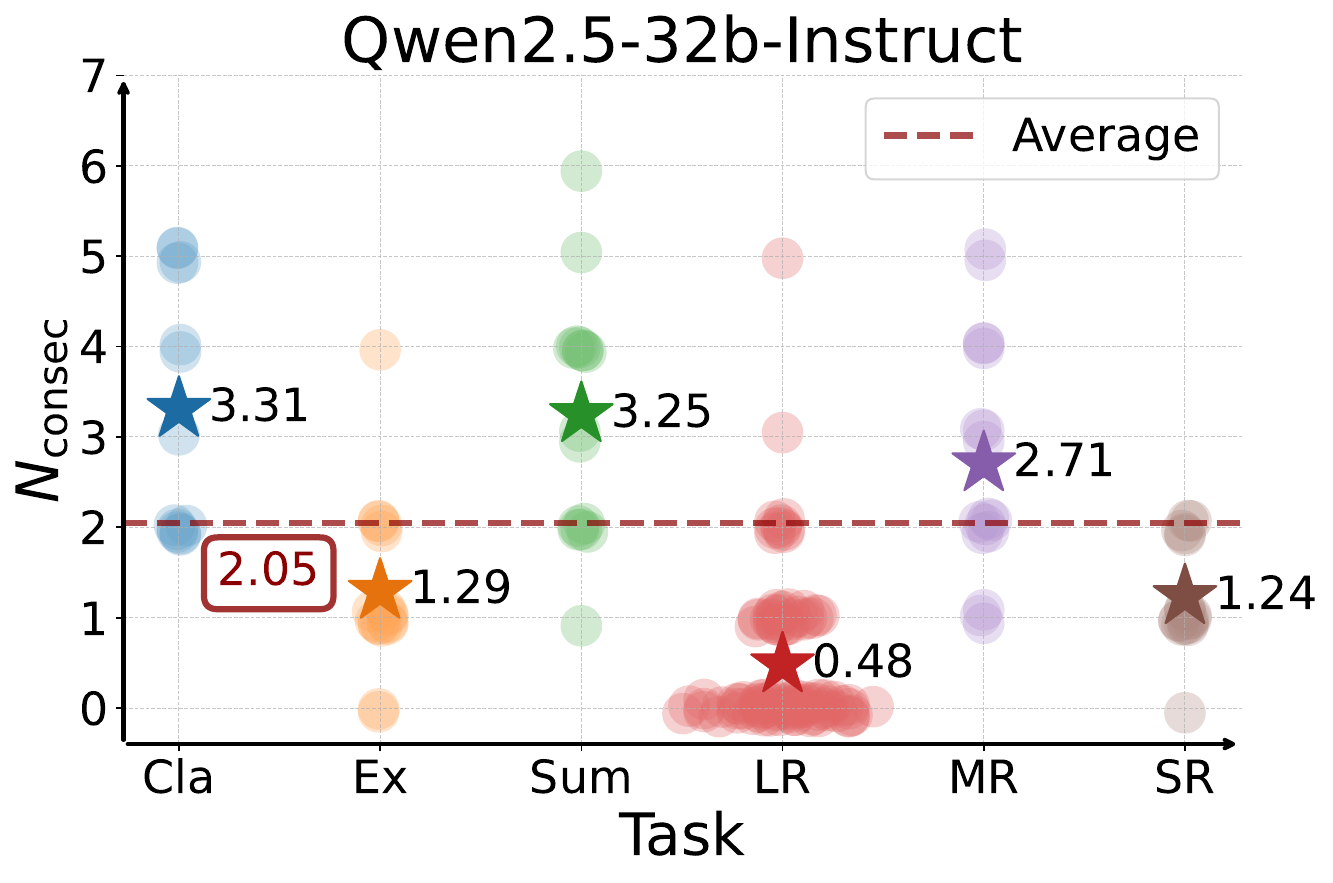}}
    \caption{Results of the average number of consecutive correct solutions ($N_{\mathrm{consec}}$) for feedback learning.}
    \label{fig:Feedback N consec}
\end{figure}

\begin{figure}[htbp]
    \centering
    \subfigure[]{\includegraphics[width=0.325\textwidth]{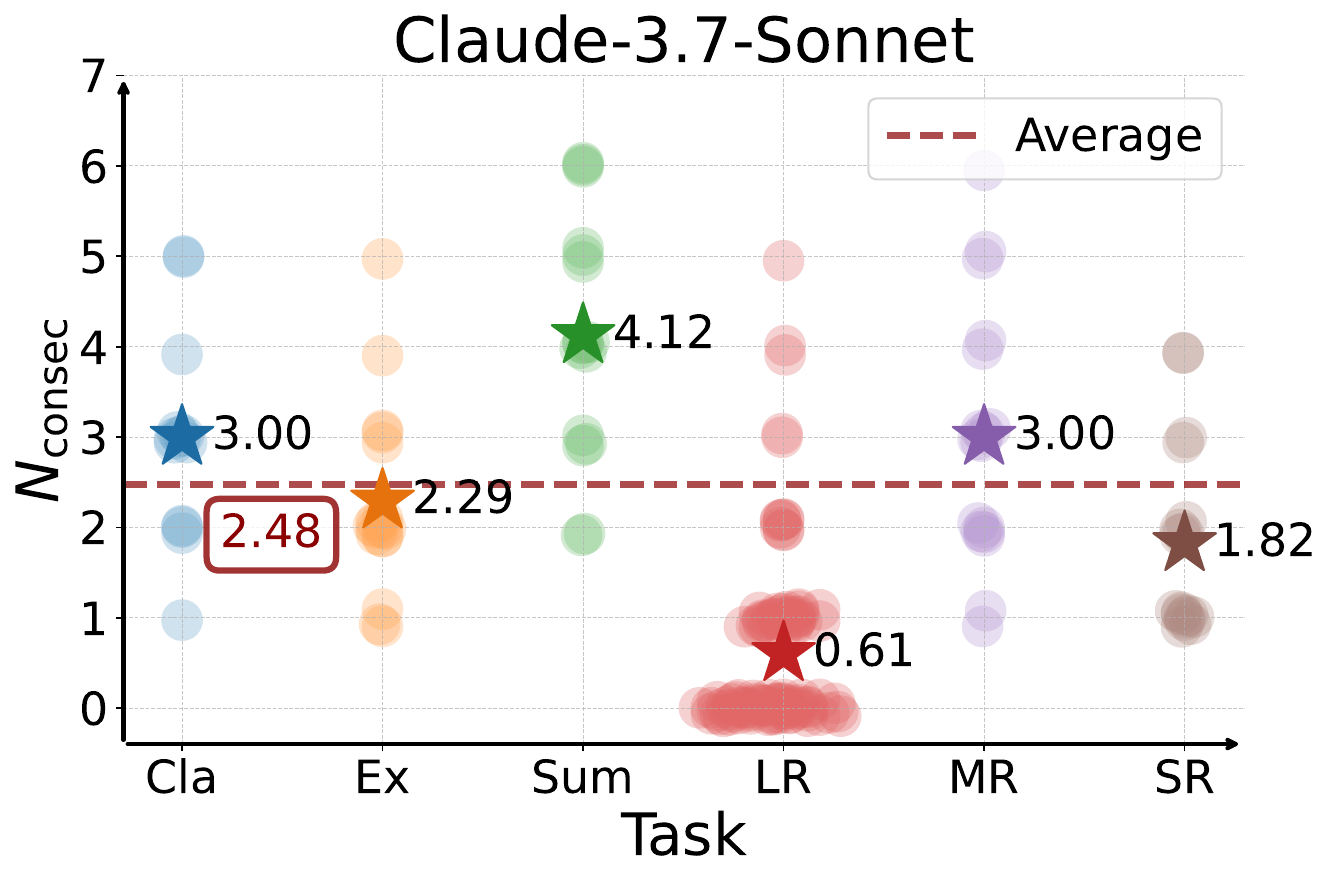}}
    \subfigure[]{\includegraphics[width=0.325\textwidth]{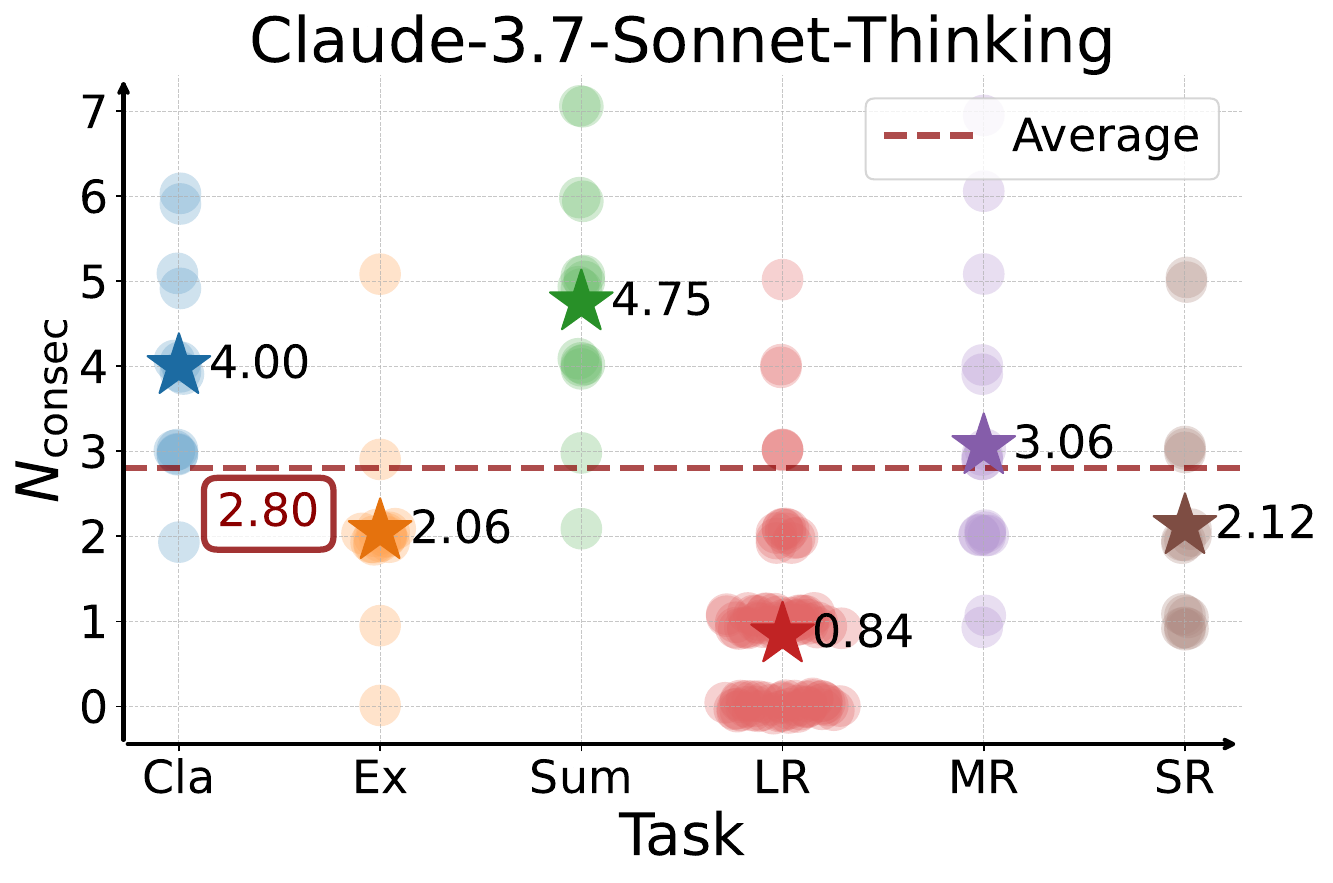}}
    \subfigure[]{\includegraphics[width=0.325\textwidth]{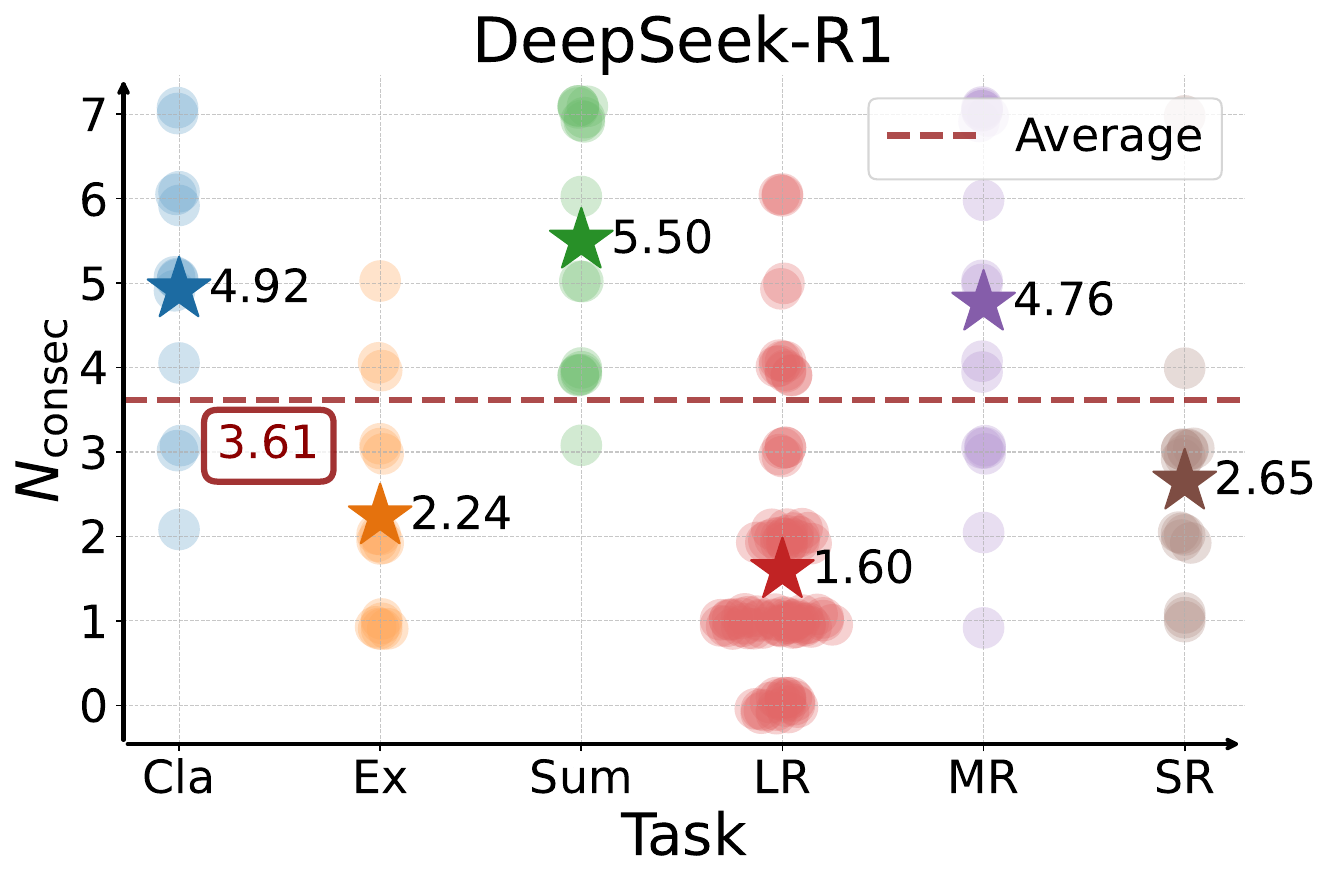}} 
    
    \subfigure[]{\includegraphics[width=0.325\textwidth]{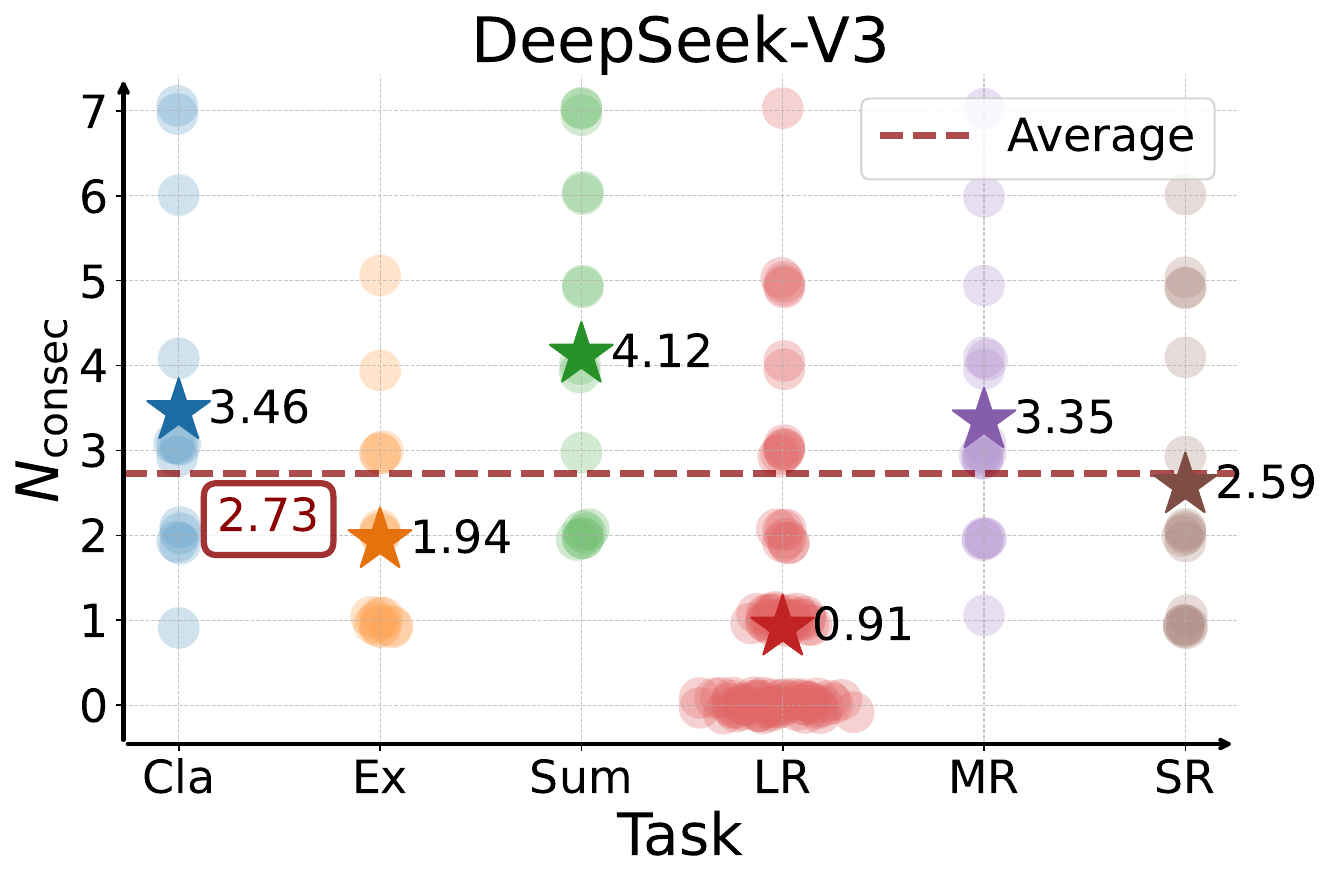}}
    \subfigure[]{\includegraphics[width=0.325\textwidth]{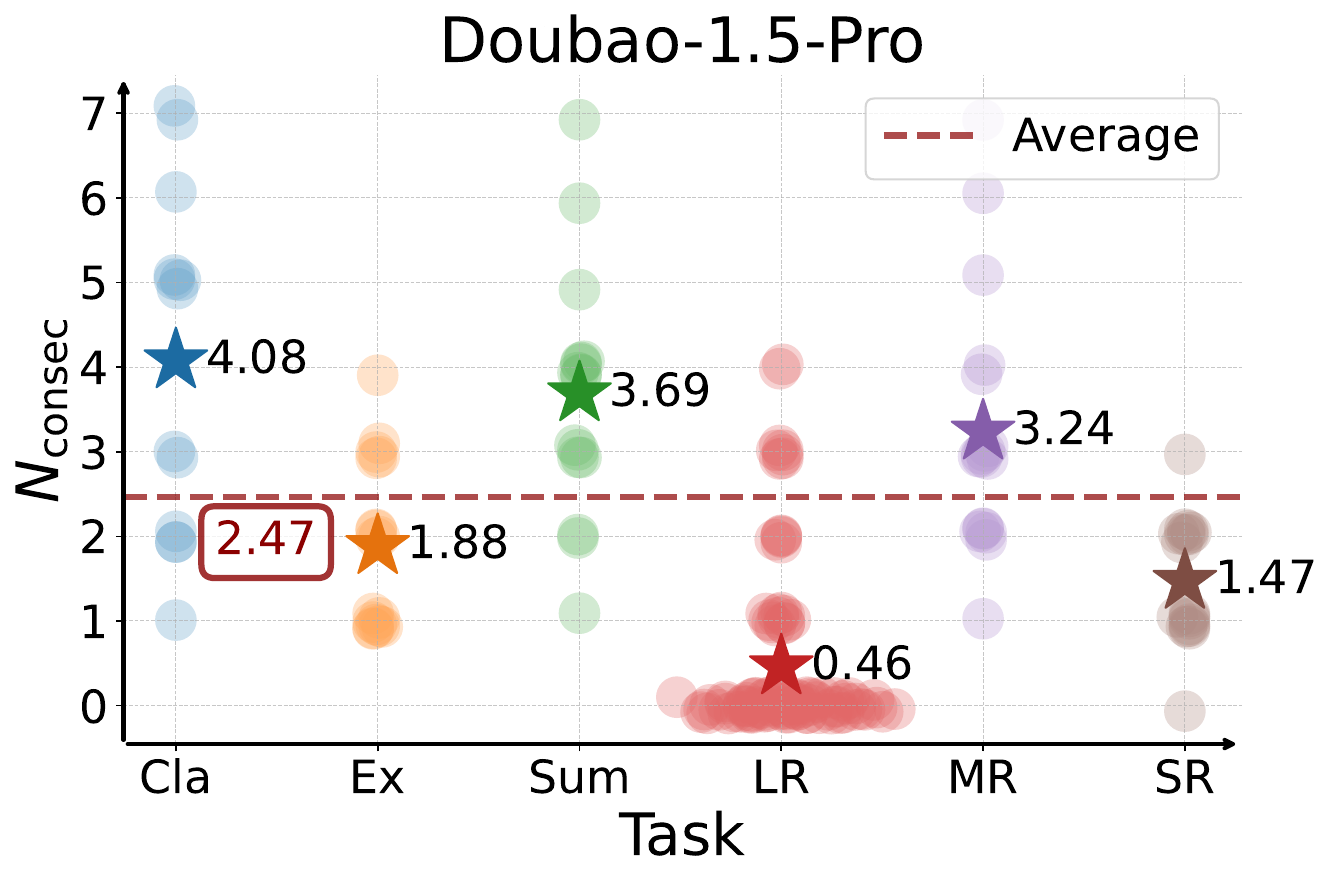}}
    \subfigure[]{\includegraphics[width=0.325\textwidth]{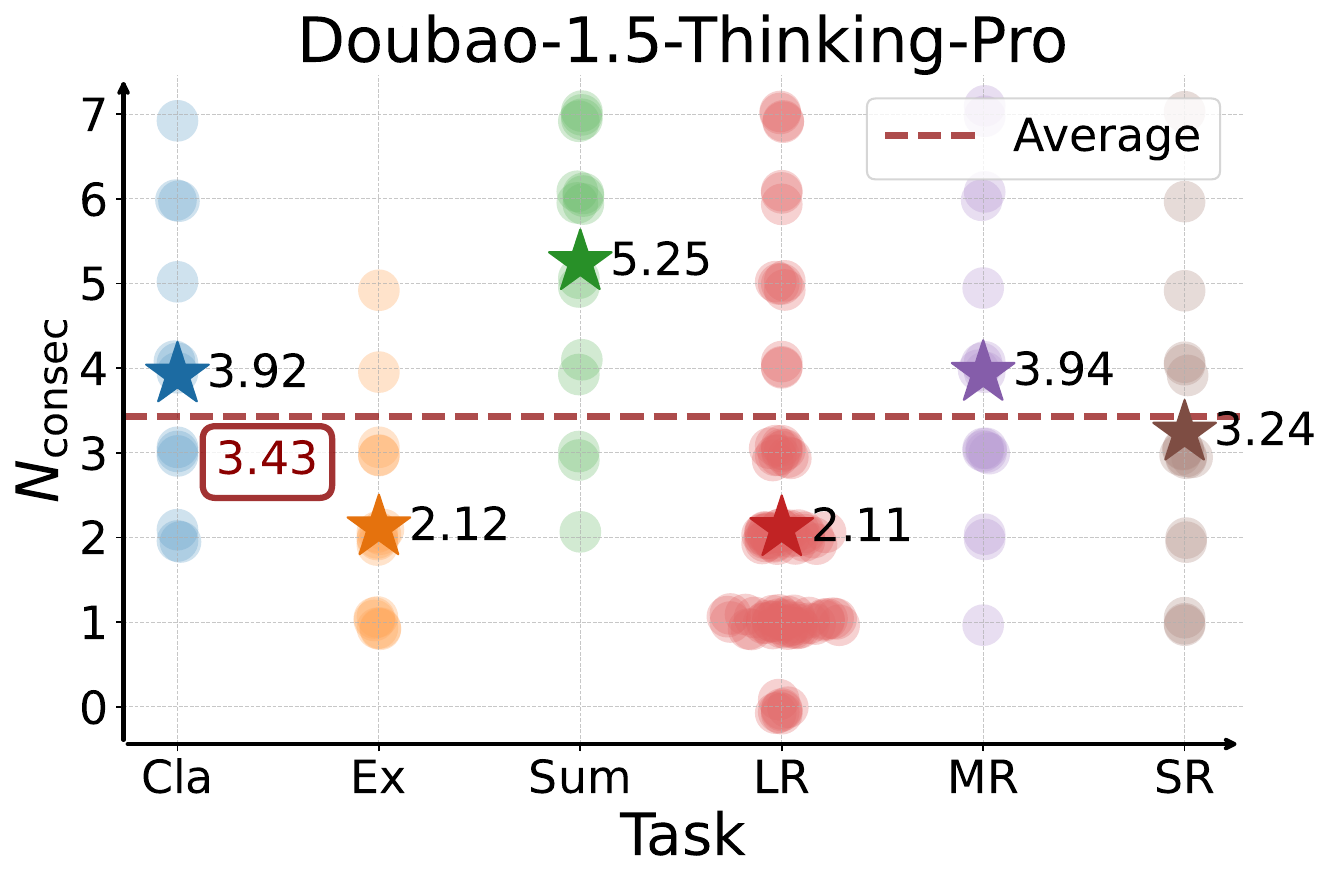}} 
    
    \subfigure[]{\includegraphics[width=0.325\textwidth]{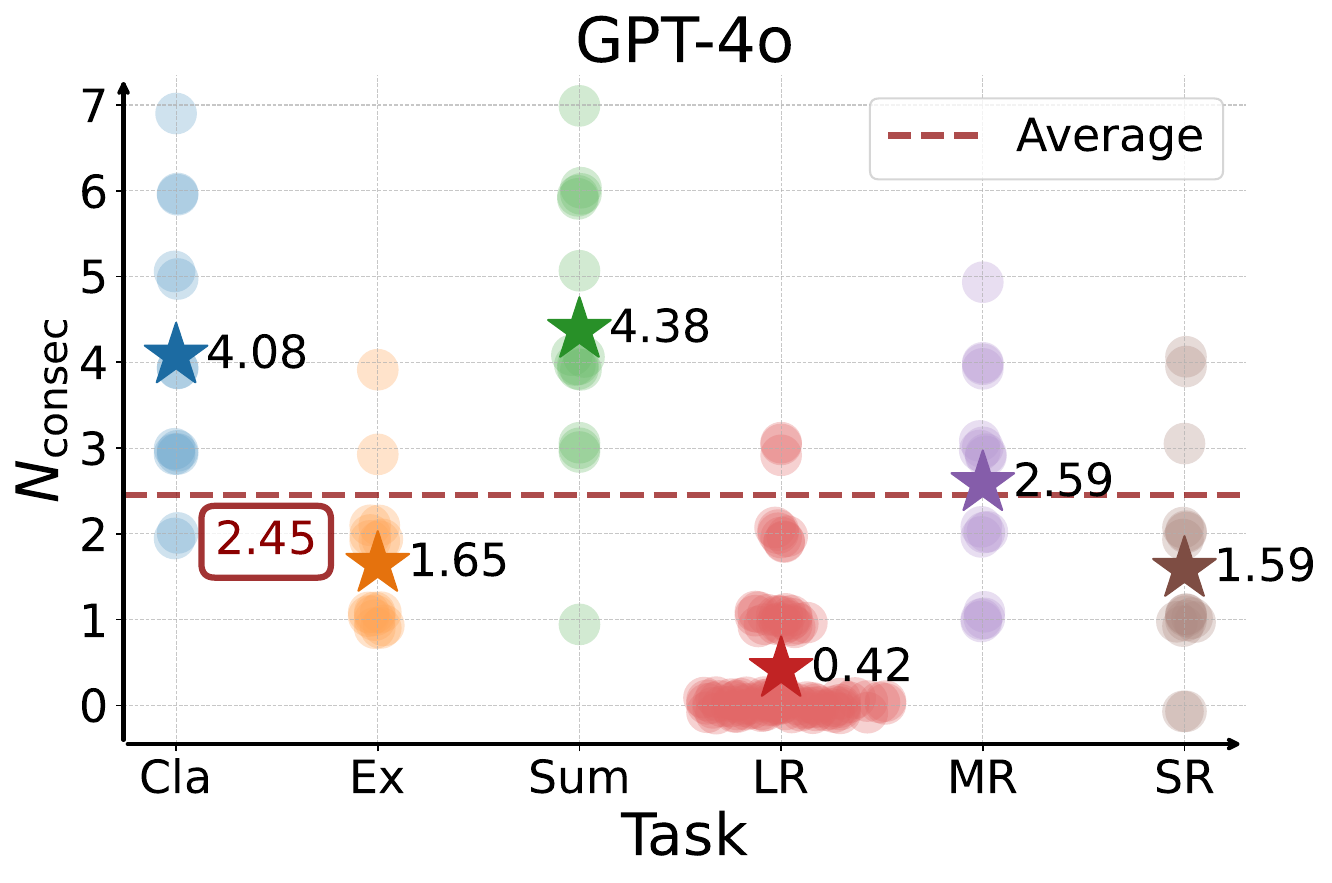}}
    \subfigure[]{\includegraphics[width=0.325\textwidth]{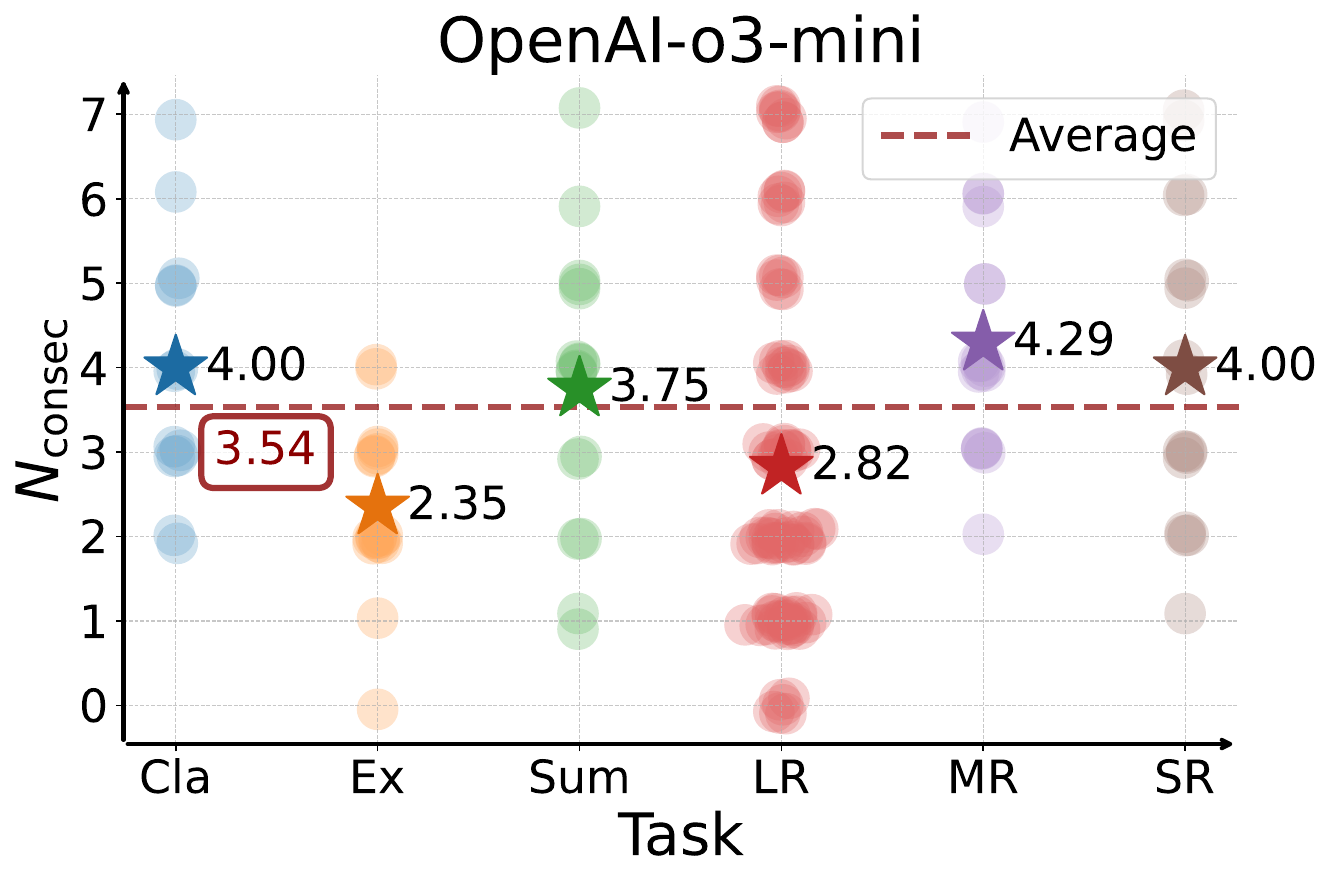}}
    \subfigure[]{\includegraphics[width=0.325\textwidth]{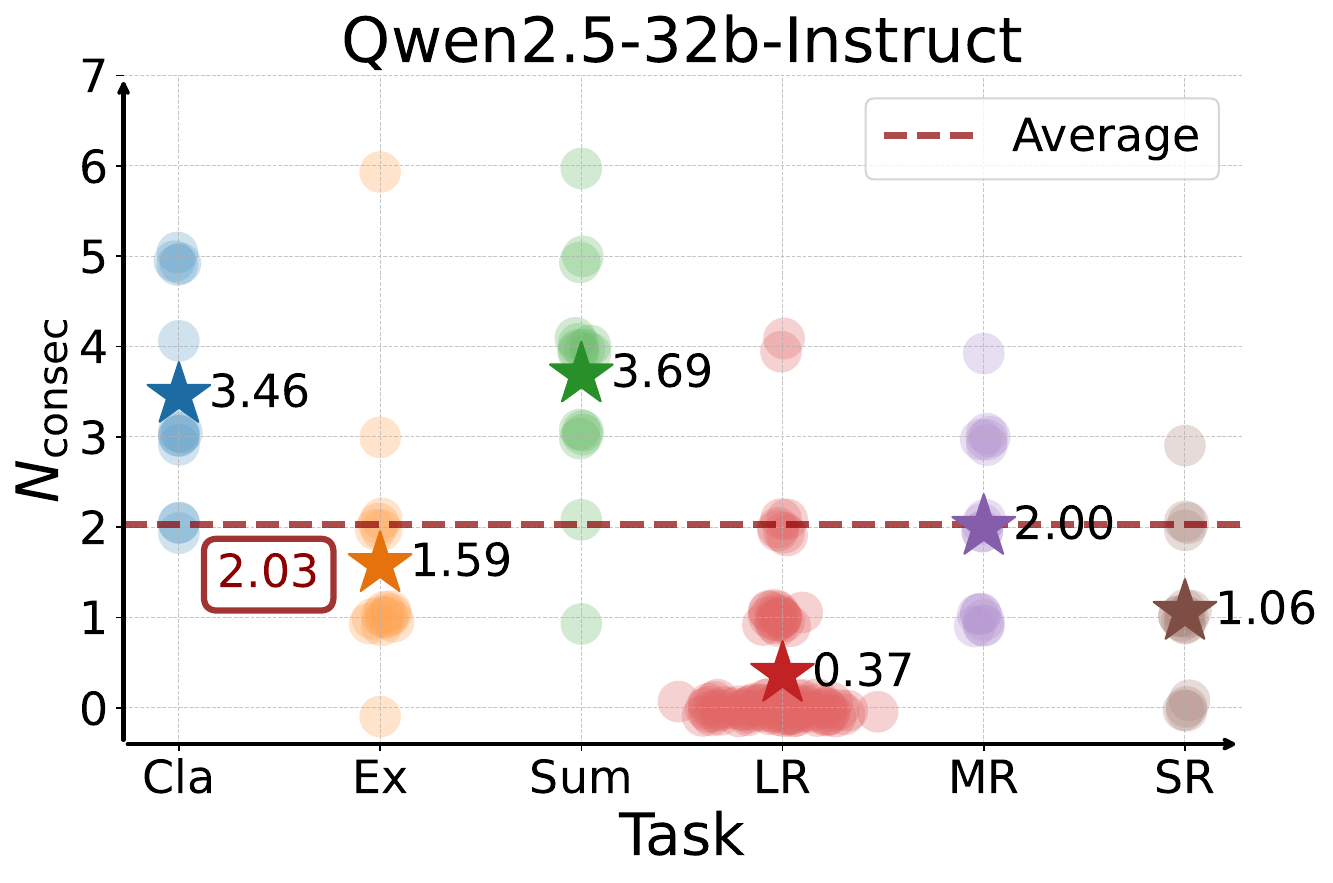}}
    \caption{Results of the average number of consecutive correct solutions ($N_{\mathrm{consec}}$) for demonstration learning.}
    \label{fig:Demonstration N consec}
\end{figure}

Figures \ref{fig:Feedback N consec} and \ref{fig:Demonstration N consec} report the results of the average number of consecutive correct solutions ($N_{\mathrm{consec}}$) for feedback learning and demonstration learning, respectively. 
This metric reflects the model's ability to leverage experience to consistently solve problems within a sequence, resulting in fewer errors.

\clearpage

\subsection{Results of Post-Warmup Accuracy (\texorpdfstring{$\mathrm{Acc}_{\mathrm{pw}-K}$}{Acc\_pw-K})}
\label{app:metric-5-pwk}

\begin{figure}[htbp]
    \centering
    \includegraphics[width=1\linewidth]{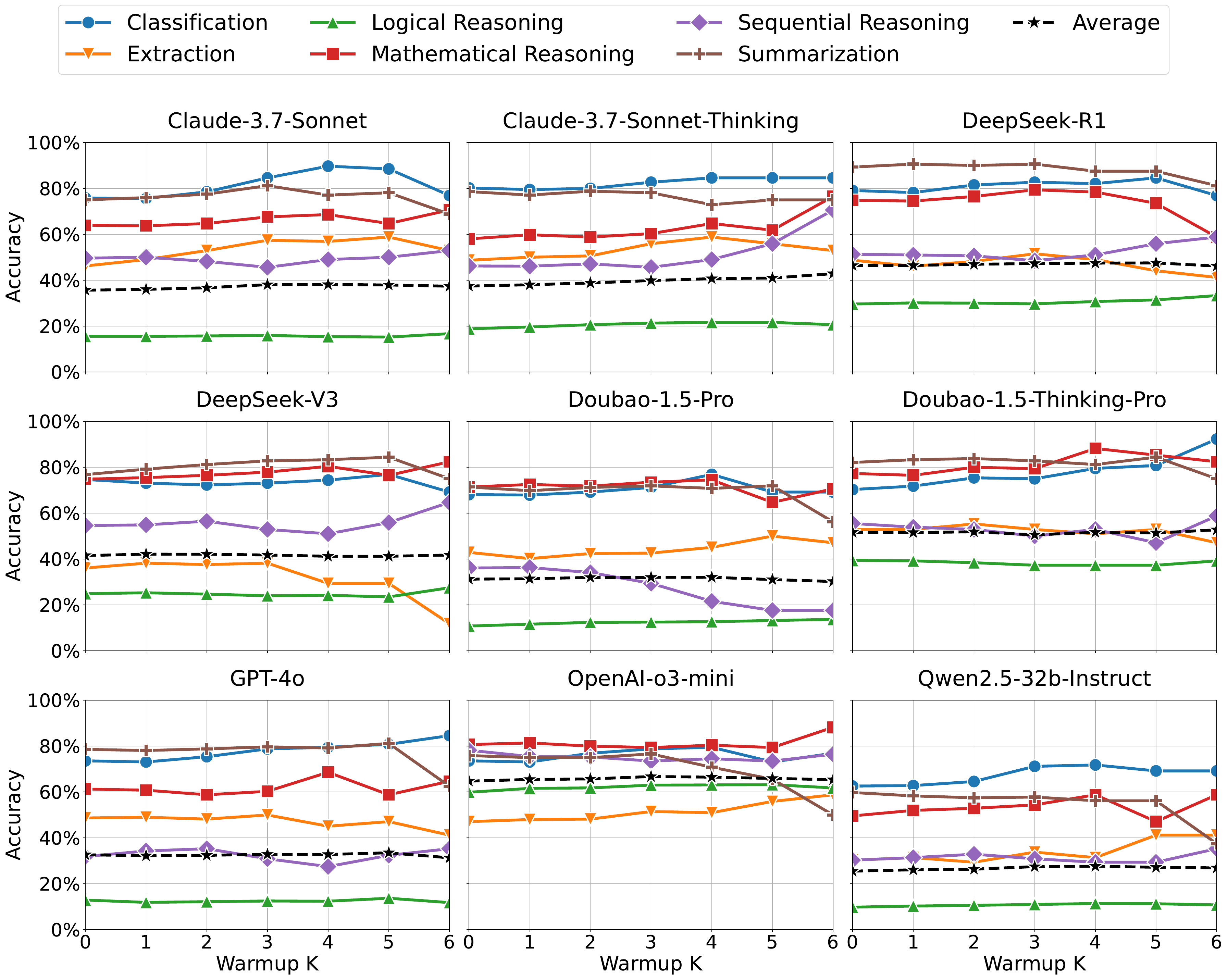}
    \caption{Results of post-warmup accuracy ($\mathrm{Acc}_{\mathrm{pw}}\text{-K}$) for feedback learning.}
    \label{fig:feedback-acc-per-k}
\end{figure}

\begin{figure}[htbp]
    \centering
    \includegraphics[width=1\linewidth]{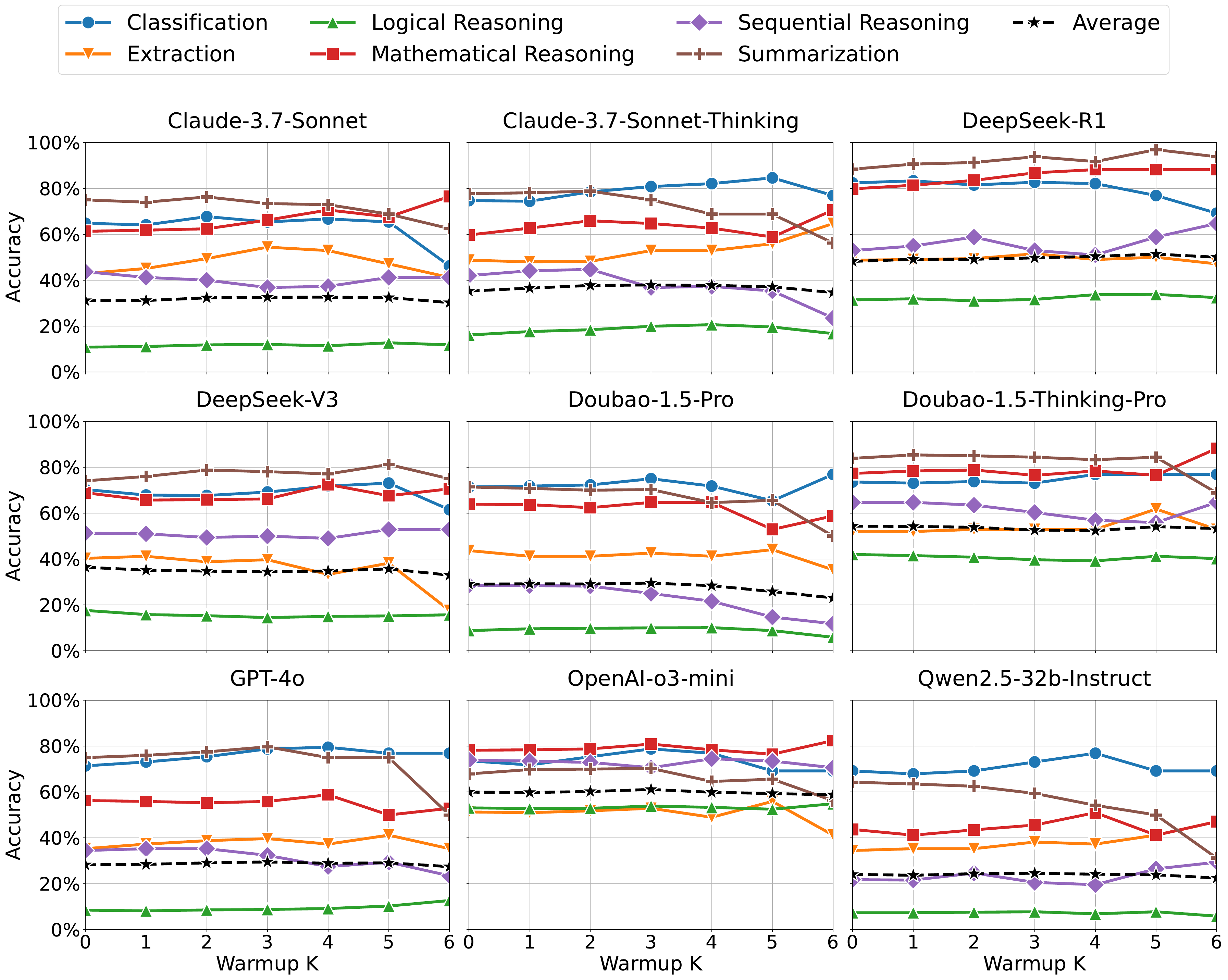}
    \caption{Results of post-warmup accuracy ($\mathrm{Acc}_{\mathrm{pw}}\text{-K}$) for demonstration learning.}
    \label{fig:demonstration-acc-per-k}
\end{figure}

Figures \ref{fig:feedback-acc-per-k} and \ref{fig:demonstration-acc-per-k} show the results of post-warmup accuracy ($\mathrm{Acc}_{\mathrm{pw}}\text{-K}$) for feedback learning and demonstration learning, respectively. 
This metric reflects model performance after an initial ``warmup'' phase, i.e., after some experience has been accumulated.

\clearpage

\subsection{Results of Position-wise Accuracy Curve}
\label{app:acc-per-position}

\begin{figure}[htbp]
    \centering
    \includegraphics[width=0.95\linewidth]{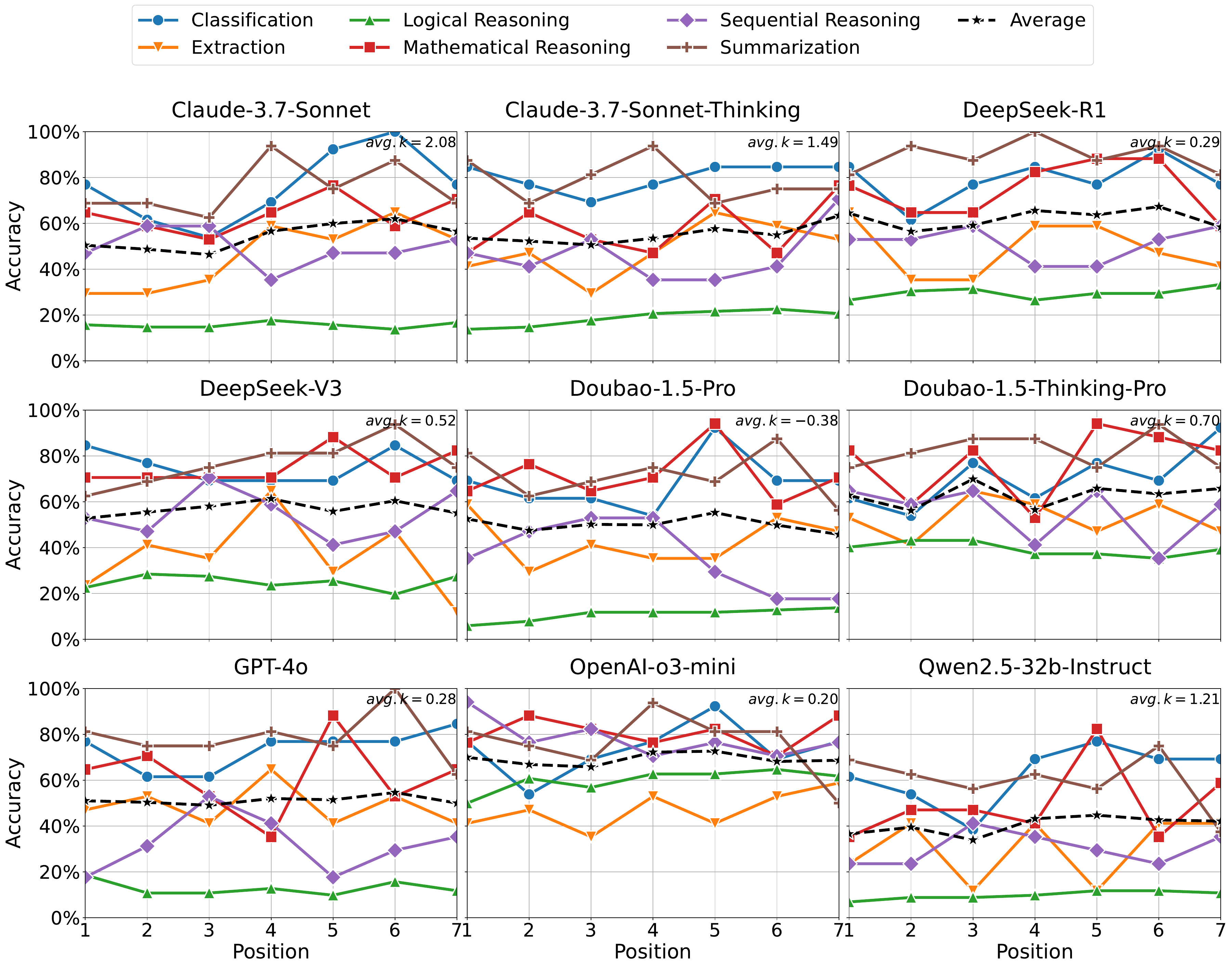}
    \caption{Position-wise Accuracy Curves for feedback learning.}
    \label{fig:Feedback-acc-per-position}
\end{figure}

\begin{figure}[htbp]
    \centering
    \includegraphics[width=0.95\linewidth]{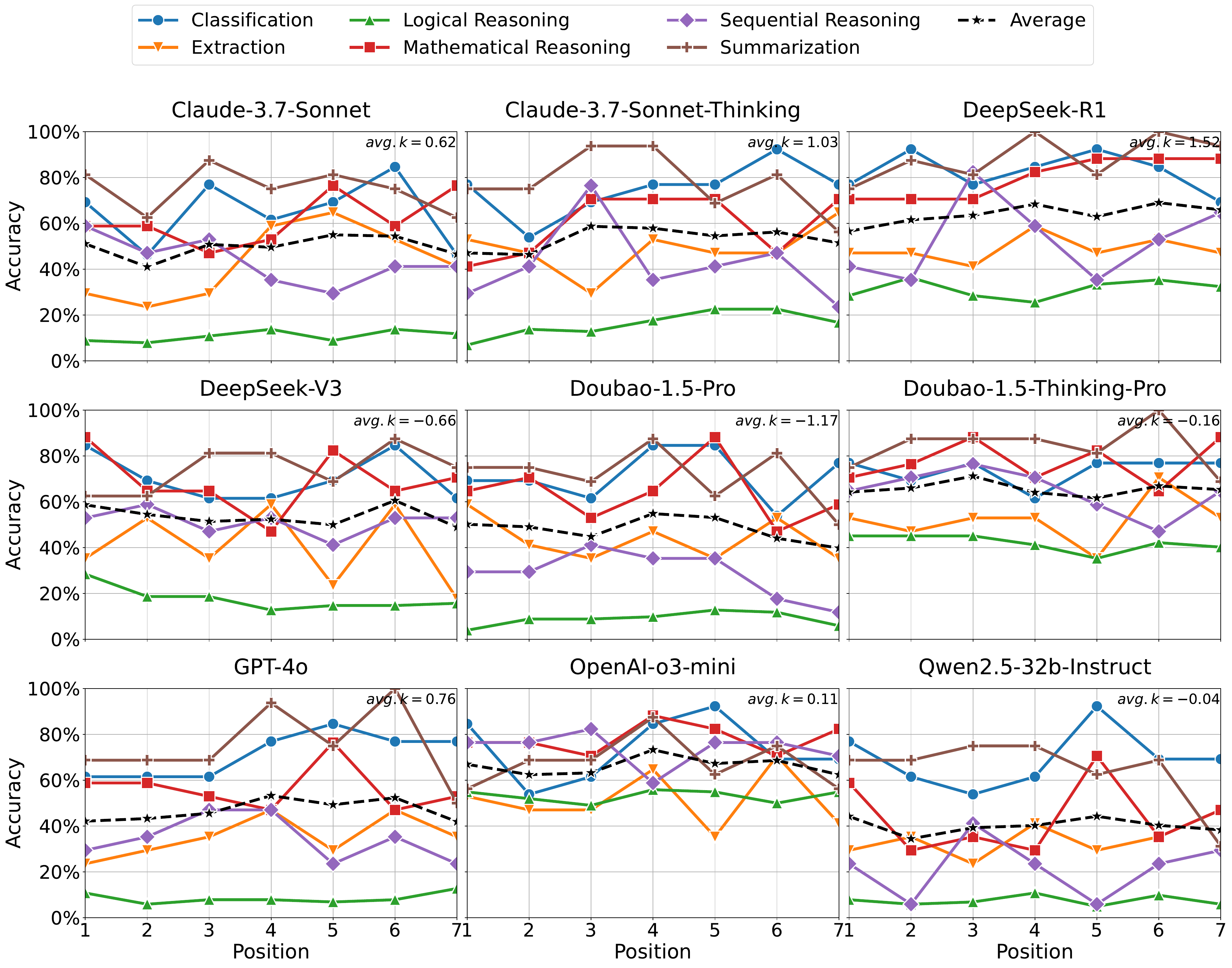}
    \caption{Position-wise Accuracy Curves for demonstration learning.}
    \label{fig:Demonstration-acc-per-position}
\end{figure}

Figures~\ref{fig:Feedback-acc-per-position} and~\ref{fig:Demonstration-acc-per-position} show the position-wise accuracy curves for feedback learning and demonstration learning, respectively.
These curves display the average accuracy at each position across all sequences.

\clearpage

\section{System Prompts}

\subsection{Evaluation Prompt for LLM-as-a-judge}

Figure \ref{fig:prompt-judge} presents the evaluation prompt used in LLM-as-a-judge.

\begin{tcolorbox}[breakable, title=Evaluation prompt for LLM-as-a-judge, colback=white,colframe=mydeepblue!90,colback=myblue!50,]
From now on, your role is a rigorous instruction-following grading teacher. Your task is to grade the student's answer strictly according to the standard answer.
You need to follow the steps below for grading. This is very important to me. Before you begin, please note the following two points:
\begin{enumerate}
    \item You have a two-level grading scale: 0 points and 1 point.0 points means the student's answer does not meet all the requirements in the standard answer. Each requirement in the standard answer is equally important. If even one requirement is not met, the score must be 0 points.1 point means the student's answer fully meets all the requirements in the standard answer.
    \item When you're ready to begin grading, remain calm and focused. Analyze and think through the problem step by step, following these steps:
    \begin{itemize}
        \item Carefully read and understand each requirement in the standard answer.
        \item Analyze whether the student's answer fully follows all the requirements in the standard answer, comparing each part of the student's answer with the standard answer.
        \item Do not rush to a conclusion. Before finalizing your grading, perform a Self-Reflection on the analysis: Ensure that your grading criteria take into account all the requirements of the standard answer and that no requirement is overlooked just because it seems "unimportant." Verify that your grading criteria and the score are logical and consistent. If there are any errors or omissions, correct them on time.
        \item Once you're confident your analysis is correct, assign the grade based on your analysis and display it in the following "JSON" format. Be strict in adhering to the output format requirements.
    \end{itemize}
\end{enumerate}

\textbf{Output Format:} 
Your output should follow this exact format: 

[Grading Rationale]:

[Score]: x points

[JSON]: \{"answer\_score": score\}

% verbatim常用于代码块展示，不缩进排版，原样保留字符\ {
% \begin{verbatim}
% \begin{lstlisting}
% \end{lstlisting}
% \end{verbatim}

\textbf{Example 1} 

<Standard Answer>: The student's answer must include an emoji after the word "jump rope". 

<Student's Answer>: Jump rope is an effective aerobic exercise that can help you burn calories. However, jump rope for weight loss requires consistent effort over time, and should be combined with a proper diet and other forms of exercise. If you want to lose weight through jump rope, it's recommended to do it for at least 30 minutes per day and gradually increase the difficulty and intensity. Also, watch your diet and avoid high-calorie, high-fat, high-sugar foods.

[Grading Rationale]: The word "jump rope" is not followed by an emoji. 

[Score]: 0 points

[JSON]: \{"answer\_score": 0\}

 \textbf{Example 2} \\ 
 <Standard Answer>: The student's answer must describe Beijing using a mix of Chinese and Korean. \\ 
 <Student's Answer>:
 \begin{CJK}{UTF8}{gbsn} 
 北京啊，北京是中国的首都，是中国的政治中心、文化中心、国际交往中心、科技创新中心。北京有着悠久的历史和丰富的文化遗产，如故宫、长城、颐和园等。北京还是中国的经济中心之一，拥有众多的跨国公司和金融机构。北京是一个充满活力和机遇的城市，吸引着来自世界各地的人们前来旅游、学习和工作。 
 \end{CJK}
 
[Grading Rationale]: The student's answer uses only Chinese and does not include any Korean as required.

[Score]: 0 points

[JSON]: \{"answer\_score": 0\} 

\textbf{Example 3}

<Standard Answer>: The student's answer must ask about the user's needs. \\
<Student's Answer>:Can you tell me what problem you're facing? That way, I can answer more accurately

[Grading Rationale]: The student asked about the user's needs, meeting all the requirements of the standard answer.
 
[Score]: 1 point

[JSON]:\{"answer\_score": 1\}

\textbf{Final Note:}
I hope you will be able to take on the role of a grading teacher effectively, as this is very important to my work. If you perform well, I will reward you appropriately. Otherwise, I may impose some penalties. Here is the official question:

\end{tcolorbox}
\begin{figure}[ht]
    \centering
    \vspace{-5pt}
    \caption{The evaluation prompt used in LLM-as-a-judge.}
    \label{fig:prompt-judge}
\end{figure}

\subsection{System Prompts for Different Solving Methods}

\begin{tcolorbox}[breakable,title=Prompt for Zero-shot,colback=white,colframe=mydeepblue!90,colback=myblue!50,]
 You are a student, you need to complete a question about \textbf{\{type\}} ability.
 
[Question]:\textbf{\{question\}}
\end{tcolorbox}
\begin{figure}[ht]
    \centering
    \vspace{-5pt}
    \caption{
System prompt for zero-shot (parallel solving).
    }
    \label{fig:Prompt for Zero-shot}
\end{figure}

\begin{tcolorbox}[breakable,title=Prompt for Few-shot,colback=white,colframe=mydeepblue!90,colback=myblue!50,]
You are a student, you need to complete a question about \textbf{\{type\}} ability. 

Before giving you the final question, I will provide some examples of other questions and their standard answers.

[History 1 Start]

[Question Start]

\textbf{\{example\_question\_1\}}

[Question End]

[Standard Answer Start]

\textbf{\{example\_answer\_1\}}

[Standard Answer End]

[History 1 End]

...

All history records have ended, now please begin answering the final question.

[Question]:\textbf{\{question\}}
\end{tcolorbox}
\begin{figure}[ht]
    \centering
    \vspace{-5pt}
    \caption{
System prompt for few-shot (parallel solving).
    }
    \label{fig:Prompt for Few-shot}
\end{figure}

\begin{tcolorbox}[breakable,title=Prompt for Demonstration Learning,colback=white,colframe=mydeepblue!90,colback=myblue!50,]
You are a student, you need to complete a question about \textbf{\{type\}} ability. 

Before giving you the final question, I will provide some examples of other questions and their standard answers.

[History 1 Start]

[Question Start]

\textbf{\{previous\_question\_1\}}

[Question End]

[Standard Answer Start]

\textbf{\{previous\_answer\_1\}}

[Standard Answer End]

[History 1 End]

...

All history records have ended, now please begin answering the final question.

[Question]:\textbf{\{question\}}
\end{tcolorbox}
\begin{figure}[ht]
    \centering
    \vspace{-5pt}
    \caption{
System prompt for demonstration learning (sequential solving).
    }
    \label{fig:Prompt for Demonstration Learning}
\end{figure}

\begin{tcolorbox}[breakable,title=Prompt for Feedback Learning,colback=white,colframe=mydeepblue!90,colback=myblue!50,]
You are a student, you need to complete a question about \textbf{\{type\}} ability. 

Before giving you the final question, I will provide some examples of other questions and their standard answers.

[History 1 Start]

[Question Start]

\textbf{\{example\_question\_1\}}

[Question End]

[Answer Evaluation Criteria Start]

\textbf{\{previous\_rubric\_1\}}

[Answer Evaluation Criteria End]

[Teacher's Evaluation of Student's Answer Start]

\textbf{\{previous\_judge\_1\}}

[Teacher's Evaluation of Student's Answer End]

[History 1 End]

...

All history records have ended, now please begin answering the final question.

[Question]:\textbf{\{question\}}
\end{tcolorbox}
\begin{figure}[ht]
    \centering
    \vspace{-5pt}
    \caption{
System prompt for feedback learning (sequential solving).
    }
    \label{fig:Prompt for Feedback Learning}
\end{figure}

\section{Instances}
\label{sec:dataset-case}

Figures \ref{fig:datasets-case1} and \ref{fig:datasets-case2} showcase two instances in the logical reasoning task of EvaLearn.

The instance in Figure~\ref{fig:datasets-case1} involves solving a $4\times4$ sliding puzzle. 
In this case, the puzzle consists of 16 tiles, one of which is a blank space (denoted by `>'), while the remaining tiles are filled with different symbols. 
By moving the blank space and swapping it with adjacent tiles, the goal is to restore the puzzle from a given initial state to a target configuration through a series of logical moves. 
This task is designed to assess the model's logical reasoning ability, and all problems within this sequence are centered around the sliding puzzles and logical reasoning.

Figure~\ref{fig:datasets-case2} presents another problem from the same sequence, which differs mainly in grid size, the initial state of the puzzle, and symbol representation. 
These problems are interrelated and collectively challenge the model's logical reasoning skills. 
Humans are able to gain experience and improve their performance by solving a series of such problems. 
By requiring models to solve these problems sequentially within a sequence, EvaLearn evaluates whether models can similarly learn from experience, thereby measuring their learning efficiency.

% Case Study 3
\begin{tcolorbox}[breakable,title=Instance 1 in the Logic Reasoning task,colback=white,colframe=mydeepblue!90,colback=myblue!50,]

\textbf{Problem Unique Id}: $557$ 

\textbf{Task Type}: Logical Reasoning

\textbf{Prompt}: 

You are a virtual expert in solving a $4 \times 4$ sliding puzzle problem. Please follow the instructions and rules below to complete the solution. Your goal is to reach the goal state with valid moves.

% \texttt{[The goal state]}\\[0.5em]
[The goal state]\\[0.5em]
\begin{tabular}{llll}
\{ & D & $>$ & - \\
q & ] & \_ & * \\
Z & n & N & C \\
P & S & | & E \\
\end{tabular} \\

[Instructions]\\
The puzzle consists of a $4 \times 4$ grid containing various symbols, with \textquotesingle >\textquotesingle representing the movable space. Only \textquotesingle >\textquotesingle can be moved horizontally or vertically, and the objective is to reach the goal state from a given initial state. \\

[Rules] \\
1. Only \textquotesingle >\textquotesingle can be moved horizontally or vertically.\\
2. Each move is chosen from the following set of options: \\

- \textquotesingle Left\textquotesingle: move \textquotesingle >\textquotesingle to the left; \\
- \textquotesingle Down\textquotesingle: move \textquotesingle >\textquotesingle downward; \\
- \textquotesingle Right\textquotesingle: move \textquotesingle >\textquotesingle to the right; \\
- \textquotesingle Up\textquotesingle: move \textquotesingle >\textquotesingle upward \\

3. The next move must be chosen from the valid move set depending on the position of \textquotesingle >\textquotesingle.\\
For example:\\
\begin{tabular}{llll}
p1 & p2 & p3 & p4 \\
p5 & p6 & p7 & p8 \\
p9 & p10 & p11 & p12 \\
p13 & p14 & p15 & p16 \\
\end{tabular}

- If \textquotesingle >\textquotesingle is located at position \textquotesingle 
 p1\textquotesingle, the valid move set is [\textquotesingle Down\textquotesingle, \textquotesingle Right\textquotesingle];\\
- If \textquotesingle >\textquotesingle is located at position \textquotesingle p2\textquotesingle, the valid move set is [\textquotesingle Down\textquotesingle, \textquotesingle Left\textquotesingle, \textquotesingle Right\textquotesingle];\\
- If \textquotesingle > \textquotesingle is located at position \textquotesingle p3 \textquotesingle, the valid move set is [\textquotesingle Down \textquotesingle, \textquotesingle Left \textquotesingle, \textquotesingle Right \textquotesingle];\\
- If \textquotesingle > \textquotesingle is located at position \textquotesingle p4 \textquotesingle, the valid move set is [\textquotesingle Down \textquotesingle, \textquotesingle Left \textquotesingle];\\
- If \textquotesingle > \textquotesingle is located at position \textquotesingle p5 \textquotesingle, the valid move set is [\textquotesingle Up \textquotesingle, \textquotesingle Down \textquotesingle, \textquotesingle Right \textquotesingle];\\
- If \textquotesingle > \textquotesingle is located at position \textquotesingle p6 \textquotesingle, the valid move set is [\textquotesingle Up \textquotesingle, \textquotesingle Down \textquotesingle, \textquotesingle Left \textquotesingle, \textquotesingle Right \textquotesingle];\\
- If \textquotesingle > \textquotesingle is located at position \textquotesingle p7 \textquotesingle, the valid move set is [\textquotesingle Up \textquotesingle, \textquotesingle Down \textquotesingle, \textquotesingle Left \textquotesingle, \textquotesingle Right \textquotesingle];\\
- If \textquotesingle > \textquotesingle is located at position \textquotesingle p8 \textquotesingle, the valid move set is [\textquotesingle Up \textquotesingle, \textquotesingle Down \textquotesingle, \textquotesingle Left \textquotesingle];\\
- If \textquotesingle > \textquotesingle is located at position \textquotesingle p9 \textquotesingle, the valid move set is [\textquotesingle Up \textquotesingle, \textquotesingle Down \textquotesingle, \textquotesingle Right \textquotesingle];\\
- If \textquotesingle > \textquotesingle is located at position \textquotesingle p10 \textquotesingle , the valid move set is [\textquotesingle Up \textquotesingle , \textquotesingle Down \textquotesingle , \textquotesingle Left \textquotesingle , \textquotesingle Right \textquotesingle ];\\
- If \textquotesingle > \textquotesingle is located at position \textquotesingle p11 \textquotesingle , the valid move set is [\textquotesingle Up \textquotesingle , \textquotesingle Down \textquotesingle , \textquotesingle Left \textquotesingle , \textquotesingle Right \textquotesingle ];\\
- If \textquotesingle > \textquotesingle is located at position \textquotesingle p12 \textquotesingle , the valid move set is [\textquotesingle Up \textquotesingle , \textquotesingle Down \textquotesingle , \textquotesingle Left \textquotesingle ];\\
- If \textquotesingle > \textquotesingle is located at position \textquotesingle p13 \textquotesingle , the valid move set is [\textquotesingle Up \textquotesingle , \textquotesingle Right \textquotesingle ];\\
- If \textquotesingle > \textquotesingle is located at position \textquotesingle p14 \textquotesingle , the valid move set is [\textquotesingle Up \textquotesingle , \textquotesingle Left \textquotesingle , \textquotesingle Right \textquotesingle ];\\
- If \textquotesingle > \textquotesingle is located at position \textquotesingle p15 \textquotesingle , the valid move set is [\textquotesingle Up \textquotesingle , \textquotesingle Left \textquotesingle , \textquotesingle Right \textquotesingle ];\\
- If \textquotesingle > \textquotesingle is located at position \textquotesingle p16 \textquotesingle, the valid move set is [\textquotesingle Up \textquotesingle, \textquotesingle Left \textquotesingle]\\

4. Diagonal moves are not allowed. \\
5. The objective is to return the moves which can reach the goal state. \\

Please solve the following problem. Your answer should be in the format of a list of moves, for example:[\textquotesingle Left \textquotesingle, \textquotesingle Up \textquotesingle, \textquotesingle Up \textquotesingle].\\
This problem can be solved within $1$ to $12$ steps. The next move must be chosen from the valid move set [\textquotesingle Left \textquotesingle, \textquotesingle Right \textquotesingle, \textquotesingle Up \textquotesingle, \textquotesingle Down \textquotesingle]. The maximum step number you can take is $12$. Try to reach the goal state using the least number of steps ($\leq 12$)
). DO NOT exceed $12$ steps.

[Initial State]\\[0.5em]
\begin{tabular}{llll}
q & \{ & \_ & - \\
Z & D & N & * \\
P & ] & | & C \\
$>$ & n & S & E \\
\end{tabular}

\textbf{Canonical Answer}:

[ \textquotesingle Up \textquotesingle , \textquotesingle Up \textquotesingle , \textquotesingle Up \textquotesingle , \textquotesingle Right \textquotesingle , \textquotesingle Down \textquotesingle , \textquotesingle Down \textquotesingle , \textquotesingle Down \textquotesingle , \textquotesingle Right \textquotesingle , \textquotesingle Up \textquotesingle , \textquotesingle Up \textquotesingle , \textquotesingle Up \textquotesingle ] \\
\textbf{Rubric}:

The student's final answer must exactly match the content and order of the standard answer.The reasoning process, letter case, extra spaces, and other formatting issues can be ignored,but the final answer must contain all the correct elements of the standard answer.\\
Standard answer: 

[ \textquotesingle Up \textquotesingle , \textquotesingle Up \textquotesingle , \textquotesingle Up \textquotesingle , \textquotesingle Right \textquotesingle , \textquotesingle Down \textquotesingle , \textquotesingle Down \textquotesingle , \textquotesingle Down \textquotesingle , \textquotesingle Right \textquotesingle , \textquotesingle Up \textquotesingle , \textquotesingle Up \textquotesingle , \textquotesingle Up \textquotesingle ]\\
If the student's answer differs in content from the standard answer (after ignoring case and formatting), the score is $0$.\\
\textbf{Sequence Id}: $51$\\
\textbf{Position In Sequence}: $1$\\

\end{tcolorbox}
\begin{figure}[ht]
    \centering
    \vspace{-5pt}
\caption{
Instance 1 in the logical reasoning task of EvaLearn.
    }
    \label{fig:datasets-case1}
\end{figure}

\begin{tcolorbox}[breakable,title=Instance 2 in the Logic Reasoning task,colback=white,colframe=mydeepblue!90,colback=myblue!50,]

\textbf{Problem Unique Id}: $557$ 

\textbf{Task Type}: Logical Reasoning

\textbf{Prompt}: 

You are a virtual expert in solving a $6 \times 6$ sliding puzzle problem. Please follow the instructions and rules below to complete the solving. Your goal is to reach the goal state with valid moves.

[The goal state]\\[0.5em]
\begin{tabular}{cccccc}
\texttt{a} & \texttt{9} & \texttt{\~{}} & \texttt{\}} & \texttt{[} & \texttt{e} \\
\texttt{o} & \texttt{8} & \texttt{i} & \texttt{E} & \texttt{S} & \texttt{-} \\
\texttt{!} & \texttt{+} & \texttt{w} & \texttt{l} & \texttt{P} & \texttt{=} \\
\texttt{q} & \texttt{B} & \texttt{0} & \texttt{6} & \texttt{n} & \texttt{F} \\
\texttt{7} & \texttt{<} & \texttt{3} & \texttt{\$} & \texttt{K} & \texttt{A} \\
\texttt{Y} & \texttt{u} & \texttt{\&} & \texttt{p} & \texttt{O} & \texttt{U} \\
\end{tabular} \\

[Instructions]\\
The puzzle consists of a $6 \times 6$ grid containing various symbols, with \textquotesingle B\textquotesingle\ representing the movable space. Only \textquotesingle B\textquotesingle\ can be moved horizontally or vertically, and the objective is to reach the goal state from a given initial state.\\

[Rules]\\
1. Only \textquotesingle B\textquotesingle\ can be moved horizontally or vertically.\\
2. Each move is chosen from the following set of options: \\

- \textquotesingle Left\textquotesingle: move \textquotesingle B\textquotesingle to the left;\\
- \textquotesingle Down\textquotesingle: move \textquotesingle B\textquotesingle downward;\\
- \textquotesingle Right\textquotesingle: move \textquotesingle B\textquotesingle to the right;\\
- \textquotesingle Up\textquotesingle: move \textquotesingle B\textquotesingle upward\\

3. The next move must be chosen from the valid move set depending on the position of \textquotesingle B\textquotesingle.\\
For example:\\

\begin{tabular}{cccccc}
p1 & p2 & p3 & p4 & p5 & p6 \\
p7 & p8 & p9 & p10 & p11 & p12 \\
p13 & p14 & p15 & p16 & p17 & p18 \\
p19 & p20 & p21 & p22 & p23 & p24 \\
p25 & p26 & p27 & p28 & p29 & p30 \\
p31 & p32 & p33 & p34 & p35 & p36 \\
\end{tabular} \\
- If 'B' is located at position 'p1', the valid move set is ['Down', 'Right'];\\
- If 'B' is located at position 'p2', the valid move set is ['Down', 'Left', 'Right'];\\
- If 'B' is located at position 'p3', the valid move set is ['Down', 'Left', 'Right'];\\
- If 'B' is located at position 'p4', the valid move set is ['Down', 'Left', 'Right'];\\
- If 'B' is located at position 'p5', the valid move set is ['Down', 'Left', 'Right'];\\
- If 'B' is located at position 'p6', the valid move set is ['Down', 'Left'];\\
- If 'B' is located at position 'p7', the valid move set is ['Up', 'Down', 'Right'];\\
- If 'B' is located at position 'p8', the valid move set is ['Up', 'Down', 'Left', 'Right'];\\
- If 'B' is located at position 'p9', the valid move set is ['Up', 'Down', 'Left', 'Right'];\\
- If 'B' is located at position 'p10', the valid move set is ['Up', 'Down', 'Left', 'Right'];\\
- If 'B' is located at position 'p11', the valid move set is ['Up', 'Down', 'Left', 'Right'];\\
- If 'B' is located at position 'p12', the valid move set is ['Up', 'Down', 'Left'];\\
- If 'B' is located at position 'p13', the valid move set is ['Up', 'Down', 'Right'];\\
- If 'B' is located at position 'p14', the valid move set is ['Up', 'Down', 'Left', 'Right'];\\
- If 'B' is located at position 'p15', the valid move set is ['Up', 'Down', 'Left', 'Right'];\\
- If 'B' is located at position 'p16', the valid move set is ['Up', 'Down', 'Left', 'Right'];\\
- If 'B' is located at position 'p17', the valid move set is ['Up', 'Down', 'Left', 'Right'];\\
- If 'B' is located at position 'p18', the valid move set is ['Up', 'Down', 'Left'];\\
- If 'B' is located at position 'p19', the valid move set is ['Up', 'Down', 'Right'];\\
- If 'B' is located at position 'p20', the valid move set is ['Up', 'Down', 'Left', 'Right'];\\
- If 'B' is located at position 'p21', the valid move set is ['Up', 'Down', 'Left', 'Right'];\\
- If 'B' is located at position 'p22', the valid move set is ['Up', 'Down', 'Left', 'Right'];\\
- If 'B' is located at position 'p23', the valid move set is ['Up', 'Down', 'Left', 'Right'];\\
- If 'B' is located at position 'p24', the valid move set is ['Up', 'Down', 'Left'];\\
- If 'B' is located at position 'p25', the valid move set is ['Up', 'Down', 'Right'];\\
- If 'B' is located at position 'p26', the valid move set is ['Up', 'Down', 'Left', 'Right'];\\
- If 'B' is located at position 'p27', the valid move set is ['Up', 'Down', 'Left', 'Right'];\\
- If 'B' is located at position 'p28', the valid move set is ['Up', 'Down', 'Left', 'Right'];\\
- If 'B' is located at position 'p29', the valid move set is ['Up', 'Down', 'Left', 'Right'];\\
- If 'B' is located at position 'p30', the valid move set is ['Up', 'Down', 'Left'];\\
- If 'B' is located at position 'p31', the valid move set is ['Up', 'Right'];\\
- If 'B' is located at position 'p32', the valid move set is ['Up', 'Left', 'Right'];\\
- If 'B' is located at position 'p33', the valid move set is ['Up', 'Left', 'Right'];\\
- If 'B' is located at position 'p34', the valid move set is ['Up', 'Left', 'Right'];\\
- If 'B' is located at position 'p35', the valid move set is ['Up', 'Left', 'Right'];\\
- If 'B' is located at position 'p36', the valid move set is ['Up', 'Left'];\\

4. Diagonal moves are not allowed.\\
5. The objective is to return the moves which can reach the goal state. Please solve the following problem. Your answer should be in the format of a list of moves, for example: ['Left', 'Up', 'Up'].\\
This problem can be solved within $1$ to $18$ steps. The next move must be chosen from the valid move set ['Left', 'Right', 'Up', 'Down']. The maximum step number you can take is $18$. Try to reach the goal state using the least number of steps ($\leq 18$). DO NOT exceed $18$ steps.

[Initial State]\\[0.5em]
\begin{tabular}{cccccc}
\texttt{o} & \texttt{B} & \texttt{\~{}} & \texttt{\}} & \texttt{[} & \texttt{e} \\
\texttt{!} & \texttt{a} & \texttt{i} & \texttt{E} & \texttt{S} & \texttt{-} \\
\texttt{q} & \texttt{9} & \texttt{w} & \texttt{l} & \texttt{P} & \texttt{=} \\
\texttt{7} & \texttt{8} & \texttt{0} & \texttt{6} & \texttt{n} & \texttt{F} \\
\texttt{Y} & \texttt{+} & \texttt{3} & \texttt{\$} & \texttt{K} & \texttt{A} \\
\texttt{u} & \texttt{<} & \texttt{\&} & \texttt{p} & \texttt{O} & \texttt{U} \\
\end{tabular} \\

\textbf{Canonical Answer}: \\
\texttt{['Down', 'Down', 'Down', 'Down', 'Down', 'Left', 'Up', 'Up', 'Up', 'Up', 'Up', 'Right', 'Down', 'Down', 'Down']}

\textbf{Rubric}: The student's final answer must exactly match the content and order of the standard answer. The reasoning process, letter case, extra spaces, and other formatting issues can be ignored, but the final answer must contain all the correct elements of the standard answer.

Standard answer: \\
\texttt{['Down', 'Down', 'Down', 'Down', 'Down', 'Left', 'Up', 'Up', 'Up', 'Up', 'Up', 'Right', 'Down', 'Down', 'Down']}

If the student's answer differs in content from the standard answer (after ignoring case and formatting), the score is $0$.

\textbf{Sequence Id}: $51$ \\
\textbf{Position In Sequence}: $6$ \\

\end{tcolorbox}

\begin{figure}[ht]
    \centering
    \vspace{-5pt}
\caption{
Instance 2 in the logical reasoning task of EvaLearn.
This instance and Figure~\ref{fig:datasets-case1} are from the same sequence.
The two problems differ primarily in grid size, the initial state of the puzzle, and symbol representation.
    }
    \label{fig:datasets-case2}
\end{figure}

\end{document}